\documentclass[a4paper,fleqn]{cas-sc}

\usepackage[numbers]{natbib}
\usepackage{graphicx}
\usepackage{caption}
\usepackage{subcaption}
\usepackage{algorithm,algpseudocode,algorithmicx}
\usepackage{bm}
\usepackage{tikz}
\usepackage{amsmath}

\def\tsc#1{\csdef{#1}{\textsc{\lowercase{#1}}\xspace}}
\tsc{WGM}
\tsc{QE}
\tsc{EP}
\tsc{PMS}
\tsc{BEC}
\tsc{DE}

\begin{document}
\let\WriteBookmarks\relax
\def\floatpagepagefraction{1}
\def\textpagefraction{.001}
\shorttitle{Federated Learning on Non-IID Data}
\shortauthors{Hangyu Zhu et~al.}

\title [mode = title]{Federated Learning on Non-IID Data: A Survey}                      



\author[1]{Hangyu Zhu}
\ead{hangyu.zhu@surrey.ac.uk}


\address[1]{Department of Computer Science, University of Surrey, Guildford, Surrey GU2 7XH, Surrey, UK}

\author[2]{Jinjin Xu}
\ead{jin.xu@mail.ecust.edu.cn}

\author[1]{Shiqing Liu}
\ead{s.liu@surrey.ac.uk}


\address[2]{Key Laboratory of Smart Manufacturing in Energy Chemical Process, Ministry of Education, East China University of Science of Technology, Shanghai, 200237, China.}


\author[1]%
{Yaochu Jin}
\cormark[1]
\ead{yaochu.jin@surrey.ac.uk}


\cortext[cor1]{Corresponding author}


\begin{abstract}
Federated learning is an emerging distributed machine learning framework for privacy preservation. However, models trained in federated learning usually have worse performance than those trained in the standard centralized learning mode, especially when the training data are not independent and identically distributed (Non-IID) on the local devices. In this survey, we provide a detailed analysis of the influence of Non-IID data on both parametric and non-parametric machine learning models in both horizontal and vertical federated learning. In addition, current research work on handling challenges of Non-IID data in federated learning are reviewed, and both advantages and disadvantages of these approaches are discussed. Finally, we suggest several future research directions before concluding the paper. 
\end{abstract}

\maketitle

\section{Introduction}
\label{intro}

Traditional centralized learning requires all data collected on local devices such as mobile phones to be stored centrally on a data center or cloud server. This requirement not only raises the concern of privacy risks and data leakage, but also poses high demands on storage and computing capacities of the server when the amount of data is huge. Although distributed data parallelism \cite{dean2012large}, which enables multiple machines to train a model replica with different data groups in parallel, may serve as a potential solution to the issue of storage and computational capacity, it still needs access to the whole training data to split it into evenly distributed shards, causing possible security and privacy problems to the data.

Federated learning (FL) aims to train a global model that can be trained on data distributed on different devices while protecting the data privacy. In 2016, McMahan \textit{et al}., for the first time, introduce the concept of FL \cite{mcmahan2017communication} based on data parallelism and proposed a Federated Averaging (FedAvg) algorithm. As a decentralized machine learning approach, FedAvg allows multiple devices to train a machine learning model cooperatively, while keeping the user data stored locally. FedAvg obviates the need for uploading the users' sensitive data to a centralized server, and makes it possible for the edge devices to train a shared model locally within their own local dataset. By aggregating the updates (gradients) of local models, FedAvg meets the basic requirements for privacy protection and data security.  


While FL provides a promising approach to privacy protection, many challenges arise in comparison with centralized learning when FL is applied to the real world \cite{yang2019federated}. These include communication cost needed for transmitting parameters between the server and local devices, computing power and energy consumption required for local devices, and the heterogeneity and randomness of possibly a huge number of local devices in the learning process. A large body of research to address the above challenges, including reducing communication cost \cite{chen2019communication,mcmahan2017communication,mills2019communication,xu2020}, FL considering hardware constraints \cite{duan2019astraea,lim2020federated}, and additional protections against adversarial attacks \cite{kairouz2019advances,254465,zhu2020distributed}.

Although the authors in \cite{mcmahan2017communication} claim that FedAvg is able to cope with the not independent and identically distributed (Non-IID) data to a certain degree, a lot of research have indicated that a deterioration in accuracy of FL is almost inevitable on Non-IID or heterogeneous data \cite{zhao2018federated}. The performance degradation can mainly be attributed to weight divergence of the local models resulting from Non-IID. That is, local models having the same initial parameters will convergence to different models because of the heterogeneity in local data distributions. During the FL, the divergence between the shared global model acquired by averaging the uploaded local models and the ideal model (the model obtained when the data on the local devices is IID) continues to increase, slowing down the convergence and worsening the learning performance.

Due to the rapid increase of research interest in FL, several valuable review papers on federated learning have been published in the literature. A general introduction to FL and its applications are given in \cite{yang2019federated,ZHANG2021106775}, detailed discussions of advances and challenges can be found in \cite{kairouz2019advances,li2020federated}. Analyses of threats and additional privacy preservation techniques in FL are presented in \cite{vepakomma2018no,li2019survey,briggs2020review,lyu2020threats}. Overviews of FL applications to IoT and edge devices \cite{lan2019introduction,imteaj2020federated,shi2020communication}, wireless networks \cite{hosseinalipour2020federated}, mobile devices \cite{lim2020federated}, and healthcare \cite{xujie2021federated} have also been reported. 

Although Kulkarni \textit{et al}. \cite{kulkarni2020survey} have provided a brief introduction to personalization approaches to handling Non-IID data in FL, none of the existing work have explored the impact of Non-IID data on FL in great detail. To fill the gap, this paper is dedicated to a comprehensive survey of FL on Non-IID data, including an in-depth analysis of various data distributions, their influences on model aggregation, a categorization and discussion of pros and cons of existing techniques for handling skewed data distributions, and an outline of remaining challenges and future research in the area of FL in the presence of Non-IID data.

\section{Federated Learning}
\label{fl}
FL aims to find an optimal global model $\theta$ (Eq. \eqref{eq:global_loss}) that can minimize the aggregated local loss function $f_k(\theta^k)$ (Eq. \eqref{eq:local loss}), where $\bm{x}$ is the data feature, $y$ is the data label, $n_{k}$ is the local data size, $n=\sum_{k=1}^{C \times K} n_k$ is the total number of sample pairs, $C$ is the participation ratio assuming that not all local clients participate in each round of model updates, $l$ is the loss function and $k$ is the client index.

\begin{equation}
\label{eq:local loss}
f_k(\theta^k) = \frac{1}{n_k}\sum_i^{n_k}l(\bm{x}_i,y_i;\theta^k) \\
\end{equation}

\begin{equation}
\label{eq:global_loss}
\min_\theta f(\theta) =  \sum_{k=1}^{C \times K}\frac{n_{k}}{n} f_k(\theta^k)
\end{equation}

Generally speaking, FL can be categorized into horizontal and vertical FL according to characteristics of data distribution among the connected clients, which was originally defined in the paper \cite{yang2019federated}. In the following, we provide a brief introduction to these two FL frameworks before we discussing various data distributions. 

\subsection{Horizontal Federated Learning}
\begin{figure}
\centering 
\includegraphics[width=0.92\textwidth]{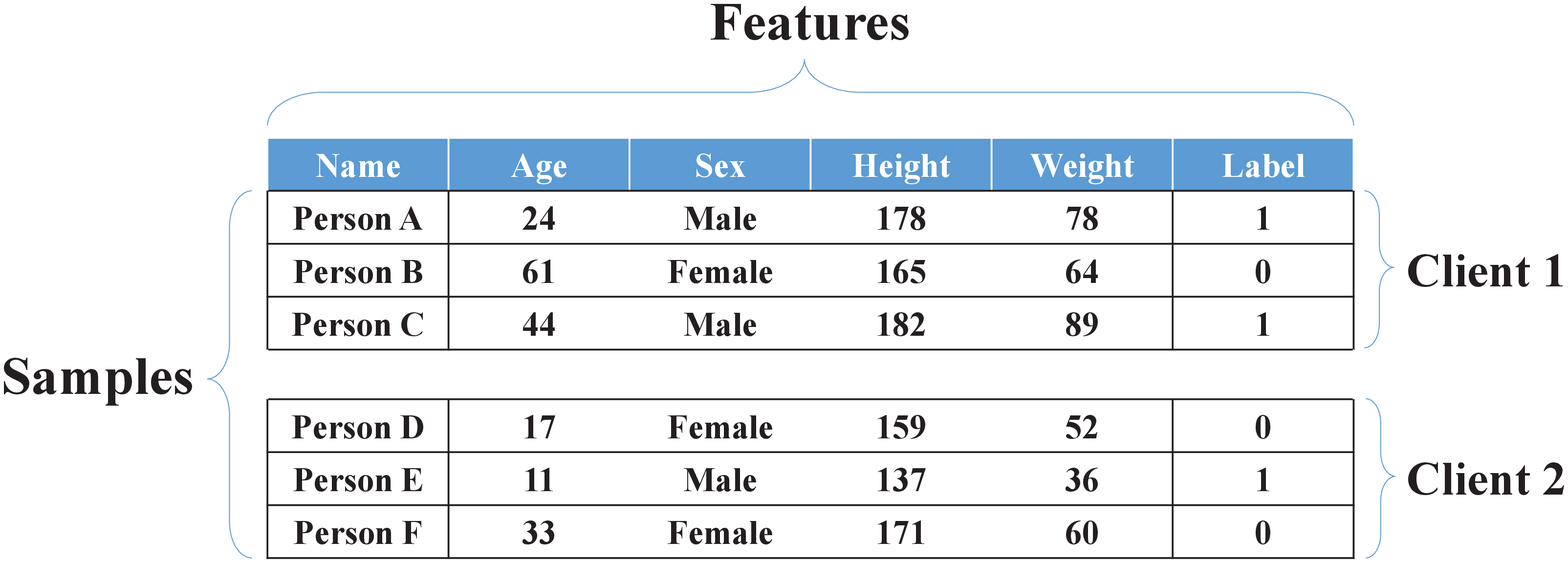} 
\caption{An illustrative example of data partition in horizontal federated learning. There are two clients, each containing five personal features including name, age, sex, height and weight. Client 1 has the data of Persons A, B, and C, while Client 2 stores the data of Persons D, E and F.} 
\label{hfl}
\end{figure}

Horizontal FL is also referred to homogeneous FL \cite{haddadpour2019convergence}, which represents the scenarios in which the training data of participating clients share the same feature space but have different sample space. As a simple example shown in Fig. \ref{hfl}, Client 1 and Client 2 contain different rows of data with the same personal features and each row indicates the particular data for one particular person. 

The FedAvg algorithm \cite{mcmahan2017communication} is a typical horizontal FL algorithm and its pseudo code is presented in Algorithm \ref{fedavg}, where $m=C\times K$ is the number of participating clients. Both the global model ${\theta _t}$ (where $t$ is the communication round) and all local models ${\theta ^k}$ have the same model structure with different model parameter values. Under this assumption, direct model aggregation can be implemented as described in line 9 of Algorithm \ref{fedavg}, in which all uploaded local models ${\theta ^k}$ are weighted averaged based on the ratio $\frac{n_{k}}{n}$, which is proportional to the amount of data on each client, to generate the global model ${\theta _t}$.

\begin{algorithm}[htbp]\footnotesize{
\caption{FedAvg. $K$ is the total numbers of clients; $B$ is the size of mini-batches, $T$ is the total number of communication rounds, $E$ is the total local training epochs, and $\eta$ is the learning rate.} 
\algblock{Begin}{End}
\label{fedavg}
\begin{algorithmic}[1]
\State \textbf{Server: }
\State Initialize global model ${\theta_{0}}$
\For {each communication round $ t = 1,2,...T\ $}
\State Select $m = C \times K$ clients, where $C \in (0,1)$ 
\For {each \textbf{Client} $k = 1,2,...m$ in parallel}
\State Download ${\theta _t}$ to \textbf{Client $k$}
\State Do \textbf{Client $k$} update and receive ${\theta ^k}$ 
\EndFor
\State Update global model ${\theta _t} \leftarrow \sum\limits_{k = 1}^m {\frac{{{n_k}}}{n}{\theta ^k}}$
\EndFor \\
\State \textbf{Client $k$} update: 
\State Replace local model ${\theta ^k} \leftarrow {\theta _t}$
\For {local epoch from 1 to $E$}
\For {batch $ b \in (1,B)\ $}
\State $ {\theta ^k} \leftarrow {\theta ^k} - \eta \nabla {L_k}({\theta ^k},b)\ $
\EndFor
\EndFor
\State \textbf{Return} ${\theta ^k}$ 
\end{algorithmic}}
\end{algorithm}

Compared to the standard centralized learning paradigm, horizontal FL provides a simple yet effective solution to prevent private local data from being leaked, because only the global model parameters $\theta_t$ and local model parameters ${\theta ^k}$ are allowed to be communicated between the server and clients, and all the training data are kept on the client devices without being accessed by any other parties. 

However, frequently downloading and uploading model parameters will consume much communication resources. Therefore, many techniques have been proposed to reduce communication costs in horizontal FL, such as client updates sub-sampling \cite{shokri2015privacy,konevcny2016federated,caldas2018expanding} and model quantization \cite{xu2020,DBLP:journals/corr/HanMD15,wen2017terngrad,mills2019communication,sattler2019robust}. Besides, Chen \emph{et al.} \cite{chen2019communication} suggest to reduce the communication frequency of deep layers of the neural network model to enhance the communication efficiency. In addition, Zhu \emph{et al.} \cite{zhu2019multi} use a multi-objective evolutionary algorithm to simultaneously increase the model performance and decrease communication costs.

Although no private data can be directly accessed by any third party, the uploaded model parameters or gradients of each client may still leak data information \cite{shokri2015privacy}, and it has been shown that private image data can be recovered from the gradient information of both shallow and deep neural networks \cite{aono2017privacy,8241854,wang2019beyond,zhao2020idlg,Zhu2020}. Therefore, additional privacy protection techniques like homomorphic encryption (HE) \cite{cryptoeprint:2015:1192} and differential privacy (DP) \cite{dwork2008differential} have been suggested to address this issue. Phong \emph{et al.} \cite{8241854} adopt lattice based additive HE to protect uploaded local model parameters. Hao \emph{et al.} \cite{hao2019towards} propose a more efficient symmetric additive HE scheme to reduce the encryption time, while Zhang \emph{et al.} \cite{254465} use introduce an encoding quantization technique to combine the model parameters into a vector and encrypt the generated vector as a whole, further reducing the encryption time in horizontal FL. More recently, Zhu \emph{et al.} \cite{zhu2020distributed} design a distributed additive encryption scheme suited for horizontal FL, in which key pairs are collaboratively generated between the server and clients. On top of the distributed key generation, ternary gradient quantization \cite{wen2017terngrad} of the model parameters together with an approximate aggregation method is adopted to significantly save the encryption time, in particular for deep neural networks.

Compared to HE approaches, DP is more favorable in terms of computation time. For instance, local DP is used in \cite{shokri2015privacy,geyer2017differentially,zhao2020local,truex2020ldp,seif2020wireless,wei2020federated}, in which the model gradients on the client side are perturbed by adding Gaussian or Laplacian noise before uploading them to the server. Meanwhile, perturbation noise can also be added on the server side to the global model \cite{mcmahan2017learning,naseri2020toward}, known as central DP, to indistinguish the outputs of the aggregation function. Moreover, Truex \emph{et al.} propose a hybrid privacy-preserving FL scheme by combining Paillier encryption \cite{pascal1999public} with local DP. The disadvantage of using DP is that the perturbation noise will lead to performance degradation of the global model. Even worse, DP cannot deal with some inverting gradient attacks as introduced in \cite{8241854,hitaj2017deep}.

Besides HE and DP based protection methods, other techniques \cite{xu2019hybridalpha,li2019end,yang2021computationefficient} have also been used, and integrating  secure aggregation protocol with a double masking technique \cite{bonawitz2017practical} is a very popular approach. In this scheme, key pairs are generated by exchanging Diffle-Hellman keys \cite{diffie1976new} among connected clients and perturbed random numbers can be canceled out after model aggregation on the server. In addition, double masking is robust to the issue caused by offline clients, since the global model parameters can still be retrieved if some clients are offline during training. However, this method applies the Diffle-Hellman key exchange and Shamir secret sharing \cite{shamir1979share} to every client, which requires a large amount of communication resources.

Finally, horizontal FL often has worse global model performance than centralized learning, especially when parametric models are used in FL. One reason for this is that, horizontal FL needs to do weighted model averaging to update the global model, which has limited theoretical evidence to support the effectiveness of this approach. Although, data parallelism  \cite{seide20141,li2015malt,campos2017distributed,yadan2013multi,gupta2020fast,shallue2018measuring} in distributed machine learning \cite{dean2012large,mcdonald2010distributed,keuper2016distributed}, which has a very similar learning scheme with horizontal FL, has been empirically proved to be valid and widely used in multi-GPU computation to accelerate the learning speed in deep learning, designers need to ensure the training data on multiple devices are evenly distributed. In practice, however, the client training data in horizontal FL are always Non-IID, which may degrade the global model performance. 

\subsection{Vertical Federated Learning} \label{verticalfl}
\begin{figure}
\centering 
\includegraphics[width=0.92\textwidth]{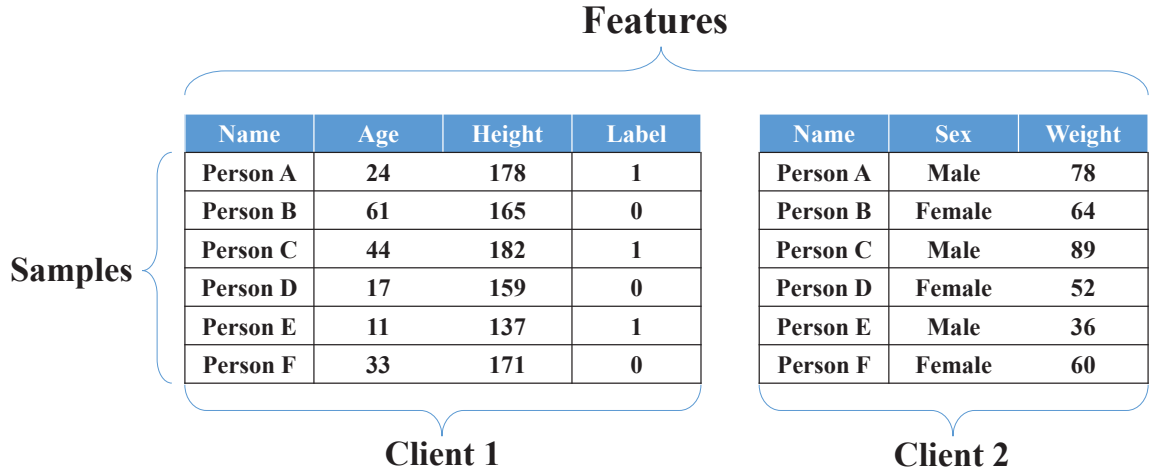} 
\caption{An illustrative example of vertical federated learning. Client 1 stores three features (name, age, and height) and a label of all six persons, while Client 2 stores two features (sex and weight) only without a label.} 
\label{vfl}
\end{figure}

Vertical FL \cite{yang2019federated,9076003} is also called heterogeneous FL \cite{yu2020heterogeneous}, in which users' training data share the same sample space but have different feature spaces. As shown in Fig. \ref{vfl}, Client 1 and Client 2 have the same data samples with different features. Unlike horizontal FL where all clients have their own local data labels, in vertical FL it is often assumed that only one client stores all the data labels (e.g. in Fig. \ref{vfl}, only Client 1 has the data labels and Client 2 has no labels). The client with data labels are called \emph{guest} party \cite{9153560} (client) or \emph{passive} party \cite{cheng2019secureboost} (client), whereas the client without data labels are termed \emph{host} party (client) or \emph{active} party (client). 

The main steps of vertical FL for training a logistic regression \cite{montgomery2021introduction} are shown in Algorithm \ref{vfllogistic}, where $X_{b}^{k}$ is the batch data on Client $k$, $z_{b}^{k}$ is the inner product of local batch data $X_{b}^{k}$ and local model $\theta_{t}^{k}$, $a$ is the activation function, $\hat{y}_{b}$ is the prediction for batch data and $L(y_{b},\hat{y}_{b})$ is the loss function. For each batch training in line 6 of Algorithm \ref{vfllogistic}, data sample identities of $X_{b}^{k}$ must be the same for all clients; otherwise, the prediction $\hat{y}$ cannot be correctly calculated. Different from FedAvg, a horizontal federated learning algorithm, a vertical federated logistic regression system does not contain a central server, nor a shared global model. Both the guest client and host clients have their own local models $\theta^{k}$ corresponding to the feature space of local data $X^{k}$, which should be kept locally without sending them to other clients. After calculating the model outputs (inner product) $z_{b}^{k}$ on all connected clients, these results will be sent to the guest client for aggregation to calculate the loss function (line 10 Algorithm \ref{vfllogistic}). Finally, the intermediate gradients $\frac{\partial L}{\partial z_{b}^{k}}$ can be calculated on the guest client and sent back to the corresponding host client for local model update. Note that the model on the guest client should also be updated. 

\begin{algorithm}[htbp]\footnotesize{
\caption{Vertical FL for logistic regression} 
\algblock{Begin}{End}
\label{vfllogistic}
\begin{algorithmic}[1]
\State \textbf{Client $1$} is \emph{guest} client and others are \emph{host} clients
\State Training data $X=\left \{ X^{1},X^{2},...X^{K} \right \}$
\State Initialize local model ${\theta_{0}^{k}}$, $k \in (1, K)$ \\
\For {each communication round $ t = 1,2,...T\ $}
\For{batch data $X_{b}^{k} \in (X_{1}^{k}, X_{2}^{k},...X_{B}^{k})$}
\For {each \textbf{Client} $k = 1,2,...K$ in parallel}
\State Compute $z_{b}^{k}=X_{b}^{k} \theta_{t}^{k}$ and send it to \textbf{Client $1$}, if $k \neq 1$
\EndFor
\State Compute $\hat{y_{b}}=a(\sum_{k=1}^{K}z_{b}^{k})$ and $L(y_{b},\hat{y}_{b})$ on \textbf{Client $1$}
\State Compute $\frac{\partial L}{\partial z_{b}}$ on \textbf{Client $1$} and send $\frac{\partial L}{\partial z_{b}^{k}}$ to \textbf{Client} $k$, $k \in (2, K)$
\For{each \textbf{Client} $k = 1,...K$ in parallel}
\State $\theta_{t}^{k} \leftarrow \theta_{t}^{k}-\eta \frac{\partial L}{\partial z_{b}^{k}}\frac{\partial z_{b}^{k}}{\partial \theta_{t}^{k}}$
\EndFor
\EndFor
\EndFor 
\end{algorithmic}}
\end{algorithm}

Apart from the client training data, there are three main differences between horizontal FL and vertical FL. First, horizontal FL has a server for global model aggregation, while vertical FL does not have this kind of central server, nor a global model. Consequently, the aggregation of local model outputs in vertical FL is performed on the guest client to construct the loss function. Second, model parameters or gradients are communicated between the server and clients in horizontal FL. In vertical FL, by contrast, local model parameters are related to the local data feature spaces that are not allowed to be transferred to other clients. Instead, the guest client receives model outputs from the connected host clients and sends the intermediate gradient values back for local model updates. Finally, the server and clients interact with each other for only once in one communication round in horizontal FL, while the guest client and host clients need to send and receive information for $B$ times in one communication round of vertical FL. 

Compared to horizontal FL, training \emph{parametric} models in vertical FL has two advantages. First, models trained in vertical FL always should be in principle have the same performance as the model trained in a centralized way. This is because the computed loss function in vertical FL (line 8 and line 10 of Algorithm \ref{vfl}) is the same as the one in centralized learning. Second, vertical FL often consumes less communication resources than horizontal FL, because only the batch local model outputs $z_{b}^{k}$ and intermediate gradients $\frac{\partial L}{\partial z_{b}^{k}}$ are required to be transmitted between the guest client and host clients, where $z_{b}^{k}$ is a batch set of scalars for logistic regression on client $k$, and $\frac{\partial L}{\partial z_{b}^{k}}$ is the intermediate gradients. Note that the local model $\theta_{b}^{k}$ can also be used to extract feature representations, and in this case $z_{b}^{k}$ becomes a set of vectors whose sizes are determined by the designer. Therefore, the communication costs in vertical FL are determined by the number of data samples and thus vertical FL may consume more communication resources than horizontal FL only if the data size is extremely large. 

Privacy preservation is one main challenge in vertical FL. Hardy \emph{et al.} \cite{hardy2017private} propose a privacy preserving identity resolution scheme for secure ID alignment in vertical logistic regression, and Nock \emph{et al.} \cite{nock2018entity} provide a detailed analysis of the impact of identity resolution. Liu \emph{et al.} introduce a protocol to protect user sample IDs in asymmetrical vertical FL. Yang \emph{et al.} introduce a simplified two-party vertical FL framework \cite{yang2019parallel} by removing the third party coordinator. Different from the above work in which parametric models are used for vertical FL, Cheng \emph{et al.} adopt xgboost \cite{chen2015xgboost} decision tree model to construct a secureboost \cite{cheng2019secureboost} system. Similarly, Wu \emph{et al.} propose a privacy preserving vertical decision tree learning scheme called Pivot \cite{wu2020privacy}, which does not need any trusted third party. 

In addition to privacy preservation, effort has also been made to enhance the learning performance in vertical FL. Yang \emph{et al.} \cite{yang2019quasi} adopt the Quasi-Newton method \cite{pfrommer1997relaxation} for vertical federated logistic regression to accelerate the convergence speed. Liu \emph{et al.} propose a FedBCD algorithm \cite{liu2019communication}, in which each party conducts multiple local updates before communication to reduce the required number of communication rounds. Feng and Yu propose an MMVFL framework \cite{feng2020multi} to solve multi-class problems with multiple parties. Finally, Chen \emph{et al.} \cite{DBLP:journals/corr/abs-2007-06081} use a perturbed local embedding technique to simultaneously protect individual's privacy and enhance the communication efficiency in asynchronous vertical FL.


\section{Categories of Non-IID Data}
The training data on each client in FL heavily depends on the usage of particular local devices, and therefore, the data distribution of connected clients may be totally different with each other. This phenomenon is known as Non-IID \cite{mcmahan2017communication}, which may cause severe model divergence, especially for \emph{parametric} models in horizontal FL. More specifically, for a supervised learning task on client $k$, assume each data sample $(\bm{x}, y)$, where $x$ is the input attribute or feature and $y$ is the label, follows a local distribution $P_{k}(\bm{x},y)$. By Non-IID, we mean $P_{k}$ differs from client to client.  Similar to \cite{kairouz2019advances,li2021federated}, we discuss categories of Non-IID data from the perspective of attribute $\bm{x}$, label $y$ and other more complicated FL scenarios.


\subsection{Attribute skew}
Attribute skew indicates the scenarios in which the feature distribution $P_{k}(\bm{x})$ across attributes on each client is different from each other. The data attributes across clients can be non-overlapped, overlapped or even the same.

\subsubsection{Non-overlapping attribute skew}
\begin{figure}[h]
\centering 
\includegraphics[width=0.46\textwidth]{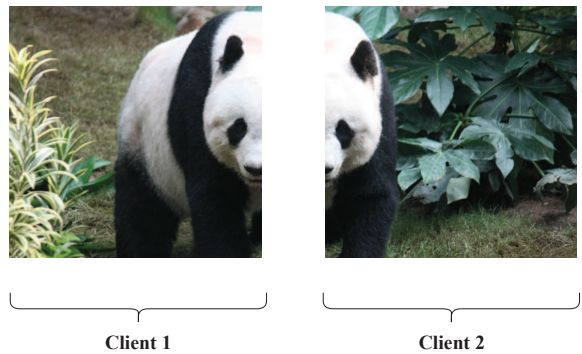} 
\caption{An example of non-overlapping attribute skew for image datasets, the left half of the image data is stored on Client $1$, while the right half of the image is on Client $2$.} 
\label{fig:vspanda}
\end{figure}

Non-overlapping attribute skew means that the data features across the clients are mutually exclusive. In this case, if data samples $\bm{x}$ on different clients $k$ with the same identities hold the same labels $P_{k}(y|\bm{x})$, it is known as vertical FL.
This non-overlapping property can ensure that the computed total loss of the logistic regression  (linear model) in vertical FL is equal to that in centralized learning (refer to line 10 of Algorithm \ref{vfllogistic}). For a dataset like personal information as shown in Fig. \ref{vfl},  client 1 owns the features of age and height, while client 2 has the features of sex and weight. And for image data as shown in Fig. \ref{fig:vspanda}, an image of panda is partitioned into two non-overlapped pieces and client 1 stores the left part and client 2 stores the right part. The main difference between these two types of datasets is that the adjacent attributes for personal information may not correlated (for example, 'Age' and 'Height' in client 1 of Fig. \ref{vfl} have no relationship), but adjacent pixels for image data are always strongly correlated.

\subsubsection{Partial overlapping attribute skew}
\begin{figure}
     \centering
     \begin{subfigure}[b]{0.44\textwidth}
         \centering
         \includegraphics[width=\textwidth]{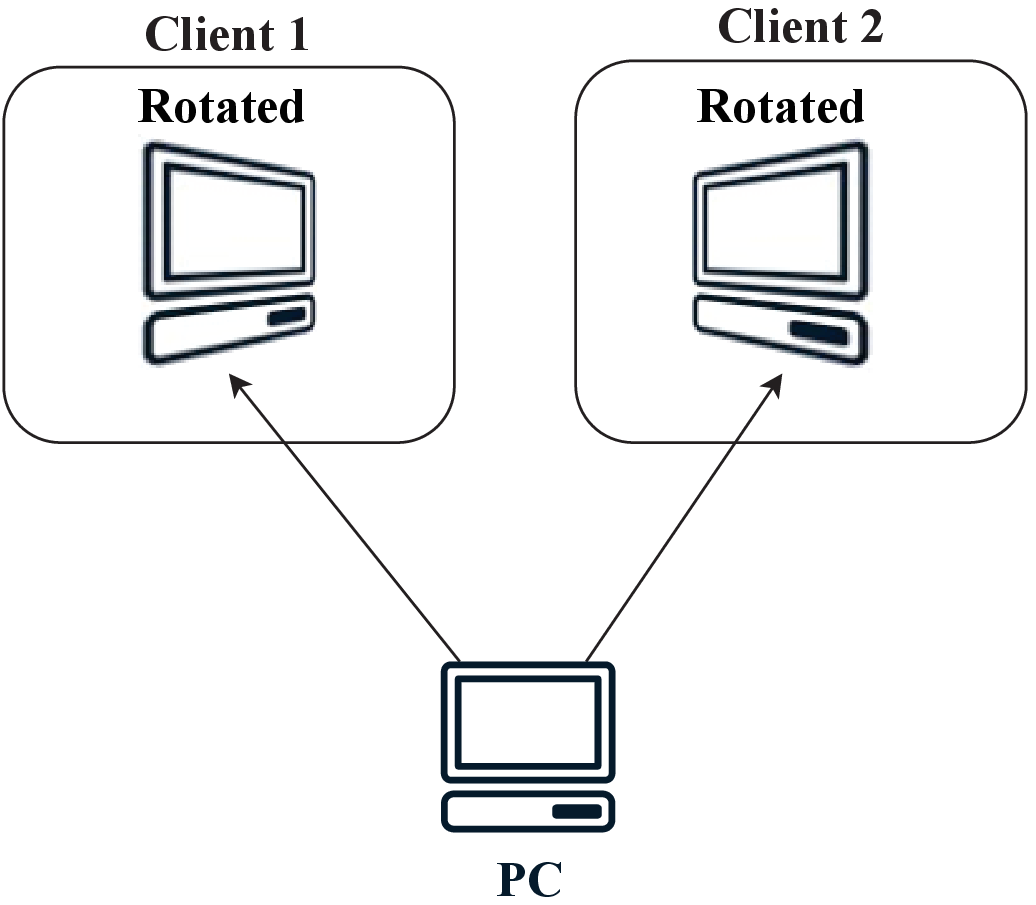}
         \caption{Multi-view of a PC image.}
         \label{fig:difangle}
     \end{subfigure}
     \hfill
     \begin{subfigure}[b]{0.46\textwidth}
         \centering
         \includegraphics[width=\textwidth]{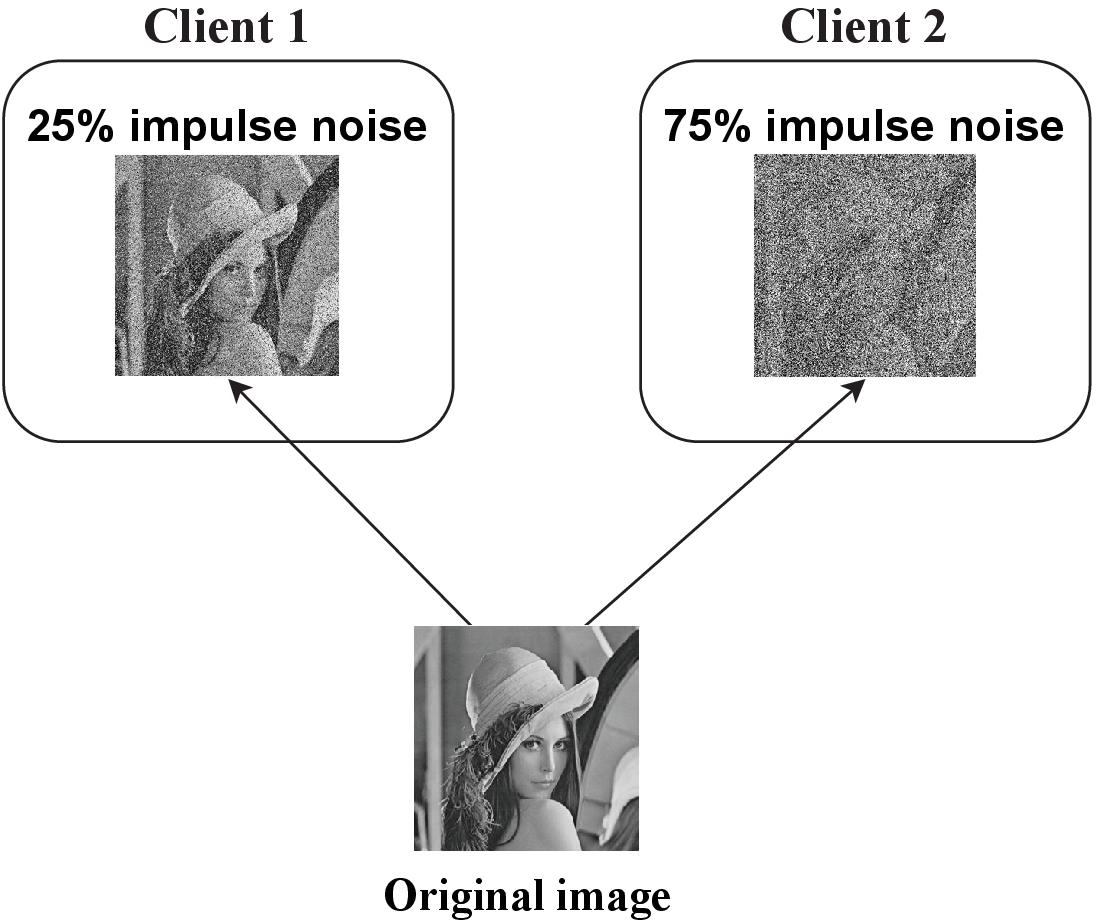}
         \caption{Images perturbed by different levels of impulse noise.}
         \label{fig:difnoise}
     \end{subfigure}
\caption{Two examples of attribute distribution skew.}
\label{fig:three graphs}
\end{figure}

The second category of attribute skew is partial overlapping attribute skew, in which some parts of the data features can be shared with each other. For instance, multi-view images \cite{su2015multi,li2021federated} are shot from different angles and each party holds single-view (single angle) images. As shown in Fig. \ref{fig:difangle}, client $1$ and client $2$ stores the same PC image with different angles. Note that the distribution of each overlapping attribute among different clients may be consistent or inconsistent.

\paragraph{Consistent distribution:}
In this case, the $i$-th overlapping attribute $x_i$ is sampled from the same distribution $P(x_i,y)$, which means that the shared attributes will not enlarge the Non-IID divergence of the data.
 
\paragraph{Inconsistent distribution:}
As the name suggests, inconsistent distributions of the shared attributes are another reason of Non-IID divergence in addition to the situation in which each client has unique attributes. For example, Xu \textit{et al}. assume that each client may encounter an infeasible domain when sampling data from the domain of input \cite{xu2021federated}.

\subsubsection{Full overlapping attribute skew}

This case is often known as horizontal FL (horizontal data partition), when the overall data is horizontally \textit{partitioned} among the clients without overlap, as shown in Fig. \ref{hfl}.

Generally, Non-IID divergence of this case usually happens in the presence of inconsistent data distributions, when there is an attribute imbalance of the training data across clients due to perturbations, for example, different degrees of impulse noise across clients, as shown in Fig. \ref{fig:difnoise}. In addition, real world feature imbalance is another feature distribution skew with the same data features. For instance, the EMNIST dataset \cite{cohen2017emnist} collects hand written digit numbers from different people, thus, even for the same digit number, the character features (e.g., stroke width and slant) are different.

\subsection{Label skew}
Label skew represents the scenarios in which the label distribution differ from client to client. There are two slightly different situations of label skew, one is label distribution skew and the other is label preference skew.

\subsubsection{Label distribution skew}
\label{labeldistskew}
Label distribution skew is a common Non-IID category, where label distributions $P_{k}(y)$ on the clients are different and a conditional feature distribution $P_{k}(x|y)$ is shared across the clients. And it is usually caused by location variations of the clients that store similar types of local training data. 
Two main kinds of label distribution skew settings \cite{li2021federated} have been considered in FL research, namely label size imbalance and label distribution imbalance.

\begin{figure}[h]
\centering 
\includegraphics[width=0.5\textwidth]{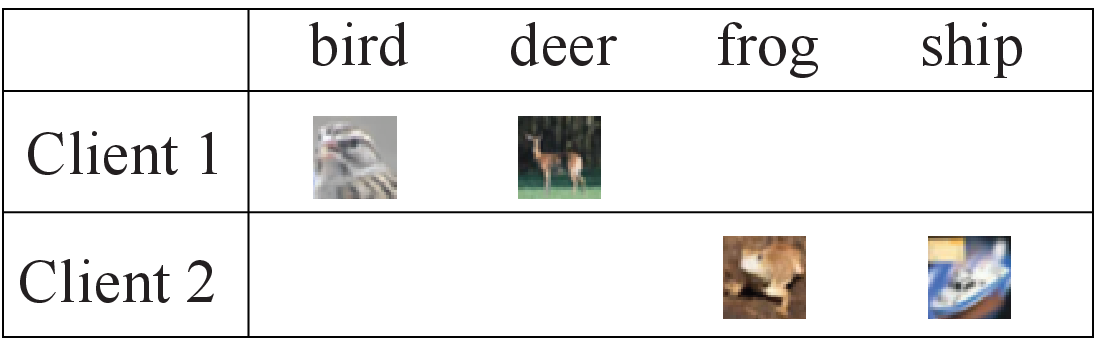} 
\caption{An illustrative example of label size imbalance with two clients and $c=2$ using CIFAR10 as an example.} 
\label{fig:label_noniid}
\end{figure}

As shown in Fig. \ref{fig:label_noniid}, label size imbalance is originally proposed in the FedAvg algorithm \cite{mcmahan2017communication}, where each client owns data samples with a fixed number $c$ label classes. $c$ is a hyper parameter that determines the degree of label imbalance, and a smaller $c$ means stronger label imbalance, and vice versa.

Label distribution imbalance is another situation considered in \cite{yurochkin2019bayesian}, in which a $\bm{p}_{c,k}$ portion of the instances of label class $c$ is distributed on client $k$ with a probability $p_{c}\sim Dir_{k}(\beta)$, where $Dir(.)$ is the Dirichlet distribution and $\beta$ is the concentration parameter influencing the imbalance level. And a larger $\beta$ value will result in more unbalanced data partition. Since the imbalance level of data distributions can be easily adjusted by $\beta$, some recent studies \cite{li2020model,lin2020ensemble,Wang2020Federated,NEURIPS2020_564127c0} adopt this method to emulate different Non-IID cases in the real-world.

\subsubsection{Label preference skew}

Different from label distribution skew, label preference skew considers the client data sample intersection issues often encountered in real-world applications, where the conditional distribution $P_{k}(y|x)$ may vary across the clients, although $P_{k}(x)$ is the same. 

\begin{figure}[h]
\centering 
\includegraphics[width=0.46\textwidth]{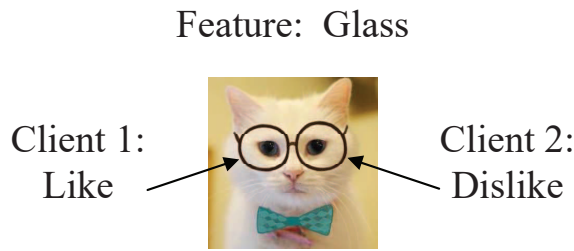} 
\caption{An example of label preference skew. For the same input feature (a cat with glass), Client 1 may label 'like' while Client 2 may label 'dislike' instead.} 
\label{fig:same_feature_noniid}
\end{figure}

If the training data across the clients are horizontally overlapped, it is likely that different users annotate different labels for the same data sample due to individuals' preferences. For the image of a cat with glass in Fig. \ref{fig:same_feature_noniid}, user $1$ labels 'like', while user $2$ labels 'dislike'.

Crowdsourcing data \cite{garcia2016challenges} is a more complex situation where data labels can be noisy, posing serious challenges to information collection in many centralized machine learning tasks, let alone in FL tasks. For example, most local devices only contain unlabelled data and require multiple workers or volunteers to label them. Therefore, it is not uncommon that some labels are incorrect, noisy and even missing.

\subsection{Temporal skew}

Temporal skew means the skew in distribution in temporal data, including spatio-temporal data and time-series data, also referred to as time-stamped data. This type of data accounts for a large proportion of real-world FL applications (e.g., various measurement data in IoT devices) and hence, it needs to be categorized separately. Different from attribute skew and label skew, temporal skew refers to the inner correlation of data observations in the time domain. Particularly for time series data, the data distributions $P_{k}(\bm{x}, y|t)$ ($t$ is the time index) on each client keeps changing  over time.

\begin{figure}[h]
\centering 
\includegraphics[width=0.46\textwidth]{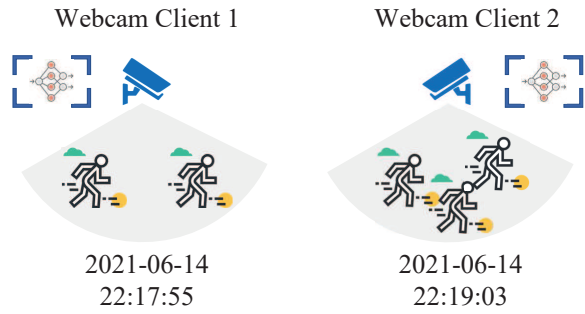} 
\caption{An example of temporal skew for webcam recordings in FL.}
\label{fig:time_series}
\end{figure}

We present an illustrative example of temporal skew, as shown in Fig. \ref{fig:time_series}, where two webcams at different locations take photos of runners over over the time and the deviation of photos on different clients comes from time difference. Although there is a temporal skew in the data collected on Client 1 and Client 2, these data have a great amount of intersections across the entire time period.



\begin{figure}[h]
\centering 
\includegraphics[width=0.64\textwidth]{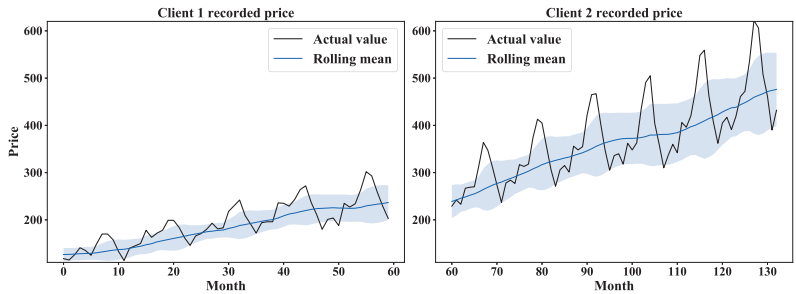} 
\caption{An example of temporal skew in distributions of stock data in FL.} 
\label{fig:stock_series}
\end{figure}

In addition, it can happen that different clients collect data during different periods of time rather than the whole time period, which also results in temporal skew. As shown in Fig. \ref{fig:stock_series}, Client 1 stores the stock prices for the first 60 months, while Client 2 holds those for the last 60 months.

\subsection{Other scenarios}
There are some other scenarios that do not belong to any Non-IID categories discussed above.


\subsubsection{Attribute \& Label skew}
In the scenario of Attribute \& Label skew, different clients hold data with different labels and different features, which integrates the characteristics of both horizontal and vertical FL.

\begin{figure}[h]
\centering 
\includegraphics[width=0.5\textwidth]{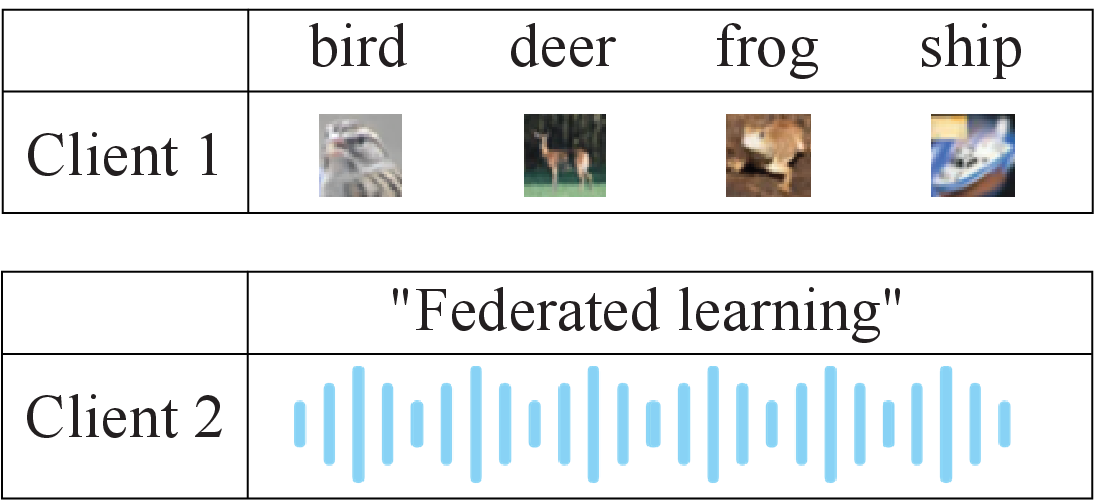} 
\caption{An example of different features, different labels Non-IID. Client $1$ and Client $2$ may hold different types of data (image and audio).} 
\label{fig:dfdl_noniid}
\end{figure}

Specifically, different features can also be extended to a generalized definition that data types may vary across clients. As shown in Fig. \ref{fig:dfdl_noniid}, client $1$ only have image dataset and client $2$ only have speech dataset. Global model aggregation becomes extremely hard in this case, since local model types or structures may be totally different across clients.

\subsubsection{Quantity skew}
Quantity skew means the number of training data varies across different clients and it can occur in all situations discussed above.

\section{Challenges of Non-IID Data to Model Training}
\label{model}
Both \emph{parametric} models and \emph{non-parametric} models have been used for FL. Due to the differences in the training mechanisms for parametric and non-parametric models in both horizontal and vertical FL, the impacts of Non-IID  on their training performances are also considerably different. In this section, we introduce the core training steps of both parametric and non-parametric models in FL at first, and then discuss the influences of Non-IID training data on their performance in horizontal and vertical FL, respectively.

\subsection{Horizontal federated learning}
If not specified, Non-IID in horizontal FL usually refers to label distribution skew, this is because, label distribution skew often causes more serious client data distribution divergence compared to the case with the same features but different labels. For those extreme cases in which for example, each client contains data samples with only one class \cite{yu2020federated}, it can happen that the global model does not converge at all.

\subsubsection{Parametric models}
\begin{figure}
\centering 
\includegraphics[width=0.75\textwidth]{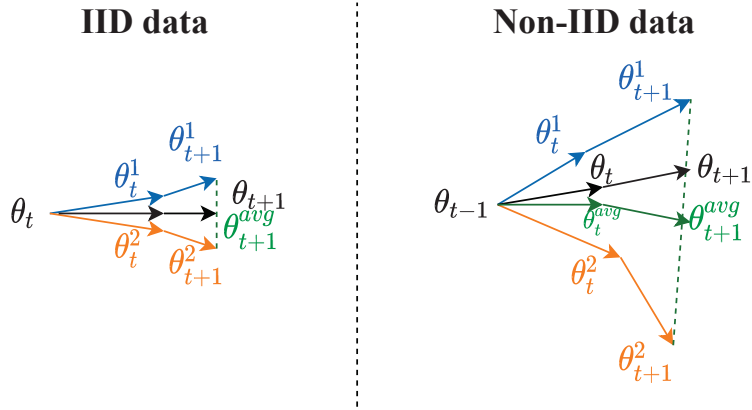} 
\caption{An illusion of parametric model divergence in horizontal FL for both IID and Non-IID data, where $\theta_{t}$ is the global model and $\theta_{t}^{avg}$ is the averaged model of the local model of Client 1 ($\theta_{t}^{1}$) and that of Client 2 ($\theta_{t}^{2}$).} 
\label{fig:modeldiver}
\end{figure}

Non-IID data does affect the learning performance on \emph{parametric} models \cite{mcmahan2017communication,zhao2018federated,sattler2019robust,wang2020optimizing,shoham2019overcoming,briggs2020federated} and will always cause global model divergence in horizontal FL. Since the local data distributions are different from the global data distribution, the averaged local model parameters may also be far away from the global model parameters \cite{zhao2018federated}, particularly when the number of epochs for local updates is large \cite{karimireddy2020scaffold,DBLP:journals/corr/abs-1812-06127,Wang2020Federated}. As shown in Fig. \ref{fig:modeldiver}, the divergence between the averaged local model parameters $\theta_{t+1}^{avg}$ and the actual global model parameters $\theta_{t+1}$ are much larger for Non-IID data. In addition, the divergence may accumulate over the communication round $t$.

\paragraph{Linear models} like linear regression and ridge regression are the simplest \emph{parametric} models, which can be directly used in FedAvg algorithm. However, the local private data are more likely to be leaked or reverse-engineered \cite{wei2020framework,lyu2020threats} for linear models than for other complex parametric models. This is because the communicated gradients of the parameters of linear models are proportional to the local input data \cite{8241854}. Therefore, additional privacy preserving methods like homomorphic encryption (HE) are usually required to use linear models in horizontal FL.

\paragraph{Neural networks} are currently the most widely used \emph{parametric} models in FL, since they have extraordinary performance in computer vision, speech recognition, among others. Three types of neural network models, multi-layer perceptrons \cite{ruck1990feature,gardner1998artificial}, convolutional neural networks \cite{lecun1995convolutional} and long-short-term-memory (LSTM) \cite{hochreiter1997long}, are originally used in the FedAvg algorithm for image classification and next word prediction tasks. 
Simulation results \cite{mcmahan2017communication} indicate that FedAvg is able to achieve good performance using \emph{shallow} neural network models on relatively simple datasets, although it performs slightly worse than centralized learning. However, in case \emph{deep} neural networks are used, the FedAvg algorithm becomes very sensitive to client data distribution and may fail to converge on very strongly Non-IID data.

\subsubsection{Non-parametric models}

\begin{figure}
\centering 
\includegraphics[width=7cm]{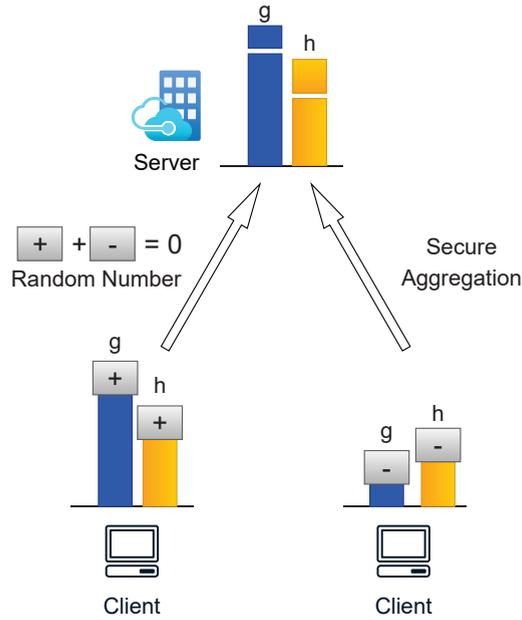} 
\caption{Secure aggregation used in horizontal federated xgboost, where g is the gradient and h is the Hessian. Secure aggregation \cite{bonawitz2017practical} is used here for information protection.}
\label{saxgboost}
\end{figure}

\paragraph{Decision trees}
are typical \emph{non-parametric} models and the gradient boosting decision tree (GBDT) \cite{ke2017lightgbm} becomes very popular at present in solving both regression and classification problems because of its good performance. Among gradient boosting tree models, xgboost \cite{chen2015xgboost} is the most powerful one, which has already been applied in both horizontal FL \cite{yang2019tradeoff} and vertical FL \cite{fate,cheng2019secureboost}.
The following modifications should be made when the FedAvg algorithm is applied to xgboost:
\begin{enumerate}
\item Set binning points (thresholds) for all data features on all connected clients and the server.
\item Each client computes the local histograms for all binning points and send all histograms to the server.
\item The server sums up the received local histograms, and calculates the corresponding impurity values for all binning points. Then, the server will find the best split binning points and send them back to the clients.
\item After getting the best split binning points, clients can split the current node and construct the next-layer of the tree. If stop conditions are not fulfilled, go back to Step 2 and repeat this procedure.
\end{enumerate}
The local histograms are the sum of the local gradients and Hessian matrix for a specific binning point and they intrinsically contain label information of the training data \cite{cheng2019secureboost}, which can be inferred by the server. In order to protect the privacy of the information about the local models, one can often adopt a secure aggregation technique that adds or subtracts an extra random number to the local histograms of each client before sending them to the server. Since the random number is generated by a pseudo random generator with the same seed value, it can be cancelled out with each other \cite{bonawitz2017practical} after summing the perturbed histograms in the server, as shown in Fig. \ref{saxgboost}.

Although different horizontally 'partitioned' local training data may lead to different local histograms of every possible split, the summed gradients and Hessian (blue and orange bar in the server of Fig. \ref{saxgboost}) in the server do not change for each split. Therefore, Non-IID data does not cause any training divergence for decision tree models.

\subsection{Vertical federated learning}
Non-IID in vertical FL usually refers to `same labels, different features', and most existing work assume that different clients do not have no overlap in features (i.e., no common features) and the data labels are only stored on one client, called the guest client. The rest clients that do not have any labels are known as host clients.


\subsubsection{Parametric models}

\paragraph{Linear models}
like logistic regression (generalized linear model) are commonly used in vertical FL. As discussed in Section \ref{verticalfl} that both model gradients and the training data are stored on local devices during the training period, which reduces both potential data leakage risk and communication costs. To further enhance the security level, some research work \cite{hardy2017private,9076003,yang2019parallel} adopts an encrypted model training scheme by encrypting all the model outputs from the host clients. In this case, the gradients of the loss with respect to model parameters can be calculated on ciphertext directly \cite{yang2019federated} due to the property of HE.

As discussed in Section \ref{verticalfl}, the loss function of training logistic regression in vertical FL has no difference to that in centralized learning. Therefore, Non-IID data does not affect the learning performance for linear models.

\begin{figure}
\centering 
\includegraphics[width=0.65\textwidth]{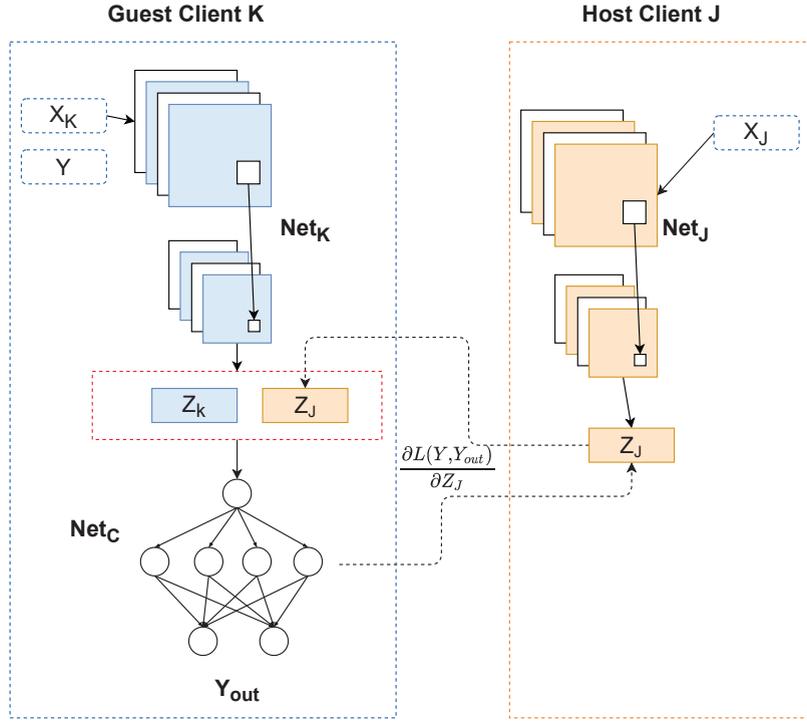} 
\caption{A simple example for a two-part vertical federated neural network scheme. The local model output $Z_{J}$ should be sent to the guest client and the merged output $Z$ will be the input of the model $Net_{C}$} 
\label{vfldeeplearning}
\end{figure}

\paragraph{Neural networks}
can also be trained in vertical FL. As shown in Fig. \ref{vfldeeplearning}, both the guest client $K$ and a host client $J$ own a local neural network model $Net_{K}$ and $Net_{J}$, respectively, which are used to extract features of the local training data. And then the feature representation $Z_{J}$ on the host client will be sent to the guest client to concatenate with $Z_{k}$ over the feature dimension. Finally the final output $Y_{out}$ can be generated by passing forward the merged feature output through an neural network model $Net_{C}$. There are two main differences between the  vertical federated logistic regression and neural network model. First, the local model output $z$ of a single data sample for logistic regression is a scalar, which should be summed up for loss computation. By contrast, the local output $Z$ in the neural network is a vector of feature representation. Second, the guest client needs to construct an extra neural network model to achieve the classification predictions.

\begin{figure}
\centering 
\includegraphics[width=0.65\textwidth]{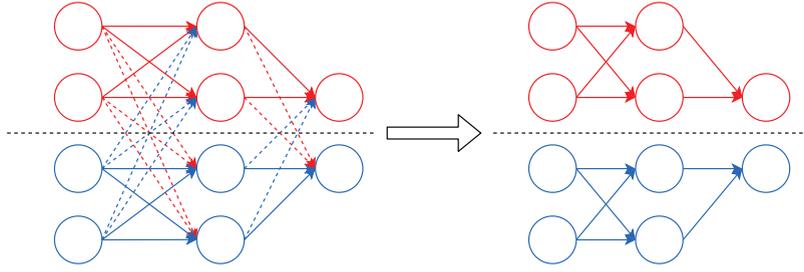} 
\caption{A simple example for a vertically partition neuron network. The dashed lines denote the coupling connections.} 
\label{fig:verticaldividenode}
\end{figure}

It should be noticed that, overall output of the neural networks in vertical FL is different from that in centralized learning, because the neural network is split into several separated sub-networks. A simple example is shown in Fig. \ref{fig:verticaldividenode}, where the  connections of the global network (denoted by dashed lines) will vanish after splitting it into two sub-networks. In fact, training neural networks in vertical FL remains an open challenge and there are very limited research work focusing on this topic. Recently, Liang \emph{et al.} present a self-supervised vertical federated neural architecture search to collaboratively fine-tuning both the architecture and model parameters \cite{liang2021selfsupervised}. However, they do not discuss the impact of Non-IID data on the model learning performance.

\subsubsection{Non-Parametric models}

For decision trees in vertical FL, all the histogram values are calculated only on the guest client, because it owns all labels of the training data, as defined in Section \ref{verticalfl}. To avoid leaking data label information of the guest client, all the gradients and Hessian values of node data samples need to be encrypted before sending them to any other host clients. The training steps \cite{cheng2019secureboost} are as follows:
\begin{enumerate}
\item Set binning points of all features for both the guest client and host clients. Initialize key pairs on the guest client.
\item On the current decision node or root node, compute and get the largest impurity value and send the encrypted gradients and Hessian values of node data samples to host clients.
\item For each host client, split data samples for all binning points and sum the corresponding encrypted gradients and Hessian values. Each split is recorded with a unique split number corresponding to the feature and split bin value. The summed values together with the split numbers are sent back to the guest client for decryption.
\item Calculate impurity values for all splits and get the best split with the largest impurity value on the guest client. If the best split is on the guest client, add the feature and split bin value with a unique record ID in the local lookup table. If the best split is on one of host clients, the guest client sends the split number to the corresponding host client. And then this host client can find the feature and split bin value based on the received split number and update the local lookup table.
\item The current decision tree can be constructed with the record ID and client ID. If the stop criterion is not fulfilled, go back to step 2 to construct the next layer of the tree.
\end{enumerate}

\begin{figure}
\centering 
\includegraphics[width=0.75\textwidth]{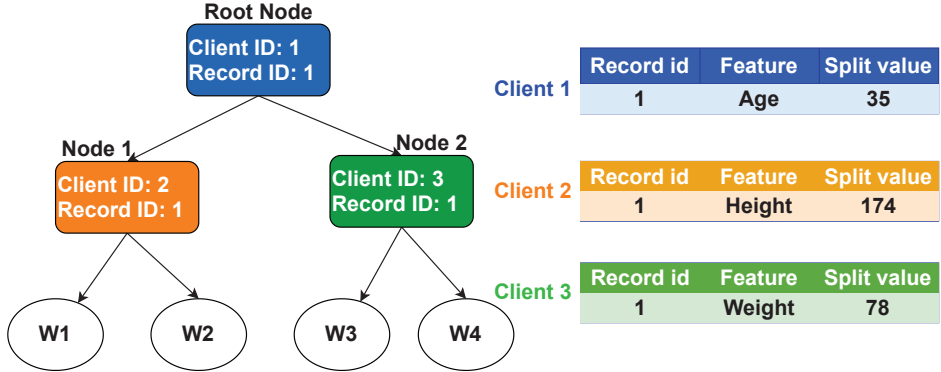} 
\caption{A simple decision tree built in vertical federated xgboost, where all the decision nodes contain the client ID and record ID only. Both the features and split bin values can be retrieved from the locally stored lookup table. W is the leaf node output.} 
\label{vflxgboost}
\end{figure}

A single generated tree model and its corresponding lookup table is shown in Fig. \ref{vflxgboost}. Same as in the aforementioned horizontal scenarios, the learning performance of decision tree models in vertical FL is not influenced by Non-IID data. This is because the calculated impurity values are only dependent on the gradients and Hessian of data samples on the decision node, which has no relationship to data feature distributions among clients.   

\section{Main Approaches to Handling Non-IID Data}
\begin{figure}
\centering 
\includegraphics[width=1\textwidth]{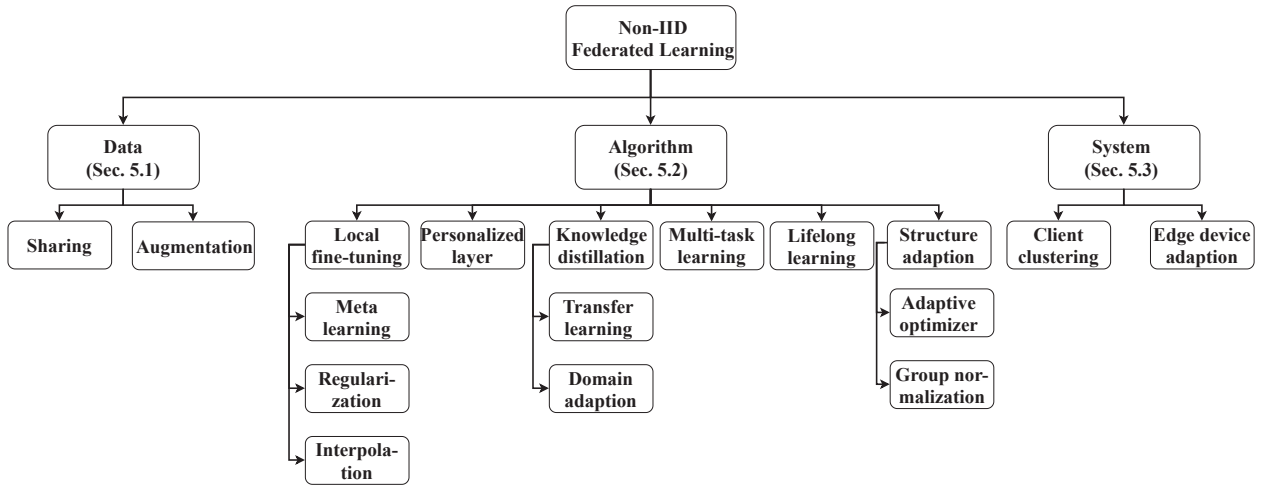} 
\caption{A summary of existing approaches to addressing Non-IID data.} 
\label{fig:noniid_FL}
\end{figure}

As discussed in the last section, Non-IID data, specifically label distribution skew, may cause severe learning divergence to \emph{parametric} models mainly in horizontal FL.
However, the FedAvg algorithm \textit{per se} cannot deal with model divergence problems caused by Non-IID data, especially when complex models such as neural networks are used in federated learning. Current approaches to dealing with Non-IID problems in horizontal FL can be classified into the data based approach, algorithm based approach, and system based approach, as shown in Fig. \ref{fig:noniid_FL}. Both advantages and disadvantages of these methods will be discussed in detail. In this section, if not specified otherwise, FL means horizontal FL, the baseline FL algorithm is FedAvg described in Algorithm \ref{hfl}, and models are neural networks.


\subsection{Data based approach}
Intuitively, the performance degeneration of FL in the presence of Non-IID data is caused by the heterogeneous data distributions and hence, data based approaches aim to address this issue by modifying the distributions. Data sharing and augmentation are two main solutions with state-of-the-art performance.
\label{sec:data}

\subsubsection{Data sharing}
\label{sec:sharing}
Data sharing \cite{zhao2018federated} is very straightforward but effective to deal with Non-IID data in horizontal FL. A globally shared dataset \textit{G} with a uniform distribution is stored on the server and the global model is warmed up by training \textit{G}. In addition, a random $\alpha$ percentage of \textit{G} should be downloaded to all connected clients so that the client model is updated by both local training data and the shared global data from \textit{G}. Experimental results show that the model test accuracy can be enhanced by approximately 30\% on the CIFAR10 \cite{krizhevsky2009learning} dataset with only 5\% of globally shared data.

Similar ideas are also used in \cite{tuor2020overcoming,DBLP:journals/corr/abs-1905-07210} to alleviate the negative effect of Non-IID by sharing some local data with the server.
Although this data sharing method can significantly enhance the global model performance on Non-IID data (in fact, this approach outperforms most horizontal FL algorithms on Non-IID data), it has very obvious shortcomings. At first, it is hard to get so called uniformly distributed global dataset, since the server has no idea about the data distributions among the connected clients. Second, downloading parts of global dataset to each client for model training violates the requirement of privacy preserving learning, which is the fundamental motivation of FL.

\subsubsection{Data augmentation}
\label{sec:augmentation}
Data augmentation \cite{tanner1987calculation} is originally a technique to increase the diversity of training data by some random transformations or knowledge transfer, which can also be used to mitigate local data imbalance issues in FL. There are three main kinds of data augmentation methods used in FL: the vanilla method, the mixup method \cite{zhang2018mixup}, and the generative adversarial network (GAN) \cite{NIPS2014_5ca3e9b1} based method.

The basic idea of using the vanilla data augmentation in FL is proposed in \cite{duan2019astraea}, in which each client needs to send its label distribution information like the number of data samples for each class $c$ to the server. Then the server can calculate the number of samples $C_{c}$ for each data class $c$ and their mean value $\bar{C}$. If $C_{c} < \bar{C}$, client $k$ needs to generate $(\bar{C}/C_{y})^{\alpha}$ number of augmentations for each local data sample $(x,y)$, where $\alpha$ is the hyperparameter to control the degree of augmentation and label $y$ is equal to $c$. Both the original local data sample and the augmented data are used to update the local model parameters.

The mixup method is another data augmentation approach to tackling Non-IID data. Shin \emph{et al.} \cite{Shin2020XORMP} first use this method to propose a XorMixFL framework. The core idea is that each client uploads its encoded seed samples (encoded using the XOR operator) to the server for decoding and the base data samples together with decoded samples in the server can construct a new balanced dataset. After that, a global model is trained on this reconstructed data and downloaded to each client until the training converges. Yoon \emph{et al.} propose a mean augmented method \cite{yoon2021fedmix} by exchanging the averaged batch local data with the server. The exchanged mean data will be combined and sent back to each client to reduce the degree of local data imbalance.

Different from the two previous methods, the purpose of federated GAN data augmentation is to train a good generator in the presence of Non-IID data. A general approach is that each client has to send its local seed data samples to the server at first, as does in the data sharing strategy. And then, the server can train the generator and discriminator of GAN based on these seed samples and the well trained generator will be sent to all connected clients. With the received generative model, each client can replenish the training data for the missing labels to construct an IID local dataset. 
However, sending local data samples to the server violates the data privacy requirement in FL and Yonetani \emph{et al.} suggest to locally train the discriminators, which will be weighted averaged to construct an aggregated global discriminator on the server, where the  weights are determined by the softmax function value calculated across the outputs of discriminators uploaded from connected clients. The generator can be updated by applying the loss gradients computed by the global discriminator. This approach can the protect local data privacy to some extent, but the information of label distribution must be revealed to the server. In order to address this issue, Jeong \emph{et al.} introduce a multi-hop federated augmentation with sample compression (MultFAug) strategy, which can protect not only data privacy but also the information about label distributions. 

Overall, data augmentation techniques can significantly improve the learning performance of the model trained on Non-IID data by replenishing the local imbalanced data with augmentations. However, most of these techniques can only be implemented with the help of aforementioned data sharing, which may increase the risk of data privacy leakage.

\subsection{Algorithm based approach}
\label{sec:algorithm}

As we stated in Section \ref{fl}, the goal of federated learning is to collaboratively train a shared model without sharing private data. However, the local models trained by FL may be harmed by the shared model in terms of performance \cite{yu2020salvaging}, and may fail to generalize due to the drift of heterogeneous data shards \cite{dinh2020personalized}. This can be seen from Eq. \eqref{eq:local loss}, which indicates that a certain client trains a local model without information exchange between clients, and its model may generalize poorly on unseen data. In addition, the conventional global loss function of a FL system is to minimize Eq. (\ref{eq:global_loss}), the output of the system is common for all clients and hence, each client may lose their precision on its own task, especially for heterogeneous data or objectives \cite{fallah2020personalized}. For example, in a FL system consisting of two clients \emph{A} and \emph{B}, client \emph{A} needs an inference efficiency model and client \emph{B} emphasizes the accuracy. Eventually, it may be difficult to apply the global model to client \emph{A} due to limited computing budget. 

\begin{figure}
\centering 
\includegraphics[width=0.75\textwidth]{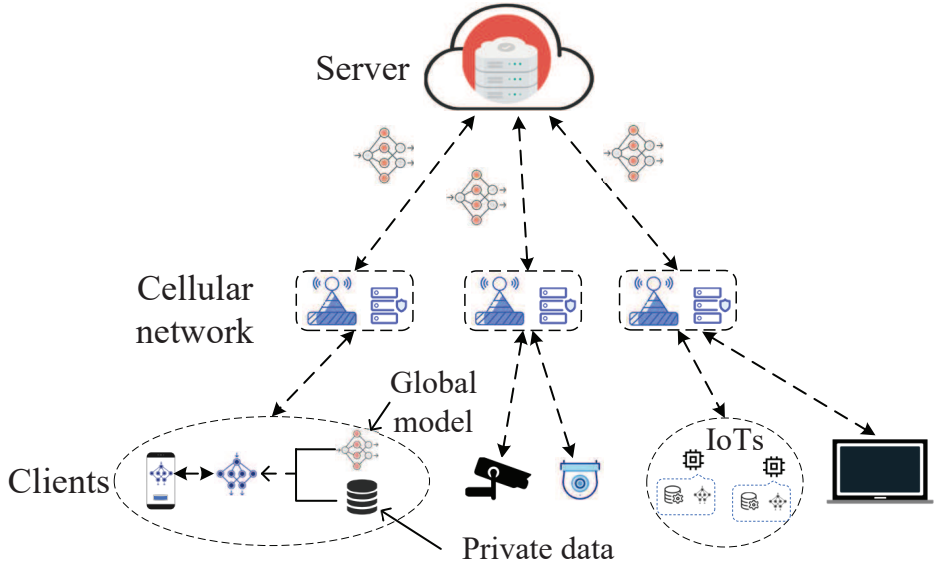} 
\caption{An illustration of personalization federated learning.} 
\label{fig:personalization_FL}
\end{figure}

To address these issues, personalization approaches have received much attention recently. Personalization, as shown in Fig. \ref{fig:personalization_FL}, aims to adjust the model according to the local tasks. In general, there are several major types of personalization methods, including conducting local fine-tuning (personalization via regularization \& interpolation, or meta learning), including a personalization layer, multi-task learning, and knowledge distillation \cite{deng2020adaptive}.

\subsubsection{Local fine-tuning}
Local fine-tuning, as the most classic and powerful personalization method, aims to fine-tune the local models after receiving the global model from the server using local data \cite{wang2019federated}, and FedAvg is the basic form of local fine-tuning. The purposes of local fine-tuning are two-folds, i.e., finding a suitable initial shared model, and combining local and global information.

One common idea for fine-tuning is to build a high quality initial global model based on meta-learning methods. A representative method called Personalized FedAvg (Per-FedAvg) \cite{fallah2020personalized}, which leverages Model-Agnostic Meta-Learning (MAML) \cite{finn2017model} to find an initial global model, making it easier for the local clients to obtain good performance with little computation costs. In Per-FedAvg, the authors modify the loss function in Eq. (\ref{eq:global_loss}) as follows:
\begin{equation}
\label{eq:perfedavg_loss}
\min_w F(\theta) =  \sum_{k=1}^{C \times K}\frac{n_{k}}{n} f_k \left(\theta-\alpha\nabla{f_k(\theta)} \right),
\end{equation}
where $\alpha$ is the step size, and the cost function $F$ is the average of meta functions $F_1$, $F_2$,..., $F_k$,..., $F_N$ on client $k$, which can be denoted by
\begin{equation}
\label{eq:meta_loss}
F_k(\theta) =  f_k \left(\theta-\alpha\nabla{f_k(\theta)}\right),
\end{equation}
where the local model $\theta^k$ is synchronized with $\theta$ and the gradient of local functions can be calculated by
\begin{equation}
\label{eq:meat_gradient}
\nabla{F_k(\theta)} = \left(I -\alpha^2\nabla{f_k(\theta)}\right)\nabla{f_k \left(\theta-\alpha\nabla{f_k(\theta)} \right)},
\end{equation}
where the first and second order information can be replaced by the unbiased estimation using a batch of data. As the authors discussed, the new cost function (Eq. (\ref{eq:meta_loss})) can capture the differences between clients, and a new client can perform well after being slightly trained based on the obtained initial solution with their own data. This method achieves the first-order optimality with convergence guarantees and better performance on heterogeneous data than FedAvg, but the approximate gradient of Per-FedAvg will significantly affect the results. Similarly, Jiang \emph{et al}. combine the Reptile algorithm \cite{nichol2018first} with FedAvg for local personalization \cite{jiang2019improving}, however, the performance of their method is only verified on EMNIST-62 and MNIST datasets. Chen \textit{et al}. propose federated meta-learning (FedMeta) framework, where a meta-learner is shared among the clients instead of a global model \cite{chen2018federated}. Similar to multi-task learning, FedMeta also treats each client as a separated task, and the target is to train a well-initialized model that can be rapidly adapted to any new tasks, but the shared meta-learner may also leads to the leakage of users' privacy.

Combination of local and global information is another approach and several studies on regularization and interpolation have been conducted. The aim of regularization is to minimize the disparity between the global and local models, and FedAvg can be viewed as a special case of regularization-based personalization. Instead of solving Eq. (\ref{eq:global_loss}) for an explicit global model, Hanzely \emph{et al}. design a new form of the cost function by adding a regularization term to investigate a trade-off between local and global models \cite{hanzely2020federated}. Dinh \emph{et al}. introduce Moreau Envelopes \cite{moreau1963proprietes} into the proposed pFedMe algorithm to overcome the statistical diversity among clients \cite{dinh2020personalized}. Specifically, they add a term of $l_2$ norm for client cost function:
\begin{equation}
\label{eq:pFedMe}
f_k(\theta_k)=\frac{1}{n_k}\sum_i^{n_k}l(\bm{x}_i,y_i;\theta_k) + \frac{\gamma}{2}||\theta_k-\theta||^2,
\end{equation}
where $\theta$ is the global model and $\gamma$ is the regularization parameter. In general, pFedMe outperforms FedAvg on the convergence rate, but there are too many hyperparameters need to be adjusted. Similarly, Huang \emph{et al}. personalize FL with additional terms and a federated attentive message passing (FedAMP) strategy to relieve the influence of Non-IID data \cite{huang2020personalized}, and they guarantee the convergence of the proposed FedAMP and obtains satisfactory results on several widely used datasets.
    
Interpolation is another idea for combining local and global information, which can further be divided into data interpolation and model interpolation. Data interpolation combines local and global data for training, while model interpolation combines local and global model as the personalized model. Mansour \emph{et al}. conduct a systematic empirical study on three personalization strategies, client clustering, data interpolation and model interpolation as well as their theoretical guarantees \cite{mansour2020three}. However, similar to \cite{jiang2019improving}, this method need to be examined on more challenging tasks.

\subsubsection{Personalization layer}
\begin{figure}
\centering 
\includegraphics[width=0.75\textwidth]{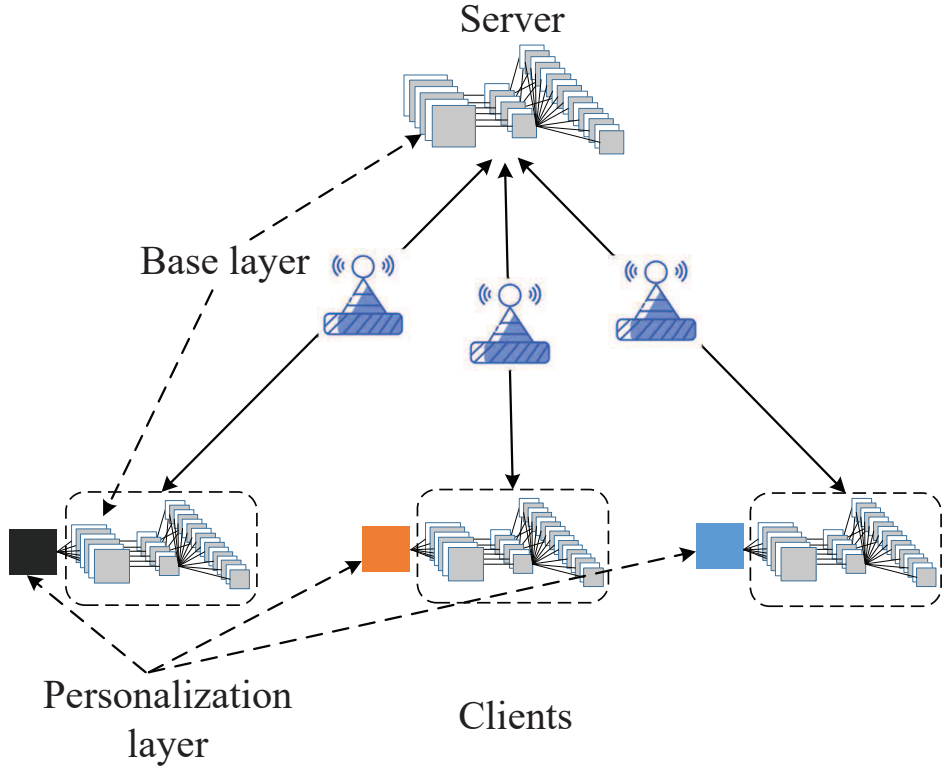} 
\caption{An illustration of horizontal FL with personalization layers.} 
\label{fig:base_personalize}
\end{figure}

As its name suggests, this type of methods allows each client to have personalized layers in the neural network models. As shown in Fig. \ref{fig:base_personalize}, each client model consists of personalization layers (filled blocks) and base layers, and only the base layers need to be uploaded to the server for global model aggregation.

A typical paradigm called FedPer is introduced in \cite{arivazhagan2019federated}, where the base layers are the shallow layers of the neural network that extracts high-level representations and the personalization layers are the deep layers for classifications. The pseudo code for FedPer is shown in Algorithm \ref{alg:fedper} and it is clear to see that except for the personalization layers, FedPer is exactly the same as the original FedAvg algorithm. Experimental results have shown that FedPer can achieve much higher test accuracy than FedAvg, especially on strongly Non-IID data. And it is surprising to find that FedPer has achieved better performance on Non-IID data than on IID data.

\begin{algorithm}[htbp]\footnotesize{
\caption{FedPer} 
\algblock{Begin}{End}
\label{alg:fedper}
\begin{algorithmic}[1]
\State \textbf{Server: }
\State Initialize the shared base model ${\theta_{B}^{0}}$
\State Initialize personalization layers ${\theta_{P_{k}}^{0}}$
\For {each communication round $ t = 1,2,...T\ $}
\State Select $m = C \times K$ clients, where $C \in (0,1)$ 
\For {each \textbf{Client} $k = 1,2,...m$ in parallel}
\State Download ${\theta_{B}^{t}}$ to \textbf{Client $k$}
\State Do \textbf{Client $k$} update and receive ${\theta_{B}^{k}}$ 
\EndFor
\State Update base model ${\theta_{B}^{t}} \leftarrow \sum\limits_{k = 1}^m {\frac{{{n_k}}}{n}{\theta_{B} ^k}}$
\EndFor \\
\State \textbf{Client $k$} update: 
\State Merge base model $\theta_{B}$ and personalization layers $\theta_{P_{k}}$
\For {local epoch from 1 to $E$}
\For {batch $ b \in (1,B)\ $}
\State $(\theta_{B}^{k},\theta_{P_{k}}) \leftarrow (\theta_{B},\theta_{P_{k}}) - \eta \nabla {L_k}(\theta_{B},\theta_{P_{k}};b)$
\EndFor
\EndFor
\State \textbf{Return} $\theta_{B}^{k}$ 
\end{algorithmic}}
\end{algorithm}

By contrast, Liang \emph{et al.} propose an LG-FEDAVG \cite{liang2020think} in which the personalization layers are shallow layers of the neural network and the base layers that are shared with the server are deep layers for class classifications. In addition to supervised local training with the shared global model, LG-FEDAVG also discuss unsupervised learning (autoencoder \cite{kramer1991nonlinear}), self-supervised (jigsaw solving \cite{10.1007/978-3-319-46466-4_5}), and adversarial training with an extra locally constructed model connected to the personalization layers. When calculating the test accuracy on  new devices, all trained local model logits are averaged before choosing the most likely class. Experiment results show that LG-FEDAVG can achieve much better local test accuracy and slightly better new test accuracy than FedAvg.

Personalization layers are able to not only enhance the learning performance on Non-IID data but also reduce communication costs, since only the base layers rather than the whole model need to be shared between the server and clients. One disadvantage is that each client needs to permanently store the personalization layers without releasing them. 

\subsubsection{Multi-task learning}
An alternative to solveing the personalization problem is to treat it as a multi-task learning problem \cite{caruana1997multitask}. For example, MOCHA, a representative framework for federated multi-task learning (FMTL), firstly considers issues of communication cost, stragglers and fault tolerance for FL \cite{NIPS2017_6211080f}. Due to the use of primal-dual optimization method, MOCHA generates separated but related models for each client, which makes it unsuitable for non-convex optimization tasks. Corinzia \textit{et al}. propose an FMTL framework VIRTUAL using a Bayesian network and approximated variational inference that can deal with non-convex models. Their method achieves satisfactory results on several Non-IID datasets, but has difficulties in converging when there is a large number of clients, due to the sequential fine-tuning \cite{corinzia2019variational}. To alleviate the performance degeneration of FL caused by incongruent data distributions, Sattler \textit{et al}. propose a non-convex FMTL framework, called clustered federated learning (CFL), to group local information \cite{sattler2020clustered}. CFL presents a computationally efficient metric of client population distributions based on the cosine similarity, and obtains remarkable results on Non-IID data. However, CFL may pose new challenges to data security due to the reliance on the data similarity.

\subsubsection{Knowledge distillation}
Knowledge distillation \cite{bucilua2006model,hinton2015distilling,gou2021knowledge} is also a promising idea for personalized federated learning. The concept of transfer information from large models to small ones is first proposed by Bucilua \emph{et al}. \cite{bucilua2006model} and popularized by Hinton \emph{et al}. as knowledge distillation. The main motivation in FL is to transfer knowledge from the server or other clients to a certain client to improve its performance on unknown heterogeneous data. In general, there are two kinds of knowledge distillation strategies adopted in Non-IID FL, namely federated transfer learning and domain adaption.

Transfer learning plays a key role in federated knowledge distillation. In \cite{wang2019federated}, the authors train a next-word prediction model using FedAvg with different hyperparameters for personalization to make sure the model can generalize in new mobile devices. Liu \emph{et al}. focus on the secure modeling under decentralized data shards and propose a federated transfer learning (FTL) framework, which leverages homomorphic encryption \cite{gentry2009fully} and secret sharing \cite{chang2018capacity} as protocols \cite{liu2020secure}. Lin \emph{et al}. \cite{lin2020ensemble} utilize an ensemble distillation strategy to robustly fuse multiple models, and the distilled model alleviates the leak risk and computational cost compared to several widely used FL algorithms. The FedDF algorithm is summarized in Algorithm \ref{alg:FedDF}. 
\begin{algorithm}[htbp]\footnotesize{
\caption{An illustration of the homogeneous FedDF.} 
\algblock{Begin}{End}
\label{alg:FedDF}
\begin{algorithmic}[1]
\State Initialize server model $\theta_0$, $K$ clients, a total of $T$ communication rounds, $n_k$ samples for a local dataset of client $k \in K$ and its weight $p_k$, participation ratio $C$.
\For {each communication round $ t = 1,2,...T\ $}
\State $S_t \leftarrow$ random subset ($C$ fraction) of $K$ clients  
\For{each client $k\in S_t$ \textbf{in parallel do}}
\State $\hat{\theta}_t^k \leftarrow$ Local update of FedAvg using $\theta_{t-1}^k$
\EndFor
\State initialize for model fusion $\theta_{t,0} \leftarrow \sum_{k\in S_t} p_k \hat{\theta}_t^k$  

\For{$j$ in $\{1,...,N\}$}
\State sample a mini-batch of samples $\bm{d}$, from e.g. (1) an unlabeled dataset, (2) a generator use ensemble of $\{\hat{\theta}_t^k\}_{k\in S_t}$ to update server student $\theta_{t,j-1}$ through AVGLOGITS

\EndFor
\State $\theta_t \leftarrow \theta_{t, N}$
\EndFor
\State \textbf{Return} $\theta_T$ 
\end{algorithmic}}
\end{algorithm}

From lines 8-10 of the algorithm, we can find that unlabeled data or artificially generated samples are used to assist the knowledge extraction from all participating clients, and the method can be applied to both homogeneous and heterogeneous settings, although the used unlabeled data may lead to an increase in the calculation budget. Similarly, Chang \emph{et al}. propose Cronus, a collaborative robust learning method, by uploading learned features instead of local models to implement local personalization \cite{chang2019cronus}. Li \emph{et al}. propose a framework called decentralized federated learning via mutual knowledge transfer (Def-KT), in which local clients exchange messages directly in a peer-to-peer manner without the participation of the cloud server \cite{li2020decentralized}. The authors state that the performance degeneration of FedAvg on heterogeneous data may be caused by transferring model parameters only, and the key point of Def-KT is to  leverage the advantages of mutual knowledge transfer (MKT) to mitigate the influence of label-shift \cite{zhang2018deep}. Concretely, during each communication round, a subset of selected clients first train their models locally, and then transmit the updated models to the second subset of clients. Then the clients in the second subset can compute two soft predictions (logits) based on their local models and received well trained models. These two calculated logits are used as the dummy labels to update both the local model and received model on each client in the second subset.

Another important chart of knowledge distillation is domain adaption, which emphasizes on eliminating the differences between data shards between clients. A federated adversarial domain adaptation (FADA) algorithm is presented in \cite{peng2019federated}, which uses adversarial adaptation techniques to solve the domain shift problem in FL systems. Li \emph{et al}. propose FedMD algorithm to enable clients train their unique models on local data \cite{li2019fedmd}. The key element of FedMD is to transfer knowledge from a public dataset (without privacy leakage risk) shared by the clients. For example, the initial model of a certain client is firstly trained on a subset of CIFAR100 (public set) and then the personalized training on CIFAR10 (private set) will be performed.

\subsubsection{Lifelong Learning}

Lifelong learning is a fundamental challenge in machine learning, where a model is trained on sequential tasks and each task can be seen only once. The main objective is to maintain the model accuracy without forgetting previously learnt tasks. Hence, it is possible to borrow the idea of lifelong learning to overcome the influence of Non-IID data.

Elastic weight consolidation (EWC) \cite{kirkpatrick2017overcoming} is an effective approach to mitigating catastrophic forgetting in lifelong learning, in which the most important parameters for a specific task \textit{A} are identified. When the model is trained on another task \textit{B}, the learner will be penalized for changing these parameters. By making an analogy between federated learning and lifelong learning, Shoham \textit{et al}. propose a federated curvature (FedCurv) algorithm based on EWC as a solution to Non-IID issues in FL \cite{shoham2019overcoming}. During each round, participants transmit updated models together with the diagonal of the Fisher information matrix, which represents the most informative parameters for the current task. A penalty term is added to the loss function for each participant to promote the convergence towards a globally shared optimum. Furthermore, they assume that the communication cost could be further reduced by a sparse version of the uploaded parameters, without practical verification though. Liu \textit{et al}. combine lifelong and reinforcement learning to form a lifelong federated reinforcement learning (LFRL) architecture \cite{liu2019lifelong}. With LFRL, they enable robots to merge and transfer experience, so that the robot can quickly adapt itself to a new environment.

\subsubsection{Structure adaption}
As mentioned before, training complex \emph{parametric} models like deep neural networks (DNNs) is very challenging in FL especially on Non-IID data. Here we introduce and discuss two useful techniques to accelerate the convergence speed of DNNs in FL. 

One technique is to use adaptive optimizers like Adagrad \cite{JMLR:v12:duchi11a}, Adam \cite{kingma2014adam}, among many others, to replace the standard SGD optimizer. However, these adaptive optimizers often require to accumulate momentum \cite{sutskever2013importance} of the previous gradient information for model update, which may double the uploading communication costs in FL. This is because model training is performed only on local devices and the accumulated gradients (with the same size as the model parameters) also need to be uploaded to the server for aggregation. Reddi \emph{et al.} propose a federated adaptive framework \cite{reddi2020adaptive} to fix this issue. The core idea is simple: the accumulated gradients are calculated upon the averaged global gradients on the server and each client performs the standard SGD for local model training. This not only saves the local computation resources but also reduces the upload communication costs, since the central server is assumed to be much more powerful than the edge devices. 

Another useful technique is to use group normalization (GN) layers \cite{wu2018group} to replace the batch normalization (BN) \cite{ioffe2015batch} layers in DNNs \cite{zhang2020improving}. BN layers calculate both the mean and variance of one batch training data along the batch dimension during the training and track the exponential moving mean and variance for model prediction. However, in FL, each client device has its own data and the calculated batch moving statistics should also be averaged on the server, which cannot represent the actual global statistics. As a result, local moving statistics are very sensitive to the client data distribution and the aggregated global statistics may fail to converge on Non-IID data \cite{NIPS2017_c54e7837}. GN is another option by partitioning the channels of each training data into groups and computing per-group statistics separately. Since the statistics of GN are calculated per data sample, it is invariant to the client data distribution. Experimental results have shown that GN exhibits a faster and more stable convergence profile than BN.

\subsection{System based approach}

\label{sec:system}
\subsubsection{Client clustering}
Most FL approaches assume that the whole system contains only one global model, which is hard to learn all client information especially in a heterogeneous data environment. Therefore, client clustering is proposed to construct a multi-center framework by grouping the clients into different clusters. And those clients with the similar local training data are allocated to the same cluster. However, the data distribution of each client is private and sensitive information and should be kept secret from the central server. To this end, two main kinds of secure data similarity evaluation methods are introduced in the literature: one is to evaluate the similarity of the loss value, and the other is the similarity of model weights.

The first similarity evaluation approach is reported in \cite{mansour2020three,kopparapu2020fedfmc,ghosh2019robust,ghosh2020efficient} by comparing the loss values of different cluster models. The general idea for this technique is straightforward: the server constructs multiple global models instead of a single model and send all cluster models to connected clients for local empirical loss computation. And then each client updates the received cluster model with the smallest loss value and returns it to the server for cluster model aggregation. A representative approach called iterative federated clustering algorithm (IFCA) \cite{ghosh2020efficient} is described in Algorithm \ref{alg:IFCA}, where a one-hot vector $s_{i,j}$ is used to identify the cluster group of the uploaded local model and option I \& II are actually the same thing with different global model updating methods. It should be noticed that IFCA reallocates clients to cluster groups based on the calculated local loss of all cluster models across each communication round, making the download payload $k$ times larger than that in FedAvg. To mitigate this issue, Kopparapu and Lin \cite{DBLP:journals/corr/abs-2006-10937} propose a fork algorithm to group clients only at some particular rounds, which reduces the frequency of doing clustering.

\begin{algorithm}[htbp]\footnotesize{
\caption{Iterative federated clustering algorithm (IFCA)} 
\algblock{Begin}{End}
\label{alg:IFCA}
\begin{algorithmic}[1]
\State \textbf{Input:} number of clusters $k$, any single cluster index $j \in [k]$, the total number of communication round $T$, number of local epochs $E$, mini-batch size $B$, learning rate $\eta$ \\
\For{$t=0,1,...T-1$}
\State \textbf{Server:} Broadcast cluster model $\theta_{j}^{t}$, $j \in [k]$
\State Randomly subsample $m$ participating clients
\For{client $i \in m$ in parallel}
\State Determine cluster group: $\hat{j}=argmin_{j \in [k]}F_{i}(\theta_{j}^{t})$
\State Generate one-hot vector $s_{i}=\left \{ s_{i,j} \right \}_{j=1}^{k}$ with $s_{i,j}=1\left \{ j=\hat{j} \right \}$
\State \textbf{option I} (gradient averaging):
\State \indent Compute gradient: $g_{i}=\hat{\triangledown}F_{i}(\theta_{\hat{j}}^{t})$
\State \textbf{option II} (model averaging):
\State \indent $\tilde{\theta_{i}}=\textrm{\textbf{ClientUpdate}}(\theta_{\hat{j}}^{t}, E, B, \eta)$
\State Send back $s_{i}$ and $g_{i}$ or $\theta_{i}$ to the \textbf{server}
\EndFor
\State \textbf{Server:}
\State \textbf{option I} (gradient averaging): $\theta_{j}^{t+1} \leftarrow \theta_{j}^{t}-\frac{\eta}{m}\sum_{i \in [m]}s_{i,j}g_{i}$
\State \textbf{option II} (model averaging): $\theta_{j}^{t+1} \leftarrow \sum_{i \in [m]}s_{i,j}\theta_{i}/\sum_{i \in [m]}s_{i,j}$
\EndFor
\State \textbf{Return} $\theta_{j}^{T}, j \in [k]$ \\
\State $\textrm{\textbf{ClientUpdate}}(\theta_{\hat{j}}, E, B, \eta)$ at the $i$-th machine
\State $\theta^{i} \leftarrow \theta_{\hat{j}}$
\For{local epoch from 1 to $E$}
\For{batch $b \in [B]$}
\State $\theta^{i} \leftarrow \theta^{i}-\eta\hat{\bigtriangledown}F_{i}(\theta^{i},b) $
\EndFor
\EndFor
\State \textbf{Return} $\theta^{i}$
\end{algorithmic}}
\end{algorithm}

The second approach evaluates the local data similarity and does clustering based on the local model weights. Before clustering the clients, some methods \cite{briggs2020federated,sattler2020clustered} train and warm up the global model with the standard FedAvg algorithm. And then the warmed-up global model is downloaded to each device for local updating (or just calculating the model gradients) and returned to the server. The server can derive the similarity scores, for instance, the cosine similarity, according to the received model weights and group the clients into clusters based on the calculated similarity scores. In addition, Chen \emph{et al.} set the fixed cluster clients before training and train cluster models with corresponding clients separately with the FedAvg algorithm. In addition, Xie \emph{et al.} adopt a stochastic expectation and maximization algorithm \cite{dempster1977maximum} to perform client clustering and model training at the same time. Instead of a similarity score, the L1 distance between the local model $W_{i}$ and the cluster model $\widetilde{W}^{k}$ is computed, as shown in line 4 of Algorithm \ref{alg:fesem}, where $m$ is the total number of clients and $K$ is the total number of clusters. In the expectation step, each local model on client $i$ measure the distance $d_{ik}$ with each global model in cluster $k$ to update cluster assignment $r_{i}^{k}$. In maximization step, all clients are grouped into clusters based on $r_{i}^{k}$ (with the shortest distance) and averaging the cluster models. After that, the derived cluster models are downloaded to the clients within the clusters for local model training. Note that the distance (similarity) calculation method can be replaced by other methods like the cosine similarity. 

\begin{algorithm}[htbp]\footnotesize{
\caption{FeSEM-Federated stochastic EM} 
\algblock{Begin}{End}
\label{alg:fesem}
\begin{algorithmic}[1]
\State Initialize $K$, $\left \{ W_{i} \right \}_{i=1}^{m}$,$\left \{  \widetilde{W}^{k} \right \}_{k=1}^{K}$
\While{stop condition is not satisfied}
\State \textbf{E-Step:}
\State Compute distance $d_{ik}=Dist(W_{i},\widetilde{W}^{k})$ \; $\forall i,k$
\State Update $r_{i}^{k}= \begin{cases}1, & \mbox{if }k=argmin_{j}Dist(W_{i},\widetilde{W}^{j}) \\
0,& \mbox{otherwise}\end{cases}$
\State \textbf{M-step:}
\State Group clients into $C_{k}$ based on $r_{i}^{k}$
\State Update $\widetilde{W}^{k}=\frac{1}{\sum_{i=1}^{m}r_{i}^{k}}\sum_{i=1}^{m}r_{i}^{k}W_{i}$
\For{each cluster $k=1,..K$}
\For{$i \in C_{k}$}
\State Send $\widetilde{W}^{k}$ to client $i$
\State  $W_{i} \leftarrow \textrm{\textbf{ClientUpdate}}(i, \widetilde{W}^{k})$
\EndFor
\EndFor
\EndWhile
\end{algorithmic}}
\end{algorithm}

\begin{figure}
\centering 
\includegraphics[width=0.75\textwidth]{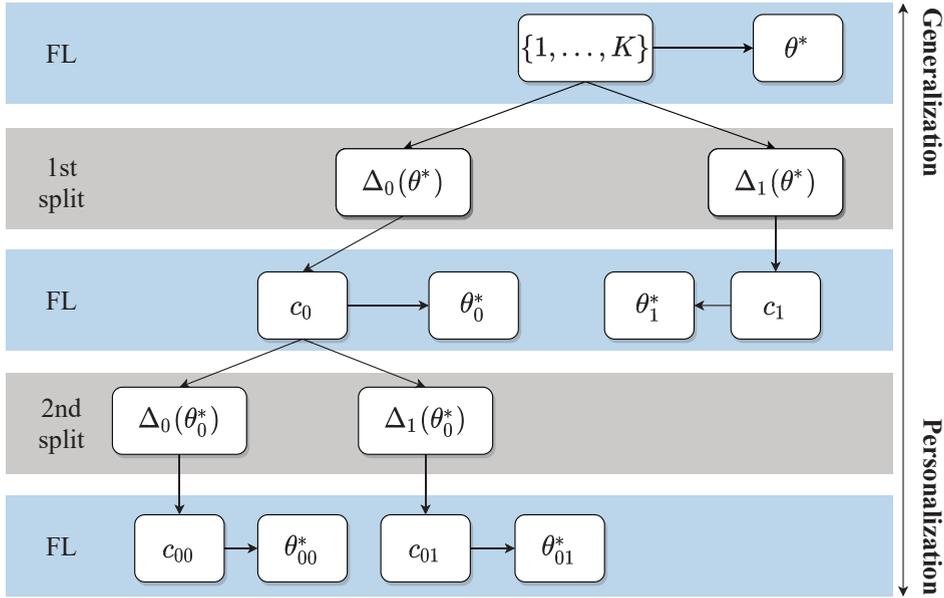} 
\caption{An exemplary parameter tree created by cluster FL.} 
\label{fig:treestructureCFL}
\end{figure}

Besides, model inference or testing in this scenarios is different from the standard FedAvg algorithm due to multiple cluster models involved. Before calculating the test metrics, all the cluster models need to be downloaded to every clients and a simple method to integrate these calculated metrics is doing (weighted) averaging for different cluster models. Another simple approach requires all clients to compute the loss function values of all cluster models and select the cluster model with the smallest loss for calculating the local test metrics. Apart from these two simple methods, Sattler \emph{et al.} propose an interesting tree-based structure \cite{ghosh2020efficient} as shown in Fig. \ref{fig:treestructureCFL}. At the root node resides, a shared global model $\theta^{\ast}$ is trained with the standard FedAvg algorithm over all connected clients. In the next layer (the 1st split layer), the client population is split into two clusters based on the cosine similarities, and the cluster models $\theta_{0}^{\ast}$ and $\theta_{1}^{\ast}$ are trained again by the FedAvg algorithm on the allocated cluster clients. This splitting continues recursively until the stop condition is satisfied. It can be seen that the cluster models near the root node are general models and those located at the bottom layers are personalized models. And different cluster models in the parameter tree can be selected for different applications. Moreover, all the tree models can also be ensembled by the aforementioned two approaches to metrics calculation.

Doing client clustering is necessary and reasonable to deal with Non-IID problems in FL, since aggregating local models with considerably different training data may cause negative knowledge transfer and degrade the shared model performance. In addition, generating multiple global models instead of a single one is beneficial for scalability and flexibility of FL systems so that designers can choose or ensemble different cluster models for specific tasks. However, this method always need to consume additional computation and communication resources for both model training and test.

\subsubsection{System level optimization}
The fundamental system design of FL is first illustrated by Google \cite{bonawitz2019towards} and then become very popular \cite{yin2020fedloc}. WeBank's AI Department has released and maintained an open-source industrial FL framework FATE \cite{webank2019fate}. Additionally, He \textit{et al}. develop an open-source FL library FedML, in which many state-of-the-art FL algorithms are implemented \cite{chaoyanghe2020fedml}. In terms of eliminating the influence of Non-IID data distribution, Jing \textit{et al}. quantify the performance of federated transfer learning (FTL) using FATE on homogeneous and heterogeneous tasks. And they conclude that inter-process communication, data encryption, internet networking condition of FTL are the main bottlenecks and can be improved further \cite{jing2019quantifying}.

\section{Remaining Challenges and Future Directions}
Although a large number of remedies have been proposed for handling Non-IID data distributions in FL, many challenges remain open. 
\begin{itemize}

    \item Privacy protection is a basic and vital purpose of FL, but a plenty of methods designed for addressing Non-IID data such as data sharing, knowledge distillation inevitably increase the risk of privacy exposure. It is still not clear to what extent these methods harm the data privacy, and there is no quantitative measures to identify the degree of privacy leakage.
    
    
    \item FL contains a large number of hyperparameters, e.g., the total number of clients, the number of local epochs, and client dropout probability, which strongly differ from algorithm to algorithm, making it hard to benchmark the real Non-IID performance of these algorithms.
    
    \item Although two real image dataset are introduced in \cite{luo2019real,hsu2020federated}, a universal homogeneous and heterogeneous benchmark dataset still lacks for FL. Generate synthetic Non-IID data by arbitrary partitioning datasets cannot effectively evaluate the performance of a method proposed for handling Non-IID data \cite{li2021federated}. 
    
    
    
    \item Personalized federated learning is promising for IOT edge devices, although it has not yet received enough attention. The system design, model deployment, communication cost reduction in unstable and limited wireless networks and task adaption with limited computation budget remain an open question.
    
    
    

    \item Although the vertical FL framework is able to be widely adopted in practical industry scenarios, only a handful work has attempted to cope with the potential problems caused by Non-IID distribution under vertical FL settings. For instance, data features with overlapping is one of the most typical cases of Non-IID scenarios for vertical FL, which deserves more attention. In addition, other Non-IID cases with both attribute and label skew, and 'different features, different labels' and 'crowdsourcing skew' have not been fully investigated. 
    
    \item There is an increasing demand for automated machine learning (AutoML) and lots of algorithms have been presented on neural architecture search (NAS) in practical scenarios \cite{cui2019fast,DBLP:journals/corr/abs-2011-00826}. However, only a limited amount of research on federated NAS has been reported \cite{DBLP:journals/corr/abs-2002-06352,singh2020differentially,FedNAS,zhu2020real} and little work has considered the influence of Non-IID distributions.
    
\end{itemize}

Given the above remaining challenges, we suggest the following future directions:

\begin{itemize}
    \item Define quantitative criteria for measuring the degree of privacy leakage so that the maximum amount of shared data can be bounded. 
    \item Construct FL benchmark problems reflecting real-world requirements and challenges and provide standard FL hyperparameter settings so that FL algorithms can be fairly compared.
    \item More cases of Non-IID scenarios for vertical FL need to be further explored, and the corresponding model performance in non-standard Non-IID settings (e.g. with overlapped data features) can be compared to that in standard Non-IID setting. In addition, corresponding algorithms can also be proposed to different Non-IID cases in vertical FL.
    \item Federated neural architecture search (FNAS) \cite{zhu2021FNAS} is an emerging research direction and the effect of NON-IID has not been clearly investigated. Therefore, handling Non-IID problems in FNAS is an interesting future direction.
    
\end{itemize}

\section{Conclusion}
This paper aims to provide a systematic understanding of Non-IID data in federated learning systems and provide a comprehensive overview of existing techniques for handling Non-IID data. A detailed categorization of Non-IID data distributions are given with illustrative examples, several of which have not been discussed in the literature. Different from existing surveys on FL, we focus on the impact of Non-IID data on both parametric and non-parametric models in horizontal and vertical FL. We point out that Non-IID data distributions mainly affect the learning performance of parametric models in horizontal FL and complex models like DNN mor sensitive to client data distributions. We indicate that some existing work on handling Non-IID data, such as local fine-tuning and data sharing achieve better convergence performance often at the cost of either increased local computation and communication resources or even data privacy. Other methods including personalization and client clustering need to change the structure of the vanilla FL framework, making it no longer possible to generate a global model for all clients. Finally, we discuss remaining challenges in handling Non-IID in FL, and suggest a few research directions to handle the open questions.




\bibliographystyle{cas-model2-names}

\bibliography{cas-refs}

\begin{thebibliography}{172}
\expandafter\ifx\csname natexlab\endcsname\relax\def\natexlab#1{#1}\fi
\providecommand{\url}[1]{\texttt{#1}}
\providecommand{\href}[2]{#2}
\providecommand{\path}[1]{#1}
\providecommand{\DOIprefix}{doi:}
\providecommand{\ArXivprefix}{arXiv:}
\providecommand{\URLprefix}{URL: }
\providecommand{\Pubmedprefix}{pmid:}
\providecommand{\doi}[1]{\href{http://dx.doi.org/#1}{\path{#1}}}
\providecommand{\Pubmed}[1]{\href{pmid:#1}{\path{#1}}}
\providecommand{\bibinfo}[2]{#2}
\ifx\xfnm\relax \def\xfnm[#1]{\unskip,\space#1}\fi
\bibitem[{Aledhari et~al.(2020)Aledhari, Razzak, Parizi and Saeed}]{9153560}
\bibinfo{author}{Aledhari, M.}, \bibinfo{author}{Razzak, R.},
  \bibinfo{author}{Parizi, R.M.}, \bibinfo{author}{Saeed, F.},
  \bibinfo{year}{2020}.
\newblock \bibinfo{title}{Federated learning: A survey on enabling
  technologies, protocols, and applications}.
\newblock \bibinfo{journal}{IEEE Access} \bibinfo{volume}{8},
  \bibinfo{pages}{140699--140725}.
\newblock \DOIprefix\doi{10.1109/ACCESS.2020.3013541}.
\bibitem[{Aono et~al.(2017)Aono, Hayashi, Wang, Moriai
  et~al.}]{aono2017privacy}
\bibinfo{author}{Aono, Y.}, \bibinfo{author}{Hayashi, T.},
  \bibinfo{author}{Wang, L.}, \bibinfo{author}{Moriai, S.}, et~al.,
  \bibinfo{year}{2017}.
\newblock \bibinfo{title}{Privacy-preserving deep learning: Revisited and
  enhanced}, in: \bibinfo{booktitle}{International Conference on Applications
  and Techniques in Information Security}, \bibinfo{organization}{Springer}.
  pp. \bibinfo{pages}{100--110}.
\bibitem[{Arivazhagan et~al.(2019)Arivazhagan, Aggarwal, Singh and
  Choudhary}]{arivazhagan2019federated}
\bibinfo{author}{Arivazhagan, M.G.}, \bibinfo{author}{Aggarwal, V.},
  \bibinfo{author}{Singh, A.K.}, \bibinfo{author}{Choudhary, S.},
  \bibinfo{year}{2019}.
\newblock \bibinfo{title}{Federated learning with personalization layers}.
\newblock \bibinfo{journal}{arXiv preprint arXiv:1912.00818} .
\bibitem[{Armknecht et~al.(2015)Armknecht, Boyd, Carr, Gj{\o}steen,
  J{\"a}schke, Reuter and Strand}]{cryptoeprint:2015:1192}
\bibinfo{author}{Armknecht, F.}, \bibinfo{author}{Boyd, C.},
  \bibinfo{author}{Carr, C.}, \bibinfo{author}{Gj{\o}steen, K.},
  \bibinfo{author}{J{\"a}schke, A.}, \bibinfo{author}{Reuter, C.A.},
  \bibinfo{author}{Strand, M.}, \bibinfo{year}{2015}.
\newblock \bibinfo{title}{A guide to fully homomorphic encryption}.
\newblock \bibinfo{howpublished}{Cryptology ePrint Archive, Report 2015/1192}.
\newblock \bibinfo{note}{\url{https://eprint.iacr.org/2015/1192}}.
\bibitem[{Bonawitz et~al.(2019)Bonawitz, Eichner, Grieskamp, Huba, Ingerman,
  Ivanov, Kiddon, Kone{\v{c}}n{\`y}, Mazzocchi, McMahan
  et~al.}]{bonawitz2019towards}
\bibinfo{author}{Bonawitz, K.}, \bibinfo{author}{Eichner, H.},
  \bibinfo{author}{Grieskamp, W.}, \bibinfo{author}{Huba, D.},
  \bibinfo{author}{Ingerman, A.}, \bibinfo{author}{Ivanov, V.},
  \bibinfo{author}{Kiddon, C.}, \bibinfo{author}{Kone{\v{c}}n{\`y}, J.},
  \bibinfo{author}{Mazzocchi, S.}, \bibinfo{author}{McMahan, H.B.}, et~al.,
  \bibinfo{year}{2019}.
\newblock \bibinfo{title}{Towards federated learning at scale: System design}.
\newblock \bibinfo{journal}{arXiv preprint arXiv:1902.01046} .
\bibitem[{Bonawitz et~al.(2017)Bonawitz, Ivanov, Kreuter, Marcedone, McMahan,
  Patel, Ramage, Segal and Seth}]{bonawitz2017practical}
\bibinfo{author}{Bonawitz, K.}, \bibinfo{author}{Ivanov, V.},
  \bibinfo{author}{Kreuter, B.}, \bibinfo{author}{Marcedone, A.},
  \bibinfo{author}{McMahan, H.B.}, \bibinfo{author}{Patel, S.},
  \bibinfo{author}{Ramage, D.}, \bibinfo{author}{Segal, A.},
  \bibinfo{author}{Seth, K.}, \bibinfo{year}{2017}.
\newblock \bibinfo{title}{Practical secure aggregation for privacy-preserving
  machine learning}, in: \bibinfo{booktitle}{proceedings of the 2017 ACM SIGSAC
  Conference on Computer and Communications Security}, pp.
  \bibinfo{pages}{1175--1191}.
\bibitem[{Briggs et~al.(2020a)Briggs, Fan and Andras}]{briggs2020federated}
\bibinfo{author}{Briggs, C.}, \bibinfo{author}{Fan, Z.},
  \bibinfo{author}{Andras, P.}, \bibinfo{year}{2020}a.
\newblock \bibinfo{title}{Federated learning with hierarchical clustering of
  local updates to improve training on non-iid data}, in:
  \bibinfo{booktitle}{2020 International Joint Conference on Neural Networks
  (IJCNN)}, \bibinfo{organization}{IEEE}. pp. \bibinfo{pages}{1--9}.
\bibitem[{Briggs et~al.(2020b)Briggs, Fan and Andras}]{briggs2020review}
\bibinfo{author}{Briggs, C.}, \bibinfo{author}{Fan, Z.},
  \bibinfo{author}{Andras, P.}, \bibinfo{year}{2020}b.
\newblock \bibinfo{title}{A review of privacy preserving federated learning for
  private iot analytics}.
\newblock \bibinfo{journal}{arXiv preprint arXiv:2004.11794} .
\bibitem[{Bucilua et~al.(2006)Bucilua, Caruana and
  Niculescu-Mizil}]{bucilua2006model}
\bibinfo{author}{Bucilua, C.}, \bibinfo{author}{Caruana, R.},
  \bibinfo{author}{Niculescu-Mizil, A.}, \bibinfo{year}{2006}.
\newblock \bibinfo{title}{Model compression}, in:
  \bibinfo{booktitle}{Proceedings of the 12th ACM SIGKDD international
  conference on Knowledge discovery and data mining}, pp.
  \bibinfo{pages}{535--541}.
\bibitem[{Caldas et~al.(2018)Caldas, Kone{\v{c}}ny, McMahan and
  Talwalkar}]{caldas2018expanding}
\bibinfo{author}{Caldas, S.}, \bibinfo{author}{Kone{\v{c}}ny, J.},
  \bibinfo{author}{McMahan, H.B.}, \bibinfo{author}{Talwalkar, A.},
  \bibinfo{year}{2018}.
\newblock \bibinfo{title}{Expanding the reach of federated learning by reducing
  client resource requirements}.
\newblock \bibinfo{journal}{arXiv preprint arXiv:1812.07210} .
\bibitem[{Campos et~al.(2017)Campos, Sastre, Yag{\"u}es, Bellver, Gir{\'o}-i
  Nieto and Torres}]{campos2017distributed}
\bibinfo{author}{Campos, V.}, \bibinfo{author}{Sastre, F.},
  \bibinfo{author}{Yag{\"u}es, M.}, \bibinfo{author}{Bellver, M.},
  \bibinfo{author}{Gir{\'o}-i Nieto, X.}, \bibinfo{author}{Torres, J.},
  \bibinfo{year}{2017}.
\newblock \bibinfo{title}{Distributed training strategies for a computer vision
  deep learning algorithm on a distributed gpu cluster}.
\newblock \bibinfo{journal}{Procedia Computer Science} \bibinfo{volume}{108},
  \bibinfo{pages}{315--324}.
\bibitem[{Caruana(1997)}]{caruana1997multitask}
\bibinfo{author}{Caruana, R.}, \bibinfo{year}{1997}.
\newblock \bibinfo{title}{Multitask learning}.
\newblock \bibinfo{journal}{Machine learning} \bibinfo{volume}{28},
  \bibinfo{pages}{41--75}.
\bibitem[{Chang et~al.(2019)Chang, Shejwalkar, Shokri and
  Houmansadr}]{chang2019cronus}
\bibinfo{author}{Chang, H.}, \bibinfo{author}{Shejwalkar, V.},
  \bibinfo{author}{Shokri, R.}, \bibinfo{author}{Houmansadr, A.},
  \bibinfo{year}{2019}.
\newblock \bibinfo{title}{Cronus: Robust and heterogeneous collaborative
  learning with black-box knowledge transfer}.
\newblock \bibinfo{journal}{arXiv preprint arXiv:1912.11279} .
\bibitem[{Chang and Tandon(2018)}]{chang2018capacity}
\bibinfo{author}{Chang, W.T.}, \bibinfo{author}{Tandon, R.},
  \bibinfo{year}{2018}.
\newblock \bibinfo{title}{On the capacity of secure distributed matrix
  multiplication}, in: \bibinfo{booktitle}{2018 IEEE Global Communications
  Conference (GLOBECOM)}, \bibinfo{organization}{IEEE}. pp.
  \bibinfo{pages}{1--6}.
\bibitem[{Chen et~al.(2018)Chen, Luo, Dong, Li and He}]{chen2018federated}
\bibinfo{author}{Chen, F.}, \bibinfo{author}{Luo, M.}, \bibinfo{author}{Dong,
  Z.}, \bibinfo{author}{Li, Z.}, \bibinfo{author}{He, X.},
  \bibinfo{year}{2018}.
\newblock \bibinfo{title}{Federated meta-learning with fast convergence and
  efficient communication}.
\newblock \bibinfo{journal}{arXiv preprint arXiv:1802.07876} .
\bibitem[{Chen et~al.(2015)Chen, He, Benesty, Khotilovich, Tang, Cho
  et~al.}]{chen2015xgboost}
\bibinfo{author}{Chen, T.}, \bibinfo{author}{He, T.}, \bibinfo{author}{Benesty,
  M.}, \bibinfo{author}{Khotilovich, V.}, \bibinfo{author}{Tang, Y.},
  \bibinfo{author}{Cho, H.}, et~al., \bibinfo{year}{2015}.
\newblock \bibinfo{title}{Xgboost: extreme gradient boosting}.
\newblock \bibinfo{journal}{R package version 0.4-2} \bibinfo{volume}{1}.
\bibitem[{Chen et~al.(2020)Chen, Jin, Sun and
  Yin}]{DBLP:journals/corr/abs-2007-06081}
\bibinfo{author}{Chen, T.}, \bibinfo{author}{Jin, X.}, \bibinfo{author}{Sun,
  Y.}, \bibinfo{author}{Yin, W.}, \bibinfo{year}{2020}.
\newblock \bibinfo{title}{{VAFL:} a method of vertical asynchronous federated
  learning}.
\newblock \bibinfo{journal}{CoRR} \bibinfo{volume}{abs/2007.06081}.
\newblock \URLprefix \url{https://arxiv.org/abs/2007.06081},
  \href{http://arxiv.org/abs/2007.06081}{\tt arXiv:2007.06081}.
\bibitem[{Chen et~al.(2019)Chen, Sun and Jin}]{chen2019communication}
\bibinfo{author}{Chen, Y.}, \bibinfo{author}{Sun, X.}, \bibinfo{author}{Jin,
  Y.}, \bibinfo{year}{2019}.
\newblock \bibinfo{title}{Communication-efficient federated deep learning with
  layerwise asynchronous model update and temporally weighted aggregation}.
\newblock \bibinfo{journal}{IEEE Transactions on Neural Networks and Learning
  Systems} \bibinfo{volume}{31}, \bibinfo{pages}{4229 -- 4238}.
\bibitem[{Cheng et~al.(2019)Cheng, Fan, Jin, Liu, Chen and
  Yang}]{cheng2019secureboost}
\bibinfo{author}{Cheng, K.}, \bibinfo{author}{Fan, T.}, \bibinfo{author}{Jin,
  Y.}, \bibinfo{author}{Liu, Y.}, \bibinfo{author}{Chen, T.},
  \bibinfo{author}{Yang, Q.}, \bibinfo{year}{2019}.
\newblock \bibinfo{title}{Secureboost: A lossless federated learning
  framework}.
\newblock \bibinfo{journal}{arXiv preprint arXiv:1901.08755} .
\bibitem[{Cohen et~al.(2017)Cohen, Afshar, Tapson and van
  Schaik}]{cohen2017emnist}
\bibinfo{author}{Cohen, G.}, \bibinfo{author}{Afshar, S.},
  \bibinfo{author}{Tapson, J.}, \bibinfo{author}{van Schaik, A.},
  \bibinfo{year}{2017}.
\newblock \bibinfo{title}{Emnist: an extension of mnist to handwritten
  letters}.
\newblock \href{http://arxiv.org/abs/1702.05373}{\tt arXiv:1702.05373}.
\bibitem[{Corinzia et~al.(2019)Corinzia, Beuret and
  Buhmann}]{corinzia2019variational}
\bibinfo{author}{Corinzia, L.}, \bibinfo{author}{Beuret, A.},
  \bibinfo{author}{Buhmann, J.M.}, \bibinfo{year}{2019}.
\newblock \bibinfo{title}{Variational federated multi-task learning}.
\newblock \bibinfo{journal}{arXiv preprint arXiv:1906.06268} .
\bibitem[{Cui et~al.(2019)Cui, Chen, Li, Liu, Shen and Jia}]{cui2019fast}
\bibinfo{author}{Cui, J.}, \bibinfo{author}{Chen, P.}, \bibinfo{author}{Li,
  R.}, \bibinfo{author}{Liu, S.}, \bibinfo{author}{Shen, X.},
  \bibinfo{author}{Jia, J.}, \bibinfo{year}{2019}.
\newblock \bibinfo{title}{Fast and practical neural architecture search}, in:
  \bibinfo{booktitle}{Proceedings of the IEEE/CVF International Conference on
  Computer Vision}, pp. \bibinfo{pages}{6509--6518}.
\bibitem[{Dean et~al.(2012)Dean, Corrado, Monga, Chen, Devin, Le, Mao, Ranzato,
  Senior, Tucker et~al.}]{dean2012large}
\bibinfo{author}{Dean, J.}, \bibinfo{author}{Corrado, G.S.},
  \bibinfo{author}{Monga, R.}, \bibinfo{author}{Chen, K.},
  \bibinfo{author}{Devin, M.}, \bibinfo{author}{Le, Q.V.},
  \bibinfo{author}{Mao, M.Z.}, \bibinfo{author}{Ranzato, M.},
  \bibinfo{author}{Senior, A.}, \bibinfo{author}{Tucker, P.}, et~al.,
  \bibinfo{year}{2012}.
\newblock \bibinfo{title}{Large scale distributed deep networks} .
\bibitem[{Dempster et~al.(1977)Dempster, Laird and Rubin}]{dempster1977maximum}
\bibinfo{author}{Dempster, A.P.}, \bibinfo{author}{Laird, N.M.},
  \bibinfo{author}{Rubin, D.B.}, \bibinfo{year}{1977}.
\newblock \bibinfo{title}{Maximum likelihood from incomplete data via the em
  algorithm}.
\newblock \bibinfo{journal}{Journal of the Royal Statistical Society: Series B
  (Methodological)} \bibinfo{volume}{39}, \bibinfo{pages}{1--22}.
\bibitem[{Deng et~al.(2020)Deng, Kamani and Mahdavi}]{deng2020adaptive}
\bibinfo{author}{Deng, Y.}, \bibinfo{author}{Kamani, M.M.},
  \bibinfo{author}{Mahdavi, M.}, \bibinfo{year}{2020}.
\newblock \bibinfo{title}{Adaptive personalized federated learning}.
\newblock \bibinfo{journal}{arXiv preprint arXiv:2003.13461} .
\bibitem[{Diffie and Hellman(1976)}]{diffie1976new}
\bibinfo{author}{Diffie, W.}, \bibinfo{author}{Hellman, M.},
  \bibinfo{year}{1976}.
\newblock \bibinfo{title}{New directions in cryptography}.
\newblock \bibinfo{journal}{IEEE transactions on Information Theory}
  \bibinfo{volume}{22}, \bibinfo{pages}{644--654}.
\bibitem[{Dinh et~al.(2020)Dinh, Tran and Nguyen}]{dinh2020personalized}
\bibinfo{author}{Dinh, C.T.}, \bibinfo{author}{Tran, N.H.},
  \bibinfo{author}{Nguyen, T.D.}, \bibinfo{year}{2020}.
\newblock \bibinfo{title}{Personalized federated learning with moreau
  envelopes}, in: \bibinfo{booktitle}{Advances in Neural Information Processing
  Systems 33: Annual Conference on Neural Information Processing Systems 2020,
  NeurIPS 2020, December 6-12, 2020, virtual}.
\bibitem[{Duan et~al.(2019)Duan, Liu, Chen, Tan, Ren, Qiao and
  Liang}]{duan2019astraea}
\bibinfo{author}{Duan, M.}, \bibinfo{author}{Liu, D.}, \bibinfo{author}{Chen,
  X.}, \bibinfo{author}{Tan, Y.}, \bibinfo{author}{Ren, J.},
  \bibinfo{author}{Qiao, L.}, \bibinfo{author}{Liang, L.},
  \bibinfo{year}{2019}.
\newblock \bibinfo{title}{Astraea: Self-balancing federated learning for
  improving classification accuracy of mobile deep learning applications}, in:
  \bibinfo{booktitle}{2019 IEEE 37th International Conference on Computer
  Design (ICCD)}, \bibinfo{organization}{IEEE}. pp. \bibinfo{pages}{246--254}.
\bibitem[{Duchi et~al.(2011)Duchi, Hazan and Singer}]{JMLR:v12:duchi11a}
\bibinfo{author}{Duchi, J.}, \bibinfo{author}{Hazan, E.},
  \bibinfo{author}{Singer, Y.}, \bibinfo{year}{2011}.
\newblock \bibinfo{title}{Adaptive subgradient methods for online learning and
  stochastic optimization}.
\newblock \bibinfo{journal}{Journal of Machine Learning Research}
  \bibinfo{volume}{12}, \bibinfo{pages}{2121--2159}.
\newblock \URLprefix \url{http://jmlr.org/papers/v12/duchi11a.html}.
\bibitem[{Dwork(2008)}]{dwork2008differential}
\bibinfo{author}{Dwork, C.}, \bibinfo{year}{2008}.
\newblock \bibinfo{title}{Differential privacy: A survey of results}, in:
  \bibinfo{booktitle}{International conference on theory and applications of
  models of computation}, \bibinfo{organization}{Springer}. pp.
  \bibinfo{pages}{1--19}.
\bibitem[{Fallah et~al.(2020)Fallah, Mokhtari and
  Ozdaglar}]{fallah2020personalized}
\bibinfo{author}{Fallah, A.}, \bibinfo{author}{Mokhtari, A.},
  \bibinfo{author}{Ozdaglar, A.E.}, \bibinfo{year}{2020}.
\newblock \bibinfo{title}{Personalized federated learning with theoretical
  guarantees: {A} model-agnostic meta-learning approach}, in:
  \bibinfo{booktitle}{Advances in Neural Information Processing Systems 33:
  Annual Conference on Neural Information Processing Systems 2020, NeurIPS
  2020, December 6-12, 2020, virtual}.
\bibitem[{Feng and Yu(2020)}]{feng2020multi}
\bibinfo{author}{Feng, S.}, \bibinfo{author}{Yu, H.}, \bibinfo{year}{2020}.
\newblock \bibinfo{title}{Multi-participant multi-class vertical federated
  learning}.
\newblock \bibinfo{journal}{arXiv preprint arXiv:2001.11154} .
\bibitem[{Finn et~al.(2017)Finn, Abbeel and Levine}]{finn2017model}
\bibinfo{author}{Finn, C.}, \bibinfo{author}{Abbeel, P.},
  \bibinfo{author}{Levine, S.}, \bibinfo{year}{2017}.
\newblock \bibinfo{title}{Model-agnostic meta-learning for fast adaptation of
  deep networks}, in: \bibinfo{booktitle}{International Conference on Machine
  Learning}, \bibinfo{organization}{PMLR}. pp. \bibinfo{pages}{1126--1135}.
\bibitem[{Garcia-Molina et~al.(2016)Garcia-Molina, Joglekar, Marcus,
  Parameswaran and Verroios}]{garcia2016challenges}
\bibinfo{author}{Garcia-Molina, H.}, \bibinfo{author}{Joglekar, M.},
  \bibinfo{author}{Marcus, A.}, \bibinfo{author}{Parameswaran, A.},
  \bibinfo{author}{Verroios, V.}, \bibinfo{year}{2016}.
\newblock \bibinfo{title}{Challenges in data crowdsourcing}.
\newblock \bibinfo{journal}{IEEE Transactions on Knowledge and Data
  Engineering} \bibinfo{volume}{28}, \bibinfo{pages}{901--911}.
\bibitem[{Gardner and Dorling(1998)}]{gardner1998artificial}
\bibinfo{author}{Gardner, M.W.}, \bibinfo{author}{Dorling, S.},
  \bibinfo{year}{1998}.
\newblock \bibinfo{title}{Artificial neural networks (the multilayer
  perceptron)—a review of applications in the atmospheric sciences}.
\newblock \bibinfo{journal}{Atmospheric environment} \bibinfo{volume}{32},
  \bibinfo{pages}{2627--2636}.
\bibitem[{Gentry et~al.(2009)}]{gentry2009fully}
\bibinfo{author}{Gentry, C.}, et~al., \bibinfo{year}{2009}.
\newblock \bibinfo{title}{A fully homomorphic encryption scheme}.
  volume~\bibinfo{volume}{20}.
\newblock \bibinfo{publisher}{Stanford university Stanford}.
\bibitem[{Geyer et~al.(2017)Geyer, Klein and Nabi}]{geyer2017differentially}
\bibinfo{author}{Geyer, R.C.}, \bibinfo{author}{Klein, T.},
  \bibinfo{author}{Nabi, M.}, \bibinfo{year}{2017}.
\newblock \bibinfo{title}{Differentially private federated learning: A client
  level perspective}.
\newblock \bibinfo{journal}{arXiv preprint arXiv:1712.07557} .
\bibitem[{Ghosh et~al.(2020)Ghosh, Chung, Yin and
  Ramchandran}]{ghosh2020efficient}
\bibinfo{author}{Ghosh, A.}, \bibinfo{author}{Chung, J.}, \bibinfo{author}{Yin,
  D.}, \bibinfo{author}{Ramchandran, K.}, \bibinfo{year}{2020}.
\newblock \bibinfo{title}{An efficient framework for clustered federated
  learning}.
\newblock \bibinfo{journal}{arXiv preprint arXiv:2006.04088} .
\bibitem[{Ghosh et~al.(2019)Ghosh, Hong, Yin and Ramchandran}]{ghosh2019robust}
\bibinfo{author}{Ghosh, A.}, \bibinfo{author}{Hong, J.}, \bibinfo{author}{Yin,
  D.}, \bibinfo{author}{Ramchandran, K.}, \bibinfo{year}{2019}.
\newblock \bibinfo{title}{Robust federated learning in a heterogeneous
  environment}.
\newblock \bibinfo{journal}{arXiv preprint arXiv:1906.06629} .
\bibitem[{Goodfellow et~al.(2014)Goodfellow, Pouget-Abadie, Mirza, Xu,
  Warde-Farley, Ozair, Courville and Bengio}]{NIPS2014_5ca3e9b1}
\bibinfo{author}{Goodfellow, I.}, \bibinfo{author}{Pouget-Abadie, J.},
  \bibinfo{author}{Mirza, M.}, \bibinfo{author}{Xu, B.},
  \bibinfo{author}{Warde-Farley, D.}, \bibinfo{author}{Ozair, S.},
  \bibinfo{author}{Courville, A.}, \bibinfo{author}{Bengio, Y.},
  \bibinfo{year}{2014}.
\newblock \bibinfo{title}{Generative adversarial nets}, in:
  \bibinfo{editor}{Ghahramani, Z.}, \bibinfo{editor}{Welling, M.},
  \bibinfo{editor}{Cortes, C.}, \bibinfo{editor}{Lawrence, N.},
  \bibinfo{editor}{Weinberger, K.Q.} (Eds.), \bibinfo{booktitle}{Advances in
  Neural Information Processing Systems}, \bibinfo{publisher}{Curran
  Associates, Inc.}
\newblock \URLprefix
  \url{https://proceedings.neurips.cc/paper/2014/file/5ca3e9b122f61f8f06494c97b1afccf3-Paper.pdf}.
\bibitem[{Gou et~al.(2021)Gou, Yu, Maybank and Tao}]{gou2021knowledge}
\bibinfo{author}{Gou, J.}, \bibinfo{author}{Yu, B.}, \bibinfo{author}{Maybank,
  S.J.}, \bibinfo{author}{Tao, D.}, \bibinfo{year}{2021}.
\newblock \bibinfo{title}{Knowledge distillation: A survey}.
\newblock \bibinfo{journal}{International Journal of Computer Vision} ,
  \bibinfo{pages}{1--31}.
\bibitem[{Gupta et~al.(2020)Gupta, Choudhary, Tang, Wei, Wang, Huang,
  Kejariwal, Ramchandran and Mahoney}]{gupta2020fast}
\bibinfo{author}{Gupta, V.}, \bibinfo{author}{Choudhary, D.},
  \bibinfo{author}{Tang, P.T.P.}, \bibinfo{author}{Wei, X.},
  \bibinfo{author}{Wang, X.}, \bibinfo{author}{Huang, Y.},
  \bibinfo{author}{Kejariwal, A.}, \bibinfo{author}{Ramchandran, K.},
  \bibinfo{author}{Mahoney, M.W.}, \bibinfo{year}{2020}.
\newblock \bibinfo{title}{Fast distributed training of deep neural networks:
  Dynamic communication thresholding for model and data parallelism}.
\newblock \bibinfo{journal}{arXiv preprint arXiv:2010.08899} .
\bibitem[{Haddadpour and Mahdavi(2019)}]{haddadpour2019convergence}
\bibinfo{author}{Haddadpour, F.}, \bibinfo{author}{Mahdavi, M.},
  \bibinfo{year}{2019}.
\newblock \bibinfo{title}{On the convergence of local descent methods in
  federated learning}.
\newblock \bibinfo{journal}{arXiv preprint arXiv:1910.14425} .
\bibitem[{Han et~al.(2016)Han, Mao and Dally}]{DBLP:journals/corr/HanMD15}
\bibinfo{author}{Han, S.}, \bibinfo{author}{Mao, H.}, \bibinfo{author}{Dally,
  W.J.}, \bibinfo{year}{2016}.
\newblock \bibinfo{title}{Deep compression: Compressing deep neural network
  with pruning, trained quantization and huffman coding}, in:
  \bibinfo{editor}{Bengio, Y.}, \bibinfo{editor}{LeCun, Y.} (Eds.),
  \bibinfo{booktitle}{4th International Conference on Learning Representations,
  {ICLR} 2016, San Juan, Puerto Rico, May 2-4, 2016, Conference Track
  Proceedings}.
\newblock \URLprefix \url{http://arxiv.org/abs/1510.00149}.
\bibitem[{Hanzely and Richt{\'a}rik(2020)}]{hanzely2020federated}
\bibinfo{author}{Hanzely, F.}, \bibinfo{author}{Richt{\'a}rik, P.},
  \bibinfo{year}{2020}.
\newblock \bibinfo{title}{Federated learning of a mixture of global and local
  models}.
\newblock \bibinfo{journal}{arXiv preprint arXiv:2002.05516} .
\bibitem[{Hao et~al.(2019)Hao, Li, Xu, Liu and Yang}]{hao2019towards}
\bibinfo{author}{Hao, M.}, \bibinfo{author}{Li, H.}, \bibinfo{author}{Xu, G.},
  \bibinfo{author}{Liu, S.}, \bibinfo{author}{Yang, H.}, \bibinfo{year}{2019}.
\newblock \bibinfo{title}{Towards efficient and privacy-preserving federated
  deep learning}, in: \bibinfo{booktitle}{ICC 2019-2019 IEEE International
  Conference on Communications (ICC)}, \bibinfo{organization}{IEEE}. pp.
  \bibinfo{pages}{1--6}.
\bibitem[{Hardy et~al.(2017)Hardy, Henecka, Ivey-Law, Nock, Patrini, Smith and
  Thorne}]{hardy2017private}
\bibinfo{author}{Hardy, S.}, \bibinfo{author}{Henecka, W.},
  \bibinfo{author}{Ivey-Law, H.}, \bibinfo{author}{Nock, R.},
  \bibinfo{author}{Patrini, G.}, \bibinfo{author}{Smith, G.},
  \bibinfo{author}{Thorne, B.}, \bibinfo{year}{2017}.
\newblock \bibinfo{title}{Private federated learning on vertically partitioned
  data via entity resolution and additively homomorphic encryption}.
\newblock \bibinfo{journal}{arXiv preprint arXiv:1711.10677} .
\bibitem[{He et~al.(2020a)He, Annavaram and Avestimehr}]{FedNAS}
\bibinfo{author}{He, C.}, \bibinfo{author}{Annavaram, M.},
  \bibinfo{author}{Avestimehr, S.}, \bibinfo{year}{2020}a.
\newblock \bibinfo{title}{Fednas: Federated deep learning via neural
  architecture search}, in: \bibinfo{booktitle}{CVPR 2020 Workshop on Neural
  Architecture Search and Beyond for Representation Learning}.
\bibitem[{He et~al.(2020b)He, Li, So, Zhang, Wang, Wang, Vepakomma, Singh, Qiu,
  Shen, Zhao, Kang, Liu, Raskar, Yang, Annavaram and
  Avestimehr}]{chaoyanghe2020fedml}
\bibinfo{author}{He, C.}, \bibinfo{author}{Li, S.}, \bibinfo{author}{So, J.},
  \bibinfo{author}{Zhang, M.}, \bibinfo{author}{Wang, H.},
  \bibinfo{author}{Wang, X.}, \bibinfo{author}{Vepakomma, P.},
  \bibinfo{author}{Singh, A.}, \bibinfo{author}{Qiu, H.},
  \bibinfo{author}{Shen, L.}, \bibinfo{author}{Zhao, P.},
  \bibinfo{author}{Kang, Y.}, \bibinfo{author}{Liu, Y.},
  \bibinfo{author}{Raskar, R.}, \bibinfo{author}{Yang, Q.},
  \bibinfo{author}{Annavaram, M.}, \bibinfo{author}{Avestimehr, S.},
  \bibinfo{year}{2020}b.
\newblock \bibinfo{title}{Fedml: A research library and benchmark for federated
  machine learning}.
\newblock \bibinfo{journal}{arXiv preprint arXiv:2007.13518} .
\bibitem[{Hinton et~al.(2015)Hinton, Vinyals and Dean}]{hinton2015distilling}
\bibinfo{author}{Hinton, G.}, \bibinfo{author}{Vinyals, O.},
  \bibinfo{author}{Dean, J.}, \bibinfo{year}{2015}.
\newblock \bibinfo{title}{Distilling the knowledge in a neural network}.
\newblock \bibinfo{journal}{arXiv preprint arXiv:1503.02531} .
\bibitem[{Hitaj et~al.(2017)Hitaj, Ateniese and Perez-Cruz}]{hitaj2017deep}
\bibinfo{author}{Hitaj, B.}, \bibinfo{author}{Ateniese, G.},
  \bibinfo{author}{Perez-Cruz, F.}, \bibinfo{year}{2017}.
\newblock \bibinfo{title}{Deep models under the gan: information leakage from
  collaborative deep learning}, in: \bibinfo{booktitle}{Proceedings of the 2017
  ACM SIGSAC Conference on Computer and Communications Security}, pp.
  \bibinfo{pages}{603--618}.
\bibitem[{Hochreiter and Schmidhuber(1997)}]{hochreiter1997long}
\bibinfo{author}{Hochreiter, S.}, \bibinfo{author}{Schmidhuber, J.},
  \bibinfo{year}{1997}.
\newblock \bibinfo{title}{Long short-term memory}.
\newblock \bibinfo{journal}{Neural computation} \bibinfo{volume}{9},
  \bibinfo{pages}{1735--1780}.
\bibitem[{Hosseinalipour et~al.(2020)Hosseinalipour, Brinton, Aggarwal, Dai and
  Chiang}]{hosseinalipour2020federated}
\bibinfo{author}{Hosseinalipour, S.}, \bibinfo{author}{Brinton, C.G.},
  \bibinfo{author}{Aggarwal, V.}, \bibinfo{author}{Dai, H.},
  \bibinfo{author}{Chiang, M.}, \bibinfo{year}{2020}.
\newblock \bibinfo{title}{From federated learning to fog learning: Towards
  large-scale distributed machine learning in heterogeneous wireless networks}.
\newblock \bibinfo{journal}{arXiv preprint arXiv:2006.03594} .
\bibitem[{Hsu et~al.(2020)Hsu, Qi and Brown}]{hsu2020federated}
\bibinfo{author}{Hsu, T.M.H.}, \bibinfo{author}{Qi, H.},
  \bibinfo{author}{Brown, M.}, \bibinfo{year}{2020}.
\newblock \bibinfo{title}{{Federated Visual Classification with Real-World Data
  Distribution}}, in: \bibinfo{booktitle}{Proceedings of the European
  Conference on Computer Vision (ECCV)}.
\bibitem[{Huang et~al.(2021)Huang, Chu, Zhou, Wang, Liu, Pei and
  Zhang}]{huang2020personalized}
\bibinfo{author}{Huang, Y.}, \bibinfo{author}{Chu, L.}, \bibinfo{author}{Zhou,
  Z.}, \bibinfo{author}{Wang, L.}, \bibinfo{author}{Liu, J.},
  \bibinfo{author}{Pei, J.}, \bibinfo{author}{Zhang, Y.}, \bibinfo{year}{2021}.
\newblock \bibinfo{title}{Personalized cross-silo federated learning on non-iid
  data}.
\newblock \bibinfo{journal}{Association for the Advancement of Artificial
  Intelligence (AAAI)} .
\bibitem[{Imteaj et~al.(2020)Imteaj, Thakker, Wang, Li and
  Amini}]{imteaj2020federated}
\bibinfo{author}{Imteaj, A.}, \bibinfo{author}{Thakker, U.},
  \bibinfo{author}{Wang, S.}, \bibinfo{author}{Li, J.}, \bibinfo{author}{Amini,
  M.H.}, \bibinfo{year}{2020}.
\newblock \bibinfo{title}{Federated learning for resource-constrained iot
  devices: Panoramas and state-of-the-art}.
\newblock \bibinfo{journal}{arXiv preprint arXiv:2002.10610} .
\bibitem[{Ioffe(2017)}]{NIPS2017_c54e7837}
\bibinfo{author}{Ioffe, S.}, \bibinfo{year}{2017}.
\newblock \bibinfo{title}{Batch renormalization: Towards reducing minibatch
  dependence in batch-normalized models}, in: \bibinfo{editor}{Guyon, I.},
  \bibinfo{editor}{Luxburg, U.V.}, \bibinfo{editor}{Bengio, S.},
  \bibinfo{editor}{Wallach, H.}, \bibinfo{editor}{Fergus, R.},
  \bibinfo{editor}{Vishwanathan, S.}, \bibinfo{editor}{Garnett, R.} (Eds.),
  \bibinfo{booktitle}{Advances in Neural Information Processing Systems},
  \bibinfo{publisher}{Curran Associates, Inc.}
\newblock \URLprefix
  \url{https://proceedings.neurips.cc/paper/2017/file/c54e7837e0cd0ced286cb5995327d1ab-Paper.pdf}.
\bibitem[{Ioffe and Szegedy(2015)}]{ioffe2015batch}
\bibinfo{author}{Ioffe, S.}, \bibinfo{author}{Szegedy, C.},
  \bibinfo{year}{2015}.
\newblock \bibinfo{title}{Batch normalization: Accelerating deep network
  training by reducing internal covariate shift}, in:
  \bibinfo{booktitle}{International conference on machine learning},
  \bibinfo{organization}{PMLR}. pp. \bibinfo{pages}{448--456}.
\bibitem[{Jiang et~al.(2019)Jiang, Kone{\v{c}}n{\`y}, Rush and
  Kannan}]{jiang2019improving}
\bibinfo{author}{Jiang, Y.}, \bibinfo{author}{Kone{\v{c}}n{\`y}, J.},
  \bibinfo{author}{Rush, K.}, \bibinfo{author}{Kannan, S.},
  \bibinfo{year}{2019}.
\newblock \bibinfo{title}{Improving federated learning personalization via
  model agnostic meta learning}.
\newblock \bibinfo{journal}{arXiv preprint arXiv:1909.12488} .
\bibitem[{Jing et~al.(2019)Jing, Wang, Zhang, Tian and
  Chen}]{jing2019quantifying}
\bibinfo{author}{Jing, Q.}, \bibinfo{author}{Wang, W.}, \bibinfo{author}{Zhang,
  J.}, \bibinfo{author}{Tian, H.}, \bibinfo{author}{Chen, K.},
  \bibinfo{year}{2019}.
\newblock \bibinfo{title}{Quantifying the performance of federated transfer
  learning}.
\newblock \bibinfo{journal}{ArXiv abs/1912.12795} .
\bibitem[{Kairouz et~al.(2019)Kairouz, McMahan, Avent, Bellet, Bennis, Bhagoji,
  Bonawitz, Charles, Cormode, Cummings et~al.}]{kairouz2019advances}
\bibinfo{author}{Kairouz, P.}, \bibinfo{author}{McMahan, H.B.},
  \bibinfo{author}{Avent, B.}, \bibinfo{author}{Bellet, A.},
  \bibinfo{author}{Bennis, M.}, \bibinfo{author}{Bhagoji, A.N.},
  \bibinfo{author}{Bonawitz, K.}, \bibinfo{author}{Charles, Z.},
  \bibinfo{author}{Cormode, G.}, \bibinfo{author}{Cummings, R.}, et~al.,
  \bibinfo{year}{2019}.
\newblock \bibinfo{title}{Advances and open problems in federated learning}.
\newblock \bibinfo{journal}{arXiv preprint arXiv:1912.04977} .
\bibitem[{Karimireddy et~al.(2020)Karimireddy, Kale, Mohri, Reddi, Stich and
  Suresh}]{karimireddy2020scaffold}
\bibinfo{author}{Karimireddy, S.P.}, \bibinfo{author}{Kale, S.},
  \bibinfo{author}{Mohri, M.}, \bibinfo{author}{Reddi, S.},
  \bibinfo{author}{Stich, S.}, \bibinfo{author}{Suresh, A.T.},
  \bibinfo{year}{2020}.
\newblock \bibinfo{title}{Scaffold: Stochastic controlled averaging for
  federated learning}, in: \bibinfo{booktitle}{International Conference on
  Machine Learning}, \bibinfo{organization}{PMLR}. pp.
  \bibinfo{pages}{5132--5143}.
\bibitem[{Ke et~al.(2017)Ke, Meng, Finley, Wang, Chen, Ma, Ye and
  Liu}]{ke2017lightgbm}
\bibinfo{author}{Ke, G.}, \bibinfo{author}{Meng, Q.}, \bibinfo{author}{Finley,
  T.}, \bibinfo{author}{Wang, T.}, \bibinfo{author}{Chen, W.},
  \bibinfo{author}{Ma, W.}, \bibinfo{author}{Ye, Q.}, \bibinfo{author}{Liu,
  T.Y.}, \bibinfo{year}{2017}.
\newblock \bibinfo{title}{Lightgbm: A highly efficient gradient boosting
  decision tree}.
\newblock \bibinfo{journal}{Advances in neural information processing systems}
  \bibinfo{volume}{30}, \bibinfo{pages}{3146--3154}.
\bibitem[{Keuper and Preundt(2016)}]{keuper2016distributed}
\bibinfo{author}{Keuper, J.}, \bibinfo{author}{Preundt, F.J.},
  \bibinfo{year}{2016}.
\newblock \bibinfo{title}{Distributed training of deep neural networks:
  Theoretical and practical limits of parallel scalability}, in:
  \bibinfo{booktitle}{2016 2nd Workshop on Machine Learning in HPC Environments
  (MLHPC)}, \bibinfo{organization}{IEEE}. pp. \bibinfo{pages}{19--26}.
\bibitem[{Kingma and Ba(2014)}]{kingma2014adam}
\bibinfo{author}{Kingma, D.P.}, \bibinfo{author}{Ba, J.}, \bibinfo{year}{2014}.
\newblock \bibinfo{title}{Adam: A method for stochastic optimization}.
\newblock \bibinfo{journal}{arXiv preprint arXiv:1412.6980} .
\bibitem[{Kirkpatrick et~al.(2017)Kirkpatrick, Pascanu, Rabinowitz, Veness,
  Desjardins, Rusu, Milan, Quan, Ramalho, Grabska-Barwinska
  et~al.}]{kirkpatrick2017overcoming}
\bibinfo{author}{Kirkpatrick, J.}, \bibinfo{author}{Pascanu, R.},
  \bibinfo{author}{Rabinowitz, N.}, \bibinfo{author}{Veness, J.},
  \bibinfo{author}{Desjardins, G.}, \bibinfo{author}{Rusu, A.A.},
  \bibinfo{author}{Milan, K.}, \bibinfo{author}{Quan, J.},
  \bibinfo{author}{Ramalho, T.}, \bibinfo{author}{Grabska-Barwinska, A.},
  et~al., \bibinfo{year}{2017}.
\newblock \bibinfo{title}{Overcoming catastrophic forgetting in neural
  networks}.
\newblock \bibinfo{journal}{Proceedings of the national academy of sciences}
  \bibinfo{volume}{114}, \bibinfo{pages}{3521--3526}.
\bibitem[{Kone{\v{c}}n{\`y} et~al.(2016)Kone{\v{c}}n{\`y}, McMahan, Yu,
  Richt{\'a}rik, Suresh and Bacon}]{konevcny2016federated}
\bibinfo{author}{Kone{\v{c}}n{\`y}, J.}, \bibinfo{author}{McMahan, H.B.},
  \bibinfo{author}{Yu, F.X.}, \bibinfo{author}{Richt{\'a}rik, P.},
  \bibinfo{author}{Suresh, A.T.}, \bibinfo{author}{Bacon, D.},
  \bibinfo{year}{2016}.
\newblock \bibinfo{title}{Federated learning: Strategies for improving
  communication efficiency}.
\newblock \bibinfo{journal}{arXiv preprint arXiv:1610.05492} .
\bibitem[{Kopparapu and Lin(2020a)}]{kopparapu2020fedfmc}
\bibinfo{author}{Kopparapu, K.}, \bibinfo{author}{Lin, E.},
  \bibinfo{year}{2020}a.
\newblock \bibinfo{title}{Fedfmc: Sequential efficient federated learning on
  non-iid data}.
\newblock \bibinfo{journal}{arXiv preprint arXiv:2006.10937} .
\bibitem[{Kopparapu and Lin(2020b)}]{DBLP:journals/corr/abs-2006-10937}
\bibinfo{author}{Kopparapu, K.}, \bibinfo{author}{Lin, E.},
  \bibinfo{year}{2020}b.
\newblock \bibinfo{title}{Fedfmc: Sequential efficient federated learning on
  non-iid data}.
\newblock \bibinfo{journal}{CoRR} \bibinfo{volume}{abs/2006.10937}.
\newblock \URLprefix \url{https://arxiv.org/abs/2006.10937},
  \href{http://arxiv.org/abs/2006.10937}{\tt arXiv:2006.10937}.
\bibitem[{Kramer(1991)}]{kramer1991nonlinear}
\bibinfo{author}{Kramer, M.A.}, \bibinfo{year}{1991}.
\newblock \bibinfo{title}{Nonlinear principal component analysis using
  autoassociative neural networks}.
\newblock \bibinfo{journal}{AIChE journal} \bibinfo{volume}{37},
  \bibinfo{pages}{233--243}.
\bibitem[{Krizhevsky et~al.(2009)Krizhevsky, Hinton
  et~al.}]{krizhevsky2009learning}
\bibinfo{author}{Krizhevsky, A.}, \bibinfo{author}{Hinton, G.}, et~al.,
  \bibinfo{year}{2009}.
\newblock \bibinfo{title}{Learning multiple layers of features from tiny
  images} .
\bibitem[{Kulkarni et~al.(2020)Kulkarni, Kulkarni and
  Pant}]{kulkarni2020survey}
\bibinfo{author}{Kulkarni, V.}, \bibinfo{author}{Kulkarni, M.},
  \bibinfo{author}{Pant, A.}, \bibinfo{year}{2020}.
\newblock \bibinfo{title}{Survey of personalization techniques for federated
  learning}, in: \bibinfo{booktitle}{2020 Fourth World Conference on Smart
  Trends in Systems, Security and Sustainability (WorldS4)},
  \bibinfo{organization}{IEEE}. pp. \bibinfo{pages}{794--797}.
\bibitem[{Lan et~al.(2019)Lan, Zhang, Du, Lin and Huang}]{lan2019introduction}
\bibinfo{author}{Lan, Q.}, \bibinfo{author}{Zhang, Z.}, \bibinfo{author}{Du,
  Y.}, \bibinfo{author}{Lin, Z.}, \bibinfo{author}{Huang, K.},
  \bibinfo{year}{2019}.
\newblock \bibinfo{title}{An introduction to communication efficient edge
  machine learning}.
\newblock \bibinfo{journal}{arXiv preprint arXiv:1912.01554} .
\bibitem[{LeCun et~al.(1995)LeCun, Bengio et~al.}]{lecun1995convolutional}
\bibinfo{author}{LeCun, Y.}, \bibinfo{author}{Bengio, Y.}, et~al.,
  \bibinfo{year}{1995}.
\newblock \bibinfo{title}{Convolutional networks for images, speech, and time
  series}.
\newblock \bibinfo{journal}{The handbook of brain theory and neural networks}
  \bibinfo{volume}{3361}, \bibinfo{pages}{1995}.
\bibitem[{Li et~al.(2020a)Li, Li and Varshney}]{li2020decentralized}
\bibinfo{author}{Li, C.}, \bibinfo{author}{Li, G.}, \bibinfo{author}{Varshney,
  P.K.}, \bibinfo{year}{2020}a.
\newblock \bibinfo{title}{Decentralized federated learning via mutual knowledge
  transfer}.
\newblock \bibinfo{journal}{arXiv preprint arXiv:2012.13063} .
\bibitem[{Li and Wang(2019)}]{li2019fedmd}
\bibinfo{author}{Li, D.}, \bibinfo{author}{Wang, J.}, \bibinfo{year}{2019}.
\newblock \bibinfo{title}{Fedmd: Heterogenous federated learning via model
  distillation}.
\newblock \bibinfo{journal}{arXiv preprint arXiv:1910.03581} .
\bibitem[{Li and Han(2019)}]{li2019end}
\bibinfo{author}{Li, H.}, \bibinfo{author}{Han, T.}, \bibinfo{year}{2019}.
\newblock \bibinfo{title}{An end-to-end encrypted neural network for gradient
  updates transmission in federated learning}.
\newblock \bibinfo{journal}{arXiv preprint arXiv:1908.08340} .
\bibitem[{Li et~al.(2015)Li, Kadav, Kruus and Ungureanu}]{li2015malt}
\bibinfo{author}{Li, H.}, \bibinfo{author}{Kadav, A.}, \bibinfo{author}{Kruus,
  E.}, \bibinfo{author}{Ungureanu, C.}, \bibinfo{year}{2015}.
\newblock \bibinfo{title}{Malt: distributed data-parallelism for existing ml
  applications}, in: \bibinfo{booktitle}{Proceedings of the Tenth European
  Conference on Computer Systems}, pp. \bibinfo{pages}{1--16}.
\bibitem[{Li et~al.(2021)Li, Diao, Chen and He}]{li2021federated}
\bibinfo{author}{Li, Q.}, \bibinfo{author}{Diao, Y.}, \bibinfo{author}{Chen,
  Q.}, \bibinfo{author}{He, B.}, \bibinfo{year}{2021}.
\newblock \bibinfo{title}{Federated learning on non-iid data silos: An
  experimental study}.
\newblock \href{http://arxiv.org/abs/2102.02079}{\tt arXiv:2102.02079}.
\bibitem[{Li et~al.(2020b)Li, He and Song}]{li2020model}
\bibinfo{author}{Li, Q.}, \bibinfo{author}{He, B.}, \bibinfo{author}{Song, D.},
  \bibinfo{year}{2020}b.
\newblock \bibinfo{title}{Model-agnostic round-optimal federated learning via
  knowledge transfer}.
\newblock \bibinfo{journal}{arXiv preprint arXiv:2010.01017} .
\bibitem[{Li et~al.(2019)Li, Wen, Wu, Hu, Wang, Li, Liu and He}]{li2019survey}
\bibinfo{author}{Li, Q.}, \bibinfo{author}{Wen, Z.}, \bibinfo{author}{Wu, Z.},
  \bibinfo{author}{Hu, S.}, \bibinfo{author}{Wang, N.}, \bibinfo{author}{Li,
  Y.}, \bibinfo{author}{Liu, X.}, \bibinfo{author}{He, B.},
  \bibinfo{year}{2019}.
\newblock \bibinfo{title}{A survey on federated learning systems: vision, hype
  and reality for data privacy and protection}.
\newblock \bibinfo{journal}{arXiv preprint arXiv:1907.09693} .
\bibitem[{Li et~al.(2020c)Li, Sahu, Talwalkar and Smith}]{li2020federated}
\bibinfo{author}{Li, T.}, \bibinfo{author}{Sahu, A.K.},
  \bibinfo{author}{Talwalkar, A.}, \bibinfo{author}{Smith, V.},
  \bibinfo{year}{2020}c.
\newblock \bibinfo{title}{Federated learning: Challenges, methods, and future
  directions}.
\newblock \bibinfo{journal}{IEEE Signal Processing Magazine}
  \bibinfo{volume}{37}, \bibinfo{pages}{50--60}.
\bibitem[{Liang et~al.(2020)Liang, Liu, Ziyin, Allen, Auerbach, Brent,
  Salakhutdinov and Morency}]{liang2020think}
\bibinfo{author}{Liang, P.P.}, \bibinfo{author}{Liu, T.},
  \bibinfo{author}{Ziyin, L.}, \bibinfo{author}{Allen, N.B.},
  \bibinfo{author}{Auerbach, R.P.}, \bibinfo{author}{Brent, D.},
  \bibinfo{author}{Salakhutdinov, R.}, \bibinfo{author}{Morency, L.P.},
  \bibinfo{year}{2020}.
\newblock \bibinfo{title}{Think locally, act globally: Federated learning with
  local and global representations}.
\newblock \bibinfo{journal}{arXiv preprint arXiv:2001.01523} .
\bibitem[{Liang et~al.(2021)Liang, Liu, Luo, He, Chen and
  Yang}]{liang2021selfsupervised}
\bibinfo{author}{Liang, X.}, \bibinfo{author}{Liu, Y.}, \bibinfo{author}{Luo,
  J.}, \bibinfo{author}{He, Y.}, \bibinfo{author}{Chen, T.},
  \bibinfo{author}{Yang, Q.}, \bibinfo{year}{2021}.
\newblock \bibinfo{title}{Self-supervised cross-silo federated neural
  architecture search}.
\newblock \href{http://arxiv.org/abs/2101.11896}{\tt arXiv:2101.11896}.
\bibitem[{Lim et~al.(2020)Lim, Luong, Hoang, Jiao, Liang, Yang, Niyato and
  Miao}]{lim2020federated}
\bibinfo{author}{Lim, W.Y.B.}, \bibinfo{author}{Luong, N.C.},
  \bibinfo{author}{Hoang, D.T.}, \bibinfo{author}{Jiao, Y.},
  \bibinfo{author}{Liang, Y.C.}, \bibinfo{author}{Yang, Q.},
  \bibinfo{author}{Niyato, D.}, \bibinfo{author}{Miao, C.},
  \bibinfo{year}{2020}.
\newblock \bibinfo{title}{Federated learning in mobile edge networks: A
  comprehensive survey}.
\newblock \bibinfo{journal}{IEEE Communications Surveys \& Tutorials}
  \bibinfo{volume}{22}, \bibinfo{pages}{2031--2063}.
\bibitem[{Lin et~al.(2020)Lin, Kong, Stich and Jaggi}]{lin2020ensemble}
\bibinfo{author}{Lin, T.}, \bibinfo{author}{Kong, L.}, \bibinfo{author}{Stich,
  S.U.}, \bibinfo{author}{Jaggi, M.}, \bibinfo{year}{2020}.
\newblock \bibinfo{title}{Ensemble distillation for robust model fusion in
  federated learning}.
\newblock \bibinfo{journal}{34th Conference on Neural Information Processing
  Systems (NeurIPS 2020)} .
\bibitem[{Liu et~al.(2019a)Liu, Wang and Liu}]{liu2019lifelong}
\bibinfo{author}{Liu, B.}, \bibinfo{author}{Wang, L.}, \bibinfo{author}{Liu,
  M.}, \bibinfo{year}{2019}a.
\newblock \bibinfo{title}{Lifelong federated reinforcement learning: a learning
  architecture for navigation in cloud robotic systems}.
\newblock \bibinfo{journal}{IEEE Robotics and Automation Letters}
  \bibinfo{volume}{4}, \bibinfo{pages}{4555--4562}.
\bibitem[{Liu et~al.(2020a)Liu, Kang, Xing, Chen and Yang}]{9076003}
\bibinfo{author}{Liu, Y.}, \bibinfo{author}{Kang, Y.}, \bibinfo{author}{Xing,
  C.}, \bibinfo{author}{Chen, T.}, \bibinfo{author}{Yang, Q.},
  \bibinfo{year}{2020}a.
\newblock \bibinfo{title}{A secure federated transfer learning framework}.
\newblock \bibinfo{journal}{IEEE Intelligent Systems} \bibinfo{volume}{35},
  \bibinfo{pages}{70--82}.
\newblock \DOIprefix\doi{10.1109/MIS.2020.2988525}.
\bibitem[{Liu et~al.(2020b)Liu, Kang, Xing, Chen and Yang}]{liu2020secure}
\bibinfo{author}{Liu, Y.}, \bibinfo{author}{Kang, Y.}, \bibinfo{author}{Xing,
  C.}, \bibinfo{author}{Chen, T.}, \bibinfo{author}{Yang, Q.},
  \bibinfo{year}{2020}b.
\newblock \bibinfo{title}{A secure federated transfer learning framework}.
\newblock \bibinfo{journal}{IEEE Intelligent Systems} \bibinfo{volume}{35},
  \bibinfo{pages}{70--82}.
\bibitem[{Liu et~al.(2019b)Liu, Kang, Zhang, Li, Cheng, Chen, Hong and
  Yang}]{liu2019communication}
\bibinfo{author}{Liu, Y.}, \bibinfo{author}{Kang, Y.}, \bibinfo{author}{Zhang,
  X.}, \bibinfo{author}{Li, L.}, \bibinfo{author}{Cheng, Y.},
  \bibinfo{author}{Chen, T.}, \bibinfo{author}{Hong, M.},
  \bibinfo{author}{Yang, Q.}, \bibinfo{year}{2019}b.
\newblock \bibinfo{title}{A communication efficient collaborative learning
  framework for distributed features}.
\newblock \bibinfo{journal}{arXiv preprint arXiv:1912.11187} .
\bibitem[{Luo et~al.(2019)Luo, Wu, Luo, Huang, Huang, Liu and
  Yang}]{luo2019real}
\bibinfo{author}{Luo, J.}, \bibinfo{author}{Wu, X.}, \bibinfo{author}{Luo, Y.},
  \bibinfo{author}{Huang, A.}, \bibinfo{author}{Huang, Y.},
  \bibinfo{author}{Liu, Y.}, \bibinfo{author}{Yang, Q.}, \bibinfo{year}{2019}.
\newblock \bibinfo{title}{Real-world image datasets for federated learning}.
\newblock \bibinfo{journal}{arXiv preprint arXiv:1910.11089} .
\bibitem[{Lyu et~al.(2020)Lyu, Yu and Yang}]{lyu2020threats}
\bibinfo{author}{Lyu, L.}, \bibinfo{author}{Yu, H.}, \bibinfo{author}{Yang,
  Q.}, \bibinfo{year}{2020}.
\newblock \bibinfo{title}{Threats to federated learning: A survey}.
\newblock \bibinfo{journal}{arXiv preprint arXiv:2003.02133} .
\bibitem[{Mansour et~al.(2020)Mansour, Mohri, Ro and Suresh}]{mansour2020three}
\bibinfo{author}{Mansour, Y.}, \bibinfo{author}{Mohri, M.},
  \bibinfo{author}{Ro, J.}, \bibinfo{author}{Suresh, A.T.},
  \bibinfo{year}{2020}.
\newblock \bibinfo{title}{Three approaches for personalization with
  applications to federated learning}.
\newblock \bibinfo{journal}{arXiv preprint arXiv:2002.10619} .
\bibitem[{McDonald et~al.(2010)McDonald, Hall and
  Mann}]{mcdonald2010distributed}
\bibinfo{author}{McDonald, R.}, \bibinfo{author}{Hall, K.},
  \bibinfo{author}{Mann, G.}, \bibinfo{year}{2010}.
\newblock \bibinfo{title}{Distributed training strategies for the structured
  perceptron}, in: \bibinfo{booktitle}{Human language technologies: The 2010
  annual conference of the North American chapter of the association for
  computational linguistics}, pp. \bibinfo{pages}{456--464}.
\bibitem[{McMahan et~al.(2017a)McMahan, Moore, Ramage, Hampson and
  y~Arcas}]{mcmahan2017communication}
\bibinfo{author}{McMahan, B.}, \bibinfo{author}{Moore, E.},
  \bibinfo{author}{Ramage, D.}, \bibinfo{author}{Hampson, S.},
  \bibinfo{author}{y~Arcas, B.A.}, \bibinfo{year}{2017}a.
\newblock \bibinfo{title}{Communication-efficient learning of deep networks
  from decentralized data}, in: \bibinfo{booktitle}{Artificial Intelligence and
  Statistics}, pp. \bibinfo{pages}{1273--1282}.
\bibitem[{McMahan et~al.(2017b)McMahan, Ramage, Talwar and
  Zhang}]{mcmahan2017learning}
\bibinfo{author}{McMahan, H.B.}, \bibinfo{author}{Ramage, D.},
  \bibinfo{author}{Talwar, K.}, \bibinfo{author}{Zhang, L.},
  \bibinfo{year}{2017}b.
\newblock \bibinfo{title}{Learning differentially private recurrent language
  models}.
\newblock \bibinfo{journal}{arXiv preprint arXiv:1710.06963} .
\bibitem[{Mills et~al.(2019)Mills, Hu and Min}]{mills2019communication}
\bibinfo{author}{Mills, J.}, \bibinfo{author}{Hu, J.}, \bibinfo{author}{Min,
  G.}, \bibinfo{year}{2019}.
\newblock \bibinfo{title}{Communication-efficient federated learning for
  wireless edge intelligence in iot}.
\newblock \bibinfo{journal}{IEEE Internet of Things Journal}
  \bibinfo{volume}{7}, \bibinfo{pages}{5986--5994}.
\bibitem[{Montgomery et~al.(2021)Montgomery, Peck and
  Vining}]{montgomery2021introduction}
\bibinfo{author}{Montgomery, D.C.}, \bibinfo{author}{Peck, E.A.},
  \bibinfo{author}{Vining, G.G.}, \bibinfo{year}{2021}.
\newblock \bibinfo{title}{Introduction to linear regression analysis}.
\newblock \bibinfo{publisher}{John Wiley \& Sons}.
\bibitem[{Moreau(1963)}]{moreau1963proprietes}
\bibinfo{author}{Moreau, J.J.}, \bibinfo{year}{1963}.
\newblock \bibinfo{title}{Propri{\'e}t{\'e}s des applications
  {\guillemotleft}prox{\guillemotright}}.
\newblock \bibinfo{journal}{Comptes rendus hebdomadaires des s{\'e}ances de
  l'Acad{\'e}mie des sciences} \bibinfo{volume}{256},
  \bibinfo{pages}{1069--1071}.
\bibitem[{Naseri et~al.(2020)Naseri, Hayes and
  De~Cristofaro}]{naseri2020toward}
\bibinfo{author}{Naseri, M.}, \bibinfo{author}{Hayes, J.},
  \bibinfo{author}{De~Cristofaro, E.}, \bibinfo{year}{2020}.
\newblock \bibinfo{title}{Toward robustness and privacy in federated learning:
  Experimenting with local and central differential privacy}.
\newblock \bibinfo{journal}{arXiv preprint arXiv:2009.03561} .
\bibitem[{Nichol et~al.(2018)Nichol, Achiam and Schulman}]{nichol2018first}
\bibinfo{author}{Nichol, A.}, \bibinfo{author}{Achiam, J.},
  \bibinfo{author}{Schulman, J.}, \bibinfo{year}{2018}.
\newblock \bibinfo{title}{On first-order meta-learning algorithms}.
\newblock \bibinfo{journal}{arXiv preprint arXiv:1803.02999} .
\bibitem[{Nock et~al.(2018)Nock, Hardy, Henecka, Ivey-Law, Patrini, Smith and
  Thorne}]{nock2018entity}
\bibinfo{author}{Nock, R.}, \bibinfo{author}{Hardy, S.},
  \bibinfo{author}{Henecka, W.}, \bibinfo{author}{Ivey-Law, H.},
  \bibinfo{author}{Patrini, G.}, \bibinfo{author}{Smith, G.},
  \bibinfo{author}{Thorne, B.}, \bibinfo{year}{2018}.
\newblock \bibinfo{title}{Entity resolution and federated learning get a
  federated resolution}.
\newblock \href{http://arxiv.org/abs/1803.04035}{\tt arXiv:1803.04035}.
\bibitem[{Noroozi and Favaro(2016)}]{10.1007/978-3-319-46466-4_5}
\bibinfo{author}{Noroozi, M.}, \bibinfo{author}{Favaro, P.},
  \bibinfo{year}{2016}.
\newblock \bibinfo{title}{Unsupervised learning of visual representations by
  solving jigsaw puzzles}, in: \bibinfo{editor}{Leibe, B.},
  \bibinfo{editor}{Matas, J.}, \bibinfo{editor}{Sebe, N.},
  \bibinfo{editor}{Welling, M.} (Eds.), \bibinfo{booktitle}{Computer Vision --
  ECCV 2016}, \bibinfo{publisher}{Springer International Publishing},
  \bibinfo{address}{Cham}. pp. \bibinfo{pages}{69--84}.
\bibitem[{Paillier(1999)}]{pascal1999public}
\bibinfo{author}{Paillier, P.}, \bibinfo{year}{1999}.
\newblock \bibinfo{title}{Public-key cryptosystems based on composite degree
  residuosity classes}, in: \bibinfo{editor}{Stern, J.} (Ed.),
  \bibinfo{booktitle}{Advances in Cryptology --- EUROCRYPT '99},
  \bibinfo{publisher}{Springer Berlin Heidelberg}, \bibinfo{address}{Berlin,
  Heidelberg}. pp. \bibinfo{pages}{223--238}.
\bibitem[{Peng et~al.(2019)Peng, Huang, Zhu and Saenko}]{peng2019federated}
\bibinfo{author}{Peng, X.}, \bibinfo{author}{Huang, Z.}, \bibinfo{author}{Zhu,
  Y.}, \bibinfo{author}{Saenko, K.}, \bibinfo{year}{2019}.
\newblock \bibinfo{title}{Federated adversarial domain adaptation}.
\newblock \bibinfo{journal}{arXiv preprint arXiv:1911.02054} .
\bibitem[{Pfrommer et~al.(1997)Pfrommer, C{\^o}t{\'e}, Louie and
  Cohen}]{pfrommer1997relaxation}
\bibinfo{author}{Pfrommer, B.G.}, \bibinfo{author}{C{\^o}t{\'e}, M.},
  \bibinfo{author}{Louie, S.G.}, \bibinfo{author}{Cohen, M.L.},
  \bibinfo{year}{1997}.
\newblock \bibinfo{title}{Relaxation of crystals with the quasi-newton method}.
\newblock \bibinfo{journal}{Journal of Computational Physics}
  \bibinfo{volume}{131}, \bibinfo{pages}{233--240}.
\bibitem[{{Phong} et~al.(2018){Phong}, {Aono}, {Hayashi}, {Wang} and
  {Moriai}}]{8241854}
\bibinfo{author}{{Phong}, L.T.}, \bibinfo{author}{{Aono}, Y.},
  \bibinfo{author}{{Hayashi}, T.}, \bibinfo{author}{{Wang}, L.},
  \bibinfo{author}{{Moriai}, S.}, \bibinfo{year}{2018}.
\newblock \bibinfo{title}{Privacy-preserving deep learning via additively
  homomorphic encryption}.
\newblock \bibinfo{journal}{IEEE Transactions on Information Forensics and
  Security} \bibinfo{volume}{13}, \bibinfo{pages}{1333--1345}.
\bibitem[{Reddi et~al.(2020)Reddi, Charles, Zaheer, Garrett, Rush,
  Kone{\v{c}}n{\`y}, Kumar and McMahan}]{reddi2020adaptive}
\bibinfo{author}{Reddi, S.}, \bibinfo{author}{Charles, Z.},
  \bibinfo{author}{Zaheer, M.}, \bibinfo{author}{Garrett, Z.},
  \bibinfo{author}{Rush, K.}, \bibinfo{author}{Kone{\v{c}}n{\`y}, J.},
  \bibinfo{author}{Kumar, S.}, \bibinfo{author}{McMahan, H.B.},
  \bibinfo{year}{2020}.
\newblock \bibinfo{title}{Adaptive federated optimization}.
\newblock \bibinfo{journal}{arXiv preprint arXiv:2003.00295} .
\bibitem[{Ruck et~al.(1990)Ruck, Rogers and Kabrisky}]{ruck1990feature}
\bibinfo{author}{Ruck, D.W.}, \bibinfo{author}{Rogers, S.K.},
  \bibinfo{author}{Kabrisky, M.}, \bibinfo{year}{1990}.
\newblock \bibinfo{title}{Feature selection using a multilayer perceptron}.
\newblock \bibinfo{journal}{Journal of Neural Network Computing}
  \bibinfo{volume}{2}, \bibinfo{pages}{40--48}.
\bibitem[{Sahu et~al.(2018)Sahu, Li, Sanjabi, Zaheer, Talwalkar and
  Smith}]{DBLP:journals/corr/abs-1812-06127}
\bibinfo{author}{Sahu, A.K.}, \bibinfo{author}{Li, T.},
  \bibinfo{author}{Sanjabi, M.}, \bibinfo{author}{Zaheer, M.},
  \bibinfo{author}{Talwalkar, A.}, \bibinfo{author}{Smith, V.},
  \bibinfo{year}{2018}.
\newblock \bibinfo{title}{On the convergence of federated optimization in
  heterogeneous networks}.
\newblock \bibinfo{journal}{CoRR} \bibinfo{volume}{abs/1812.06127}.
\newblock \URLprefix \url{http://arxiv.org/abs/1812.06127},
  \href{http://arxiv.org/abs/1812.06127}{\tt arXiv:1812.06127}.
\bibitem[{Sattler et~al.(2020)Sattler, M{\"u}ller and
  Samek}]{sattler2020clustered}
\bibinfo{author}{Sattler, F.}, \bibinfo{author}{M{\"u}ller, K.R.},
  \bibinfo{author}{Samek, W.}, \bibinfo{year}{2020}.
\newblock \bibinfo{title}{Clustered federated learning: Model-agnostic
  distributed multitask optimization under privacy constraints}.
\newblock \bibinfo{journal}{IEEE Transactions on Neural Networks and Learning
  Systems} .
\bibitem[{Sattler et~al.(2019)Sattler, Wiedemann, M{\"u}ller and
  Samek}]{sattler2019robust}
\bibinfo{author}{Sattler, F.}, \bibinfo{author}{Wiedemann, S.},
  \bibinfo{author}{M{\"u}ller, K.R.}, \bibinfo{author}{Samek, W.},
  \bibinfo{year}{2019}.
\newblock \bibinfo{title}{Robust and communication-efficient federated learning
  from non-iid data}.
\newblock \bibinfo{journal}{IEEE transactions on neural networks and learning
  systems} \bibinfo{volume}{31}, \bibinfo{pages}{3400--3413}.
\bibitem[{Seide et~al.(2014)Seide, Fu, Droppo, Li and Yu}]{seide20141}
\bibinfo{author}{Seide, F.}, \bibinfo{author}{Fu, H.}, \bibinfo{author}{Droppo,
  J.}, \bibinfo{author}{Li, G.}, \bibinfo{author}{Yu, D.},
  \bibinfo{year}{2014}.
\newblock \bibinfo{title}{1-bit stochastic gradient descent and its application
  to data-parallel distributed training of speech dnns}, in:
  \bibinfo{booktitle}{Fifteenth Annual Conference of the International Speech
  Communication Association}.
\bibitem[{Seif et~al.(2020)Seif, Tandon and Li}]{seif2020wireless}
\bibinfo{author}{Seif, M.}, \bibinfo{author}{Tandon, R.}, \bibinfo{author}{Li,
  M.}, \bibinfo{year}{2020}.
\newblock \bibinfo{title}{Wireless federated learning with local differential
  privacy}, in: \bibinfo{booktitle}{2020 IEEE International Symposium on
  Information Theory (ISIT)}, \bibinfo{organization}{IEEE}. pp.
  \bibinfo{pages}{2604--2609}.
\bibitem[{Shallue et~al.(2018)Shallue, Lee, Antognini, Sohl-Dickstein, Frostig
  and Dahl}]{shallue2018measuring}
\bibinfo{author}{Shallue, C.J.}, \bibinfo{author}{Lee, J.},
  \bibinfo{author}{Antognini, J.}, \bibinfo{author}{Sohl-Dickstein, J.},
  \bibinfo{author}{Frostig, R.}, \bibinfo{author}{Dahl, G.E.},
  \bibinfo{year}{2018}.
\newblock \bibinfo{title}{Measuring the effects of data parallelism on neural
  network training}.
\newblock \bibinfo{journal}{arXiv preprint arXiv:1811.03600} .
\bibitem[{Shamir(1979)}]{shamir1979share}
\bibinfo{author}{Shamir, A.}, \bibinfo{year}{1979}.
\newblock \bibinfo{title}{How to share a secret}.
\newblock \bibinfo{journal}{Communications of the ACM} \bibinfo{volume}{22},
  \bibinfo{pages}{612--613}.
\bibitem[{Shi et~al.(2020)Shi, Yang, Jiang, Zhang and
  Letaief}]{shi2020communication}
\bibinfo{author}{Shi, Y.}, \bibinfo{author}{Yang, K.}, \bibinfo{author}{Jiang,
  T.}, \bibinfo{author}{Zhang, J.}, \bibinfo{author}{Letaief, K.B.},
  \bibinfo{year}{2020}.
\newblock \bibinfo{title}{Communication-efficient edge ai: Algorithms and
  systems}.
\newblock \bibinfo{journal}{IEEE Communications Surveys \& Tutorials}
  \bibinfo{volume}{22}, \bibinfo{pages}{2167--2191}.
\bibitem[{Shin et~al.(2020)Shin, Hwang, Kim, Park, Bennis and
  Kim}]{Shin2020XORMP}
\bibinfo{author}{Shin, M.}, \bibinfo{author}{Hwang, C.}, \bibinfo{author}{Kim,
  J.}, \bibinfo{author}{Park, J.}, \bibinfo{author}{Bennis, M.},
  \bibinfo{author}{Kim, S.L.}, \bibinfo{year}{2020}.
\newblock \bibinfo{title}{Xor mixup: Privacy-preserving data augmentation for
  one-shot federated learning}.
\newblock \bibinfo{journal}{ArXiv} \bibinfo{volume}{abs/2006.05148}.
\bibitem[{Shoham et~al.(2019)Shoham, Avidor, Keren, Israel, Benditkis,
  Mor-Yosef and Zeitak}]{shoham2019overcoming}
\bibinfo{author}{Shoham, N.}, \bibinfo{author}{Avidor, T.},
  \bibinfo{author}{Keren, A.}, \bibinfo{author}{Israel, N.},
  \bibinfo{author}{Benditkis, D.}, \bibinfo{author}{Mor-Yosef, L.},
  \bibinfo{author}{Zeitak, I.}, \bibinfo{year}{2019}.
\newblock \bibinfo{title}{Overcoming forgetting in federated learning on
  non-iid data}.
\newblock \bibinfo{journal}{NeurIPS 2019 Workshop on Federated Learning for
  Data Privacy and Confidentiality} .
\bibitem[{Shokri and Shmatikov(2015)}]{shokri2015privacy}
\bibinfo{author}{Shokri, R.}, \bibinfo{author}{Shmatikov, V.},
  \bibinfo{year}{2015}.
\newblock \bibinfo{title}{Privacy-preserving deep learning}, in:
  \bibinfo{booktitle}{Proceedings of the 22nd ACM SIGSAC conference on computer
  and communications security}, pp. \bibinfo{pages}{1310--1321}.
\bibitem[{Singh et~al.(2020)Singh, Zhou, Yang, Ding, Lin and
  Xie}]{singh2020differentially}
\bibinfo{author}{Singh, I.}, \bibinfo{author}{Zhou, H.}, \bibinfo{author}{Yang,
  K.}, \bibinfo{author}{Ding, M.}, \bibinfo{author}{Lin, B.},
  \bibinfo{author}{Xie, P.}, \bibinfo{year}{2020}.
\newblock \bibinfo{title}{Differentially-private federated neural architecture
  search}.
\newblock \bibinfo{journal}{arXiv preprint arXiv:2006.10559} .
\bibitem[{Smith et~al.(2017)Smith, Chiang, Sanjabi and
  Talwalkar}]{NIPS2017_6211080f}
\bibinfo{author}{Smith, V.}, \bibinfo{author}{Chiang, C.K.},
  \bibinfo{author}{Sanjabi, M.}, \bibinfo{author}{Talwalkar, A.S.},
  \bibinfo{year}{2017}.
\newblock \bibinfo{title}{Federated multi-task learning}, in:
  \bibinfo{booktitle}{Advances in Neural Information Processing Systems}, pp.
  \bibinfo{pages}{4424--4434}.
\bibitem[{Su et~al.(2015)Su, Maji, Kalogerakis and
  Learned-Miller}]{su2015multi}
\bibinfo{author}{Su, H.}, \bibinfo{author}{Maji, S.},
  \bibinfo{author}{Kalogerakis, E.}, \bibinfo{author}{Learned-Miller, E.},
  \bibinfo{year}{2015}.
\newblock \bibinfo{title}{Multi-view convolutional neural networks for 3d shape
  recognition}, in: \bibinfo{booktitle}{Proceedings of the IEEE international
  conference on computer vision}, pp. \bibinfo{pages}{945--953}.
\bibitem[{Sutskever et~al.(2013)Sutskever, Martens, Dahl and
  Hinton}]{sutskever2013importance}
\bibinfo{author}{Sutskever, I.}, \bibinfo{author}{Martens, J.},
  \bibinfo{author}{Dahl, G.}, \bibinfo{author}{Hinton, G.},
  \bibinfo{year}{2013}.
\newblock \bibinfo{title}{On the importance of initialization and momentum in
  deep learning}, in: \bibinfo{booktitle}{International conference on machine
  learning}, \bibinfo{organization}{PMLR}. pp. \bibinfo{pages}{1139--1147}.
\bibitem[{Tanner and Wong(1987)}]{tanner1987calculation}
\bibinfo{author}{Tanner, M.A.}, \bibinfo{author}{Wong, W.H.},
  \bibinfo{year}{1987}.
\newblock \bibinfo{title}{The calculation of posterior distributions by data
  augmentation}.
\newblock \bibinfo{journal}{Journal of the American statistical Association}
  \bibinfo{volume}{82}, \bibinfo{pages}{528--540}.
\bibitem[{Truex et~al.(2020)Truex, Liu, Chow, Gursoy and Wei}]{truex2020ldp}
\bibinfo{author}{Truex, S.}, \bibinfo{author}{Liu, L.}, \bibinfo{author}{Chow,
  K.H.}, \bibinfo{author}{Gursoy, M.E.}, \bibinfo{author}{Wei, W.},
  \bibinfo{year}{2020}.
\newblock \bibinfo{title}{Ldp-fed: Federated learning with local differential
  privacy}, in: \bibinfo{booktitle}{Proceedings of the Third ACM International
  Workshop on Edge Systems, Analytics and Networking}, pp.
  \bibinfo{pages}{61--66}.
\bibitem[{Tuor et~al.(2020)Tuor, Wang, Ko, Liu and Leung}]{tuor2020overcoming}
\bibinfo{author}{Tuor, T.}, \bibinfo{author}{Wang, S.}, \bibinfo{author}{Ko,
  B.J.}, \bibinfo{author}{Liu, C.}, \bibinfo{author}{Leung, K.K.},
  \bibinfo{year}{2020}.
\newblock \bibinfo{title}{Overcoming noisy and irrelevant data in federated
  learning}.
\newblock \bibinfo{journal}{arXiv e-prints} , \bibinfo{pages}{arXiv--2001}.
\bibitem[{Vepakomma et~al.(2018)Vepakomma, Swedish, Raskar, Gupta and
  Dubey}]{vepakomma2018no}
\bibinfo{author}{Vepakomma, P.}, \bibinfo{author}{Swedish, T.},
  \bibinfo{author}{Raskar, R.}, \bibinfo{author}{Gupta, O.},
  \bibinfo{author}{Dubey, A.}, \bibinfo{year}{2018}.
\newblock \bibinfo{title}{No peek: A survey of private distributed deep
  learning}.
\newblock \bibinfo{journal}{arXiv preprint arXiv:1812.03288} .
\bibitem[{Wang et~al.(2020a)Wang, Kaplan, Niu and Li}]{wang2020optimizing}
\bibinfo{author}{Wang, H.}, \bibinfo{author}{Kaplan, Z.}, \bibinfo{author}{Niu,
  D.}, \bibinfo{author}{Li, B.}, \bibinfo{year}{2020}a.
\newblock \bibinfo{title}{Optimizing federated learning on non-iid data with
  reinforcement learning}, in: \bibinfo{booktitle}{IEEE INFOCOM 2020-IEEE
  Conference on Computer Communications}, \bibinfo{organization}{IEEE}. pp.
  \bibinfo{pages}{1698--1707}.
\bibitem[{Wang et~al.(2020b)Wang, Yurochkin, Sun, Papailiopoulos and
  Khazaeni}]{Wang2020Federated}
\bibinfo{author}{Wang, H.}, \bibinfo{author}{Yurochkin, M.},
  \bibinfo{author}{Sun, Y.}, \bibinfo{author}{Papailiopoulos, D.},
  \bibinfo{author}{Khazaeni, Y.}, \bibinfo{year}{2020}b.
\newblock \bibinfo{title}{Federated learning with matched averaging}, in:
  \bibinfo{booktitle}{International Conference on Learning Representations}.
\newblock \URLprefix \url{https://openreview.net/forum?id=BkluqlSFDS}.
\bibitem[{Wang et~al.(2020c)Wang, Liu, Liang, Joshi and
  Poor}]{NEURIPS2020_564127c0}
\bibinfo{author}{Wang, J.}, \bibinfo{author}{Liu, Q.}, \bibinfo{author}{Liang,
  H.}, \bibinfo{author}{Joshi, G.}, \bibinfo{author}{Poor, H.V.},
  \bibinfo{year}{2020}c.
\newblock \bibinfo{title}{Tackling the objective inconsistency problem in
  heterogeneous federated optimization}, in: \bibinfo{editor}{Larochelle, H.},
  \bibinfo{editor}{Ranzato, M.}, \bibinfo{editor}{Hadsell, R.},
  \bibinfo{editor}{Balcan, M.F.}, \bibinfo{editor}{Lin, H.} (Eds.),
  \bibinfo{booktitle}{Advances in Neural Information Processing Systems},
  \bibinfo{publisher}{Curran Associates, Inc.}. pp.
  \bibinfo{pages}{7611--7623}.
\newblock \URLprefix
  \url{https://proceedings.neurips.cc/paper/2020/file/564127c03caab942e503ee6f810f54fd-Paper.pdf}.
\bibitem[{Wang et~al.(2019a)Wang, Mathews, Kiddon, Eichner, Beaufays and
  Ramage}]{wang2019federated}
\bibinfo{author}{Wang, K.}, \bibinfo{author}{Mathews, R.},
  \bibinfo{author}{Kiddon, C.}, \bibinfo{author}{Eichner, H.},
  \bibinfo{author}{Beaufays, F.}, \bibinfo{author}{Ramage, D.},
  \bibinfo{year}{2019}a.
\newblock \bibinfo{title}{Federated evaluation of on-device personalization}.
\newblock \bibinfo{journal}{arXiv preprint arXiv:1910.10252} .
\bibitem[{Wang et~al.(2020d)Wang, Lin, Sheng, Yan and
  Shao}]{DBLP:journals/corr/abs-2011-00826}
\bibinfo{author}{Wang, Z.}, \bibinfo{author}{Lin, C.}, \bibinfo{author}{Sheng,
  L.}, \bibinfo{author}{Yan, J.}, \bibinfo{author}{Shao, J.},
  \bibinfo{year}{2020}d.
\newblock \bibinfo{title}{{PV-NAS:} practical neural architecture search for
  video recognition}.
\newblock \bibinfo{journal}{CoRR} \bibinfo{volume}{abs/2011.00826}.
\newblock \URLprefix \url{https://arxiv.org/abs/2011.00826},
  \href{http://arxiv.org/abs/2011.00826}{\tt arXiv:2011.00826}.
\bibitem[{Wang et~al.(2019b)Wang, Song, Zhang, Song, Wang and
  Qi}]{wang2019beyond}
\bibinfo{author}{Wang, Z.}, \bibinfo{author}{Song, M.}, \bibinfo{author}{Zhang,
  Z.}, \bibinfo{author}{Song, Y.}, \bibinfo{author}{Wang, Q.},
  \bibinfo{author}{Qi, H.}, \bibinfo{year}{2019}b.
\newblock \bibinfo{title}{Beyond inferring class representatives: User-level
  privacy leakage from federated learning}, in: \bibinfo{booktitle}{IEEE
  INFOCOM 2019-IEEE Conference on Computer Communications},
  \bibinfo{organization}{IEEE}. pp. \bibinfo{pages}{2512--2520}.
\bibitem[{Webank(2019)}]{fate}
\bibinfo{author}{Webank}, \bibinfo{year}{2019}.
\newblock \bibinfo{title}{Fate: an industrial grade federated learning
  framework}.
\newblock \bibinfo{howpublished}{\url{https://fate.fedai.org/.}}
\bibitem[{WeBankFinTech(2019)}]{webank2019fate}
\bibinfo{author}{WeBankFinTech}, \bibinfo{year}{2019}.
\newblock \bibinfo{title}{Webank: Fate}.
\newblock \URLprefix \url{https://github.com/webankfintech/fate}.
\bibitem[{Wei et~al.(2020a)Wei, Li, Ding, Ma, Yang, Farokhi, Jin, Quek and
  Poor}]{wei2020federated}
\bibinfo{author}{Wei, K.}, \bibinfo{author}{Li, J.}, \bibinfo{author}{Ding,
  M.}, \bibinfo{author}{Ma, C.}, \bibinfo{author}{Yang, H.H.},
  \bibinfo{author}{Farokhi, F.}, \bibinfo{author}{Jin, S.},
  \bibinfo{author}{Quek, T.Q.}, \bibinfo{author}{Poor, H.V.},
  \bibinfo{year}{2020}a.
\newblock \bibinfo{title}{Federated learning with differential privacy:
  Algorithms and performance analysis}.
\newblock \bibinfo{journal}{IEEE Transactions on Information Forensics and
  Security} \bibinfo{volume}{15}, \bibinfo{pages}{3454--3469}.
\bibitem[{Wei et~al.(2020b)Wei, Liu, Loper, Chow, Gursoy, Truex and
  Wu}]{wei2020framework}
\bibinfo{author}{Wei, W.}, \bibinfo{author}{Liu, L.}, \bibinfo{author}{Loper,
  M.}, \bibinfo{author}{Chow, K.H.}, \bibinfo{author}{Gursoy, M.E.},
  \bibinfo{author}{Truex, S.}, \bibinfo{author}{Wu, Y.}, \bibinfo{year}{2020}b.
\newblock \bibinfo{title}{A framework for evaluating gradient leakage attacks
  in federated learning}.
\newblock \bibinfo{journal}{arXiv preprint arXiv:2004.10397} .
\bibitem[{Wen et~al.(2017)Wen, Xu, Yan, Wu, Wang, Chen and
  Li}]{wen2017terngrad}
\bibinfo{author}{Wen, W.}, \bibinfo{author}{Xu, C.}, \bibinfo{author}{Yan, F.},
  \bibinfo{author}{Wu, C.}, \bibinfo{author}{Wang, Y.}, \bibinfo{author}{Chen,
  Y.}, \bibinfo{author}{Li, H.}, \bibinfo{year}{2017}.
\newblock \bibinfo{title}{Terngrad: Ternary gradients to reduce communication
  in distributed deep learning}, in: \bibinfo{booktitle}{Proceedings of the
  31st International Conference on Neural Information Processing Systems},
  \bibinfo{publisher}{Curran Associates Inc.}, \bibinfo{address}{Red Hook, NY,
  USA}. p. \bibinfo{pages}{1508–1518}.
\bibitem[{Wu et~al.(2020)Wu, Cai, Xiao, Chen and Ooi}]{wu2020privacy}
\bibinfo{author}{Wu, Y.}, \bibinfo{author}{Cai, S.}, \bibinfo{author}{Xiao,
  X.}, \bibinfo{author}{Chen, G.}, \bibinfo{author}{Ooi, B.C.},
  \bibinfo{year}{2020}.
\newblock \bibinfo{title}{Privacy preserving vertical federated learning for
  tree-based models}.
\newblock \bibinfo{journal}{arXiv preprint arXiv:2008.06170} .
\bibitem[{Wu and He(2018)}]{wu2018group}
\bibinfo{author}{Wu, Y.}, \bibinfo{author}{He, K.}, \bibinfo{year}{2018}.
\newblock \bibinfo{title}{Group normalization}, in:
  \bibinfo{booktitle}{Proceedings of the European conference on computer vision
  (ECCV)}, pp. \bibinfo{pages}{3--19}.
\bibitem[{Xu et~al.(2020a)Xu, Du, Jin, He and Cheng}]{xu2020}
\bibinfo{author}{Xu, J.}, \bibinfo{author}{Du, W.}, \bibinfo{author}{Jin, Y.},
  \bibinfo{author}{He, W.}, \bibinfo{author}{Cheng, R.}, \bibinfo{year}{2020}a.
\newblock \bibinfo{title}{Ternary compression for communication-efficient
  federated learning}.
\newblock \bibinfo{journal}{IEEE Transactions on Neural Networks and Learning
  Systems} .
\bibitem[{Xu et~al.(2021a)Xu, Glicksberg, Su, Walker, Bian and
  Wang}]{xujie2021federated}
\bibinfo{author}{Xu, J.}, \bibinfo{author}{Glicksberg, B.S.},
  \bibinfo{author}{Su, C.}, \bibinfo{author}{Walker, P.},
  \bibinfo{author}{Bian, J.}, \bibinfo{author}{Wang, F.},
  \bibinfo{year}{2021}a.
\newblock \bibinfo{title}{Federated learning for healthcare informatics}.
\newblock \bibinfo{journal}{Journal of Healthcare Informatics Research}
  \bibinfo{volume}{5}, \bibinfo{pages}{1--19}.
\bibitem[{Xu et~al.(2021b)Xu, Jin, Du and Gu}]{xu2021federated}
\bibinfo{author}{Xu, J.}, \bibinfo{author}{Jin, Y.}, \bibinfo{author}{Du, W.},
  \bibinfo{author}{Gu, S.}, \bibinfo{year}{2021}b.
\newblock \bibinfo{title}{A federated data-driven evolutionary algorithm}.
\newblock \bibinfo{journal}{arXiv preprint arXiv:2102.08288} .
\bibitem[{Xu et~al.(2020b)Xu, Zhao, Bian, Huang, Mei and
  Liu}]{DBLP:journals/corr/abs-2002-06352}
\bibinfo{author}{Xu, M.}, \bibinfo{author}{Zhao, Y.}, \bibinfo{author}{Bian,
  K.}, \bibinfo{author}{Huang, G.}, \bibinfo{author}{Mei, Q.},
  \bibinfo{author}{Liu, X.}, \bibinfo{year}{2020}b.
\newblock \bibinfo{title}{Neural architecture search over decentralized data}.
\newblock \bibinfo{journal}{CoRR} \bibinfo{volume}{abs/2002.06352}.
\newblock \URLprefix \url{https://arxiv.org/abs/2002.06352},
  \href{http://arxiv.org/abs/2002.06352}{\tt arXiv:2002.06352}.
\bibitem[{Xu et~al.(2019)Xu, Baracaldo, Zhou, Anwar and
  Ludwig}]{xu2019hybridalpha}
\bibinfo{author}{Xu, R.}, \bibinfo{author}{Baracaldo, N.},
  \bibinfo{author}{Zhou, Y.}, \bibinfo{author}{Anwar, A.},
  \bibinfo{author}{Ludwig, H.}, \bibinfo{year}{2019}.
\newblock \bibinfo{title}{Hybridalpha: An efficient approach for
  privacy-preserving federated learning}, in: \bibinfo{booktitle}{Proceedings
  of the 12th ACM Workshop on Artificial Intelligence and Security}, pp.
  \bibinfo{pages}{13--23}.
\bibitem[{Yadan et~al.(2013)Yadan, Adams, Taigman and Ranzato}]{yadan2013multi}
\bibinfo{author}{Yadan, O.}, \bibinfo{author}{Adams, K.},
  \bibinfo{author}{Taigman, Y.}, \bibinfo{author}{Ranzato, M.},
  \bibinfo{year}{2013}.
\newblock \bibinfo{title}{Multi-gpu training of convnets}.
\newblock \bibinfo{journal}{arXiv preprint arXiv:1312.5853} .
\bibitem[{Yang et~al.(2019a)Yang, Fan, Chen, Shi and Yang}]{yang2019quasi}
\bibinfo{author}{Yang, K.}, \bibinfo{author}{Fan, T.}, \bibinfo{author}{Chen,
  T.}, \bibinfo{author}{Shi, Y.}, \bibinfo{author}{Yang, Q.},
  \bibinfo{year}{2019}a.
\newblock \bibinfo{title}{A quasi-newton method based vertical federated
  learning framework for logistic regression}.
\newblock \bibinfo{journal}{arXiv preprint arXiv:1912.00513} .
\bibitem[{Yang et~al.(2019b)Yang, Song, Xu, Li and Tan}]{yang2019tradeoff}
\bibinfo{author}{Yang, M.}, \bibinfo{author}{Song, L.}, \bibinfo{author}{Xu,
  J.}, \bibinfo{author}{Li, C.}, \bibinfo{author}{Tan, G.},
  \bibinfo{year}{2019}b.
\newblock \bibinfo{title}{The tradeoff between privacy and accuracy in anomaly
  detection using federated xgboost}.
\newblock \bibinfo{journal}{arXiv preprint arXiv:1907.07157} .
\bibitem[{Yang et~al.(2019c)Yang, Liu, Chen and Tong}]{yang2019federated}
\bibinfo{author}{Yang, Q.}, \bibinfo{author}{Liu, Y.}, \bibinfo{author}{Chen,
  T.}, \bibinfo{author}{Tong, Y.}, \bibinfo{year}{2019}c.
\newblock \bibinfo{title}{Federated machine learning: Concept and
  applications}.
\newblock \bibinfo{journal}{ACM Transactions on Intelligent Systems and
  Technology (TIST)} \bibinfo{volume}{10}, \bibinfo{pages}{1--19}.
\bibitem[{Yang et~al.(2019d)Yang, Ren, Zhou and Liu}]{yang2019parallel}
\bibinfo{author}{Yang, S.}, \bibinfo{author}{Ren, B.}, \bibinfo{author}{Zhou,
  X.}, \bibinfo{author}{Liu, L.}, \bibinfo{year}{2019}d.
\newblock \bibinfo{title}{Parallel distributed logistic regression for vertical
  federated learning without third-party coordinator}.
\newblock \bibinfo{journal}{arXiv preprint arXiv:1911.09824} .
\bibitem[{Yang et~al.(2021)Yang, Feng, Fang, Shao, Tang, Xia and
  Lu}]{yang2021computationefficient}
\bibinfo{author}{Yang, X.}, \bibinfo{author}{Feng, Y.}, \bibinfo{author}{Fang,
  W.}, \bibinfo{author}{Shao, J.}, \bibinfo{author}{Tang, X.},
  \bibinfo{author}{Xia, S.T.}, \bibinfo{author}{Lu, R.}, \bibinfo{year}{2021}.
\newblock \bibinfo{title}{Computation-efficient deep model training for
  ciphertext-based cross-silo federated learning}.
\newblock \href{http://arxiv.org/abs/2002.09843}{\tt arXiv:2002.09843}.
\bibitem[{Yin et~al.(2020)Yin, Lin, Kong, Xu, Li, Theodoridis and
  Cui}]{yin2020fedloc}
\bibinfo{author}{Yin, F.}, \bibinfo{author}{Lin, Z.}, \bibinfo{author}{Kong,
  Q.}, \bibinfo{author}{Xu, Y.}, \bibinfo{author}{Li, D.},
  \bibinfo{author}{Theodoridis, S.}, \bibinfo{author}{Cui, S.R.},
  \bibinfo{year}{2020}.
\newblock \bibinfo{title}{Fedloc: Federated learning framework for data-driven
  cooperative localization and location data processing}.
\newblock \bibinfo{journal}{IEEE Open Journal of Signal Processing}
  \bibinfo{volume}{1}, \bibinfo{pages}{187--215}.
\bibitem[{Yoon et~al.(2021)Yoon, Shin, Hwang and Yang}]{yoon2021fedmix}
\bibinfo{author}{Yoon, T.}, \bibinfo{author}{Shin, S.}, \bibinfo{author}{Hwang,
  S.J.}, \bibinfo{author}{Yang, E.}, \bibinfo{year}{2021}.
\newblock \bibinfo{title}{Fedmix: Approximation of mixup under mean augmented
  federated learning}, in: \bibinfo{booktitle}{International Conference on
  Learning Representations}.
\newblock \URLprefix \url{https://openreview.net/forum?id=Ogga20D2HO-}.
\bibitem[{Yoshida et~al.(2019)Yoshida, Nishio, Morikura, Yamamoto and
  Yonetani}]{DBLP:journals/corr/abs-1905-07210}
\bibinfo{author}{Yoshida, N.}, \bibinfo{author}{Nishio, T.},
  \bibinfo{author}{Morikura, M.}, \bibinfo{author}{Yamamoto, K.},
  \bibinfo{author}{Yonetani, R.}, \bibinfo{year}{2019}.
\newblock \bibinfo{title}{Hybrid-fl: Cooperative learning mechanism using
  non-iid data in wireless networks}.
\newblock \bibinfo{journal}{CoRR} \bibinfo{volume}{abs/1905.07210}.
\newblock \URLprefix \url{http://arxiv.org/abs/1905.07210},
  \href{http://arxiv.org/abs/1905.07210}{\tt arXiv:1905.07210}.
\bibitem[{Yu et~al.(2020a)Yu, Rawat, Menon and Kumar}]{yu2020federated}
\bibinfo{author}{Yu, F.}, \bibinfo{author}{Rawat, A.S.},
  \bibinfo{author}{Menon, A.}, \bibinfo{author}{Kumar, S.},
  \bibinfo{year}{2020}a.
\newblock \bibinfo{title}{Federated learning with only positive labels}, in:
  \bibinfo{booktitle}{International Conference on Machine Learning},
  \bibinfo{organization}{PMLR}. pp. \bibinfo{pages}{10946--10956}.
\bibitem[{Yu et~al.(2020b)Yu, Zhang, Qin, Xu, Wang, Liu, Tian and
  Chen}]{yu2020heterogeneous}
\bibinfo{author}{Yu, F.}, \bibinfo{author}{Zhang, W.}, \bibinfo{author}{Qin,
  Z.}, \bibinfo{author}{Xu, Z.}, \bibinfo{author}{Wang, D.},
  \bibinfo{author}{Liu, C.}, \bibinfo{author}{Tian, Z.}, \bibinfo{author}{Chen,
  X.}, \bibinfo{year}{2020}b.
\newblock \bibinfo{title}{Heterogeneous federated learning}.
\newblock \bibinfo{journal}{arXiv preprint arXiv:2008.06767} .
\bibitem[{Yu et~al.(2020c)Yu, Bagdasaryan and Shmatikov}]{yu2020salvaging}
\bibinfo{author}{Yu, T.}, \bibinfo{author}{Bagdasaryan, E.},
  \bibinfo{author}{Shmatikov, V.}, \bibinfo{year}{2020}c.
\newblock \bibinfo{title}{Salvaging federated learning by local adaptation}.
\newblock \bibinfo{journal}{arXiv preprint arXiv:2002.04758} .
\bibitem[{Yurochkin et~al.(2019)Yurochkin, Agarwal, Ghosh, Greenewald, Hoang
  and Khazaeni}]{yurochkin2019bayesian}
\bibinfo{author}{Yurochkin, M.}, \bibinfo{author}{Agarwal, M.},
  \bibinfo{author}{Ghosh, S.}, \bibinfo{author}{Greenewald, K.},
  \bibinfo{author}{Hoang, N.}, \bibinfo{author}{Khazaeni, Y.},
  \bibinfo{year}{2019}.
\newblock \bibinfo{title}{Bayesian nonparametric federated learning of neural
  networks}, in: \bibinfo{booktitle}{International Conference on Machine
  Learning}, \bibinfo{organization}{PMLR}. pp. \bibinfo{pages}{7252--7261}.
\bibitem[{Zhang et~al.(2020a)Zhang, Li, Xia, Wang, Yan and Liu}]{254465}
\bibinfo{author}{Zhang, C.}, \bibinfo{author}{Li, S.}, \bibinfo{author}{Xia,
  J.}, \bibinfo{author}{Wang, W.}, \bibinfo{author}{Yan, F.},
  \bibinfo{author}{Liu, Y.}, \bibinfo{year}{2020}a.
\newblock \bibinfo{title}{Batchcrypt: Efficient homomorphic encryption for
  cross-silo federated learning}, in: \bibinfo{booktitle}{2020 {USENIX} Annual
  Technical Conference ({USENIX} {ATC} 20)}, \bibinfo{publisher}{{USENIX}
  Association}. pp. \bibinfo{pages}{493--506}.
\newblock \URLprefix
  \url{https://www.usenix.org/conference/atc20/presentation/zhang-chengliang}.
\bibitem[{Zhang et~al.(2021)Zhang, Xie, Bai, Yu, Li and Gao}]{ZHANG2021106775}
\bibinfo{author}{Zhang, C.}, \bibinfo{author}{Xie, Y.}, \bibinfo{author}{Bai,
  H.}, \bibinfo{author}{Yu, B.}, \bibinfo{author}{Li, W.},
  \bibinfo{author}{Gao, Y.}, \bibinfo{year}{2021}.
\newblock \bibinfo{title}{A survey on federated learning}.
\newblock \bibinfo{journal}{Knowledge-Based Systems} \bibinfo{volume}{216},
  \bibinfo{pages}{106775}.
\newblock \URLprefix
  \url{https://www.sciencedirect.com/science/article/pii/S0950705121000381},
  \DOIprefix\doi{https://doi.org/10.1016/j.knosys.2021.106775}.
\bibitem[{Zhang et~al.(2018a)Zhang, Cisse, Dauphin and
  Lopez-Paz}]{zhang2018mixup}
\bibinfo{author}{Zhang, H.}, \bibinfo{author}{Cisse, M.},
  \bibinfo{author}{Dauphin, Y.N.}, \bibinfo{author}{Lopez-Paz, D.},
  \bibinfo{year}{2018}a.
\newblock \bibinfo{title}{mixup: Beyond empirical risk minimization}.
\newblock \bibinfo{journal}{International Conference on Learning
  Representations} \URLprefix \url{https://openreview.net/forum?id=r1Ddp1-Rb}.
\bibitem[{Zhang et~al.(2018b)Zhang, Xiang, Hospedales and Lu}]{zhang2018deep}
\bibinfo{author}{Zhang, Y.}, \bibinfo{author}{Xiang, T.},
  \bibinfo{author}{Hospedales, T.M.}, \bibinfo{author}{Lu, H.},
  \bibinfo{year}{2018}b.
\newblock \bibinfo{title}{Deep mutual learning}, in:
  \bibinfo{booktitle}{Proceedings of the IEEE Conference on Computer Vision and
  Pattern Recognition}, pp. \bibinfo{pages}{4320--4328}.
\bibitem[{Zhang et~al.(2020b)Zhang, Yang, Yao, Yan, Gonzalez and
  Mahoney}]{zhang2020improving}
\bibinfo{author}{Zhang, Z.}, \bibinfo{author}{Yang, Y.}, \bibinfo{author}{Yao,
  Z.}, \bibinfo{author}{Yan, Y.}, \bibinfo{author}{Gonzalez, J.E.},
  \bibinfo{author}{Mahoney, M.W.}, \bibinfo{year}{2020}b.
\newblock \bibinfo{title}{Improving semi-supervised federated learning by
  reducing the gradient diversity of models}.
\newblock \bibinfo{journal}{arXiv preprint arXiv:2008.11364} .
\bibitem[{Zhao et~al.(2020a)Zhao, Mopuri and Bilen}]{zhao2020idlg}
\bibinfo{author}{Zhao, B.}, \bibinfo{author}{Mopuri, K.R.},
  \bibinfo{author}{Bilen, H.}, \bibinfo{year}{2020}a.
\newblock \bibinfo{title}{idlg: Improved deep leakage from gradients}.
\newblock \bibinfo{journal}{arXiv preprint arXiv:2001.02610} .
\bibitem[{Zhao et~al.(2018)Zhao, Li, Lai, Suda, Civin and
  Chandra}]{zhao2018federated}
\bibinfo{author}{Zhao, Y.}, \bibinfo{author}{Li, M.}, \bibinfo{author}{Lai,
  L.}, \bibinfo{author}{Suda, N.}, \bibinfo{author}{Civin, D.},
  \bibinfo{author}{Chandra, V.}, \bibinfo{year}{2018}.
\newblock \bibinfo{title}{Federated learning with non-iid data}.
\newblock \bibinfo{journal}{arXiv preprint arXiv:1806.00582} .
\bibitem[{Zhao et~al.(2020b)Zhao, Zhao, Yang, Wang, Wang, Lyu, Niyato and
  Lam}]{zhao2020local}
\bibinfo{author}{Zhao, Y.}, \bibinfo{author}{Zhao, J.}, \bibinfo{author}{Yang,
  M.}, \bibinfo{author}{Wang, T.}, \bibinfo{author}{Wang, N.},
  \bibinfo{author}{Lyu, L.}, \bibinfo{author}{Niyato, D.},
  \bibinfo{author}{Lam, K.Y.}, \bibinfo{year}{2020}b.
\newblock \bibinfo{title}{Local differential privacy based federated learning
  for internet of things}.
\newblock \bibinfo{journal}{IEEE Internet of Things Journal} .
\bibitem[{Zhu and Jin(2019)}]{zhu2019multi}
\bibinfo{author}{Zhu, H.}, \bibinfo{author}{Jin, Y.}, \bibinfo{year}{2019}.
\newblock \bibinfo{title}{Multi-objective evolutionary federated learning}.
\newblock \bibinfo{journal}{IEEE transactions on neural networks and learning
  systems} \bibinfo{volume}{31}, \bibinfo{pages}{1310--1322}.
\bibitem[{Zhu and Jin(2020)}]{zhu2020real}
\bibinfo{author}{Zhu, H.}, \bibinfo{author}{Jin, Y.}, \bibinfo{year}{2020}.
\newblock \bibinfo{title}{Real-time federated evolutionary neural architecture
  search}.
\newblock \bibinfo{journal}{arXiv preprint arXiv:2003.02793} .
\bibitem[{Zhu et~al.(2020)Zhu, Wang, Jin, Liang and Ning}]{zhu2020distributed}
\bibinfo{author}{Zhu, H.}, \bibinfo{author}{Wang, R.}, \bibinfo{author}{Jin,
  Y.}, \bibinfo{author}{Liang, K.}, \bibinfo{author}{Ning, J.},
  \bibinfo{year}{2020}.
\newblock \bibinfo{title}{Distributed additive encryption and quantization for
  privacy preserving federated deep learning}.
\newblock \bibinfo{journal}{arXiv preprint arXiv:2011.12623} .
\bibitem[{Zhu et~al.(2021)Zhu, Zhang and Jin}]{zhu2021FNAS}
\bibinfo{author}{Zhu, H.}, \bibinfo{author}{Zhang, H.}, \bibinfo{author}{Jin,
  Y.}, \bibinfo{year}{2021}.
\newblock \bibinfo{title}{From federated learning to federated neural
  architecture search: {A} survey}.
\newblock \bibinfo{journal}{Complex \& Intelligent Systems}
  \bibinfo{volume}{7}, \bibinfo{pages}{639–657}.
\bibitem[{Zhu and Han(2020)}]{Zhu2020}
\bibinfo{author}{Zhu, L.}, \bibinfo{author}{Han, S.}, \bibinfo{year}{2020}.
\newblock \bibinfo{title}{Deep Leakage from Gradients}.
  \bibinfo{publisher}{Springer International Publishing},
  \bibinfo{address}{Cham}.
\newblock pp. \bibinfo{pages}{17--31}.

\end{thebibliography}








\end{document}